\documentclass[conference, twocolumn]{IEEEtran}
\usepackage{microtype}
\usepackage[hyphens]{url}

\usepackage{braket}
\usepackage{listings}
\usepackage{graphicx}
\usepackage{subcaption}
\usepackage[table,xcdraw]{xcolor}
\usepackage{amsmath}
\usepackage{pifont}
\usepackage{amssymb}
\definecolor{mydarkblue}{rgb}{0,0.08,0.65}
\usepackage[colorlinks=true,
    linkcolor=mydarkblue,
    citecolor=mydarkblue,
    filecolor=mydarkblue,
    urlcolor=mydarkblue]{hyperref} 
\usepackage{cleveref}
\usepackage{soul}
\usepackage[shortlabels]{enumitem}
\usepackage{color}
\usepackage{tikz}
\usepackage{balance}
\usepackage{multirow}
\usepackage{comment}
\usepackage{wrapfig,lipsum,booktabs}
\usepackage[frozencache,cachedir=.]{minted}

\usepackage[symbol]{footmisc}
\usepackage{tablefootnote}
\usepackage{tabularx}
\usepackage{placeins}
\usepackage{algorithm}
\usepackage{algpseudocode}
\usepackage{makecell}
\usepackage{cuted}
\usepackage{float}
\usepackage{caption}

\usepackage{multicol}

\usepackage{dblfloatfix}
\usepackage{natbib}

\renewcommand{\arraystretch}{1.2}

\definecolor{codegreen}{rgb}{0,0.6,0}
\definecolor{codegray}{rgb}{0.5,0.5,0.5}
\definecolor{codepurple}{rgb}{0.58,0,0.82}
\definecolor{backcolour}{rgb}{0.95,0.95,0.92}

\makeatletter
\def\blfootnote{\xdef\@thefnmark{}\@footnotetext}
\makeatother

\lstdefinestyle{mystyle}{
  backgroundcolor=\color{backcolour},   commentstyle=\color{codegreen},
  keywordstyle=\color{magenta},
  numberstyle=\tiny\color{codegray},
  stringstyle=\color{codepurple},
  basicstyle=\ttfamily\footnotesize,
  breakatwhitespace=false,         
  breaklines=true,                 
  captionpos=b,                    
  keepspaces=true,                 
  numbers=left,                    
  numbersep=5pt,                  
  showspaces=false,                
  showstringspaces=false,
  showtabs=false,                  
  tabsize=2,
}

\AtBeginDocument{%
  \providecommand\BibTeX{{%
    \normalfont B\kern-0.5em{\scshape i\kern-0.25em b}\kern-0.8em\TeX}}}

\makeatletter
\def\section{\@startsection{section}{1}{\z@}%
  {4.0ex plus 1.5ex minus 0.4ex}% MORE space before
  {1.4ex plus 0.3ex}% space after
  {\centering\normalfont\scshape}}
\makeatother

\IEEEoverridecommandlockouts
\begin{document}

\title{ZAYA1-VL-8B Technical Report}

\newcommand{\corr}{\textsuperscript{*}}

\author{
\IEEEauthorblockN{Hassan Shapourian\corr, 
Kasra Hejazi, 
Olabode M. Sule,
Beren Millidge\corr}
\IEEEauthorblockA{Zyphra Technologies Inc,\\ San Francisco, CA}
\IEEEauthorblockA{\textsuperscript{*}Corresponding authors: \texttt{hassan@zyphra.com}, \texttt{beren@zyphra.com}}
}

\maketitle

\setcounter{page}{1}

\begin{abstract}

We present ZAYA1-VL-8B, a compact mixture-of-experts vision–language model built upon our in-house language model, ZAYA1-8B. Despite its compact size, ZAYA1-VL-8B achieves performance competitive with leading base models such as Molmo2-4B and InternVL3.5-4B, while surpassing models including Qwen2.5-VL-3B, PLM-3B, and MolmoE-8B across a range of image understanding, reasoning, and counting benchmarks. The architecture incorporates two key innovations: (1) vision-specific LoRA adapters integrated into the LLM to increase modality-specific capacity without increasing the number of experts, and (2) bidirectional attention over image tokens within the LLM to enhance visual understanding.
We detail the full training pipeline including data composition at each stage, packing sequences, and the attention masking scheme. 
The model comprises 9.2B total parameters (1.4B active including the vision encoder) and is publicly available at: \href{https://huggingface.co/Zyphra/ZAYA1-VL-8B}{(https://huggingface.co/Zyphra/ZAYA1-VL-8B)}.

\end{abstract}

\section{Introduction}
Vision Language Models (VLMs) have emerged as a central paradigm in multimodal AI, evolving rapidly from dual-encoder architectures to increasingly unified systems capable of rich cross-modal reasoning. Early models like CLIP \cite{Radford2021LearningTV} and BLIP \cite{BLIP-Chauduri} bridged modalities via contrastive learning across separate text and vision encoders, enabling a wide range of vision tasks including zero-shot and few-shot image classification \cite{TZSFSC-Martin, wang2024capsadapter}, Optical Character Recognition (OCR) \cite{ZhaoCLIP4STR}, open-vocabulary detection \cite{ZangOV-Detr} and segmentation \cite{Liang_2023_CVPR}, and semantic image and video retrieval \cite{Schuhmann-LAION-5B, JVIME-E2E-Retr-Bain}. The current generation of VLMs employs a modular framework pairing a vision encoder, such as CLIP \cite{Radford2021LearningTV}, SigLIP-2 \cite{tschannen2025siglip}, SAM \cite{Kirillov_2023_ICCV, wei2025deepseekocrcontextsopticalcompression}, or Dino-V3 \cite{simeoni2025dinov3, deria2026comevlscalingcomplementarymultiencoder}, with a Large Language Model (LLM) \cite{openai2024gpt4technicalreport, yang2025qwen3technicalreport}, leveraging the LLM's powerful and general open-world knowledge, language understanding, and reasoning capabilities for multimodal understanding. These VLMs employ various mechanisms to either extract the necessary information from vision tokens before input to the LLM, or project vision tokens directly into the language embedding space.

In the former category, Flamingo \cite{alayrac2022flamingo} employs cross-attention between input text and vision tokens, using text tokens as queries, thus enabling the text tokens to become vision-aware before being fed to an LLM. BLIP-2 \cite{BLIP-2-Li} introduces a Q-former which makes use of learnable queries to cross-attend to image tokens and extract necessary information. Within the Q-former the learnable queries also interact via self-attention. The output vision-informed query tokens from the Q-former are subsequently concatenated with language tokens and fed as input to an LLM.

Conversely, LLaVA \cite{liu2023visual} introduced an MLP adapter to align vision tokens to the language embedding space. This is the approach followed in many popular VLMs such as Qwen3-VL~\cite{bai2025qwen3vltechnicalreport}, InternVL3 \cite{zhu2025internvl3exploringadvancedtraining}, GLM4.5 \cite{vteam2026glm45vglm41vthinkingversatilemultimodal}, and Molmo \cite{deitke2025molmo}. These vision tokens are then fed as vision embeddings to the LLM without other architectural changes and the LLM itself is finetuned to understand the meaning of these new tokens. 

Beyond the choice of connector, several complementary innovations on the vision encoder side, have become central to modern VLM design. Dynamic resolution strategies, popularized by the AnyRes technique in LLaVA-NeXT \cite{liu2024llavanext}, allow VLMs to process images at their native aspect ratio and resolution by partitioning them into fixed-size tiles, a constraint imposed by the vision encoder's absolute position embeddings and fixed context window. While this substantially improves performance on detail-sensitive tasks such as OCR and fine-grained VQA, it introduces redundancy at tile boundaries and limits spatial coherence across the full image. A more recent line of work adapts Rotary Position Embedding (RoPE) from 1D sequences to 2D spatial grids \cite{kexuefm-8397, kexuefm-10040, heo2024ropevit, liu2026spiralrope}, encoding relative spatial positions directly in attention. This enables VLMs such as Qwen2-VL \cite{wang2024qwen2} to process a single image at its native resolution without tiling, yielding strong resolution extrapolation with minimal computational overhead.
Additionally, the growing computational burden of high-resolution and video inputs has spurred significant work on visual token compression \cite{shang2024LLaVA-PruMerge, yang2025pvc, bolya2023tome}, which reduces the number of vision tokens fed to the LLM through pruning, merging, or adaptive selection while preserving task-critical information. More recently and speculatively, `native' VLMs \cite{diao2026from} have attracted growing interest, proposing to discard vision encoders entirely (at the expense of adding more layers to the LLM) and process images and text end-to-end within a single model, pointing toward a future of more tightly integrated multimodal architectures.

The practical impact of VLMs already spans a broad and expanding set of domains: multimodal chatbots and intelligent assistants \cite{openai2023gpt4v}, OCR and document understanding \cite{wei2025deepseekocrcontextsopticalcompression, poznanski2025olmocr2unittest}, medical image analysis \cite{lasateam2025lingshugeneralistfoundationmodel, Chen-MedComplxReasoning}, computer-use agents \cite{wang2025uitars2technicalreportadvancing, wang2025opencua, gupta2026molmowebopenvisualweb}, surveillance monitoring \cite{benschop2025evaluationvisionllmssurveillancevideo}, navigation agents, embodied robotics \cite{pmlr-v270-kim25c, geminiroboticsteam2025geminiroboticsbringingai, ram2025from}, autonomous driving \cite{sima2024drivelm, Shao2023LMDriveCE}, manufacturing and engineering design \cite{LLM-Manuf-Li}, and scientific discovery \cite{yan2025a-sd}.

Despite this remarkable progress, several challenges remain before VLMs can be considered fully mature. These include hallucinations \cite{li-etal-2023-evaluating, kanade-ganu-2026-see, augustin2025dash}, where model reasoning diverges from the reality of visual input; high computational costs \cite{10.1609/aaai.v39i5.32567, shang2024LLaVA-PruMerge} in scenarios involving large volumes of multimodal tokens, as encountered in high-resolution image or video applications; efficient deployment in the cloud or on edge devices \cite{li2025eureka, VLM4Edge-Sharshar}; critical safety challenges \cite{qiu2025efficient}; difficulties with 3D, multi-view, and multi-sensor settings \cite{chen2025scene, SpatialVLM-Chen, HOU2026104314}; and vulnerability to adversarial attacks \cite{zhang2025anyattack, wang2025advedm}.

A further challenge concerns the ecosystem itself. The strongest VLMs today remain proprietary \cite{openai2024gpt4technicalreport, anthropic2024claude35}, and many competitive open-weight alternatives still rely heavily on synthetic data distilled from these closed models \cite{deitke2025molmo, clark2026molmo2}. This dependence limits reproducibility and leaves the community without full visibility into the ingredients that drive strong performance. At the same time, recent work has shown that data quality and curation are at least as important as architectural choices: careful construction of pre-training captions, instruction-tuning mixtures, and task-specific annotations can be the decisive factor in VLM performance \cite{deitke2025molmo, cho2025perceptionlm,an2025llavaonevision15, li2024llava}.

Beyond data, two architectural limitations persist in most current VLMs. First, the standard approach of processing vision tokens with the same causal attention mask used for language is suboptimal: unlike text, image patches have no inherent left-to-right ordering, and causal masking arbitrarily prevents earlier patches from attending to later ones, limiting the model's ability to capture global visual structure \cite{clark2026molmo2}. Second, most VLMs route vision and language tokens through identical model parameters, forcing shared representations to simultaneously serve two modalities with fundamentally different statistical properties. This parameter sharing can create interference, where optimizing for one modality degrades the other, particularly in mixture-of-expert architectures where expert routing is learned primarily from language data. Recent work on modality-specific adaptation \cite{tian2025navil,luo2025mono,lin2026moe} suggests that dedicating a subset of parameters to visual processing can alleviate this tension, but existing approaches often require training additional experts from scratch or significantly increasing model size.
 
In this work, we present ZAYA1-VL-8B, a VLM built on top of ZAYA1-8B~\cite{anthony2025training}, and report on the data engineering and training pipeline which supported its development. ZAYA1-VL-8B addresses the two limitations above with targeted architectural innovations: (1) bidirectional attention over image tokens within the LLM, allowing every image patch to attend to every other patch and restoring the spatial symmetry that causal masking destroys, and (2) vision-specific LoRA adapters integrated into the LLM, which increase modality-specific capacity without adding new experts or substantially growing model size. We detail these innovations alongside the full training pipeline and demonstrate that they enable ZAYA1-VL-8B to be highly competitive with models of comparable size and computational complexity.

Our main contributions are summarized as follows.

\begin{enumerate}
\item We demonstrate that our custom mixture-of-experts model (MoE) LLM ZAYA1-8B-A1B \cite{anthony2025training} is indeed capable of powering a strong VLM.
\item We introduce vision-specific LoRA adapters and bidirectional attention over image tokens as lightweight mechanisms to address modality interference and the limitations of causal masking for vision, respectively, offering a new approach for extending MoE LLMs to fully-fledged VLMs.
\item ZAYA1-VL-8B overperforms strongly on performance per inference flops and also shows sample efficiency in terms of training tokens.
\item We open source ZAYA1-VL-8B as a resource for the research and application development community. The training checkpoints can be found at 
\href{https://huggingface.co/Zyphra/ZAYA1-VL-8B}{(https://huggingface.co/Zyphra/ZAYA1-VL-8B)} and inference code can be found at  \href{https://github.com/Zyphra/transformers/tree/zaya1-vl}{(https://github.com/Zyphra/transformers/tree/zaya1-vl)}.
\end{enumerate}

The rest of our report is organized as follows. In Section~\ref{sec:related-work}, we discuss general VLM design and related work. In Section~\ref{sec:architecture} we describe the model architecture of ZAYA1-VL-8B, including the vision encoder, the connector module, and the integration with the ZAYA1-8B LLM backbone. Section~\ref{sec:training} details our multi-stage training pipeline, covering pre-training data curation, alignment training, and supervised fine-tuning. Section~\ref{sec:evaluation} presents a comprehensive evaluation of ZAYA1-VL-8B across a range of benchmarks spanning VQA, OCR, and visual grounding, with comparisons to models of similar size and computational cost. Section~\ref{sec:ablations} provides ablation studies analyzing the impact of key design choices on downstream performance. Finally, Section~\ref{sec:conclusions} summarizes our findings and discusses directions for future work. We provide additional details on training data composition, random examples from training datasets, and model responses to benchmark questions in the appendices.

\begin{figure*}[htbp]
    \centering
    \includegraphics[width=0.9\linewidth]{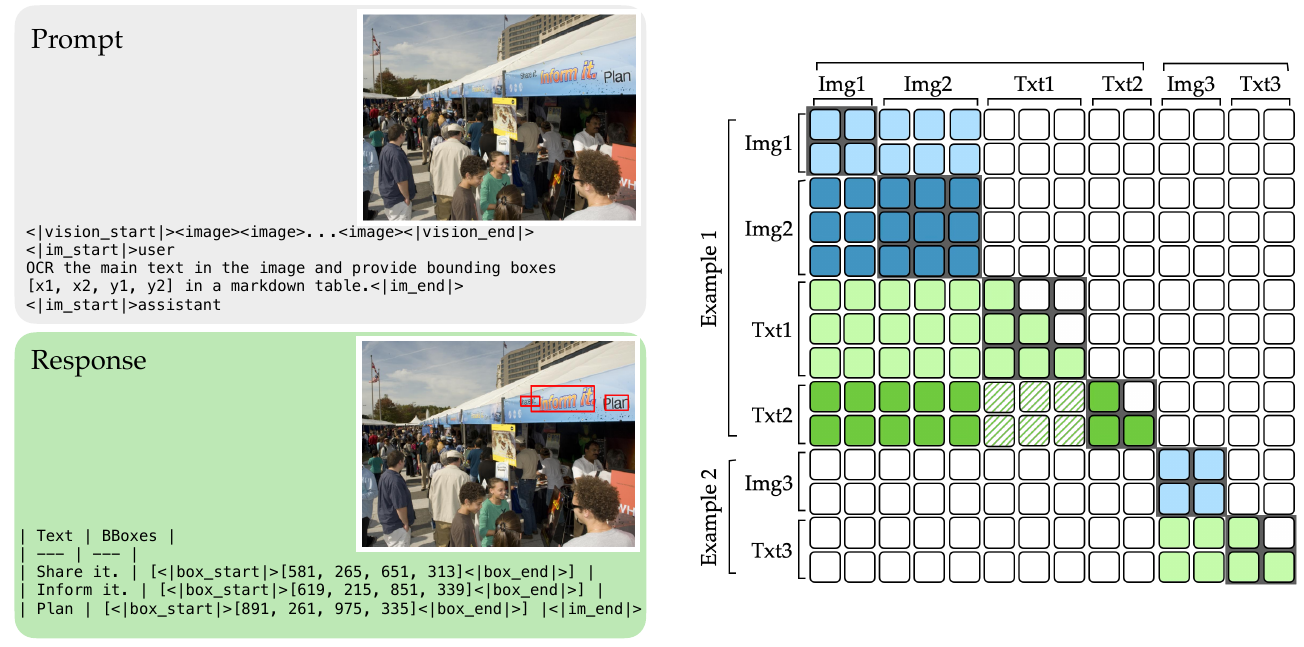}
    \caption{Left: Model chat template and a sample response. Note the model can give detailed grounding and bounding-boxes as well as having standard OCR and visual captioning/understanding. Right: Attention mask for a packed sequence of two image-text examples. Shaded green in second conversation of Example1 (Txt2) shows optional conversation masking.}
    \label{fig:attention-mask}
\end{figure*}

\section{Related Work}
\label{sec:related-work}

\subsection{VLM problem formulation}
The vision language modeling with a vision encoder and an LLM backbone is usually formulated \cite{liu2023visual, deitke2025molmo, cho2025perceptionlm} within the same basic paradigm as language modeling. There is a stream of tokens and the model autoregressively predicts these tokens in the standard way. The only difference is some tokens are vision tokens while others remain text. The VLM only generates text tokens. We focus on transformer-based architectures that concatenate, or more generally interleave, vision tokens with input text tokens. Raw vision inputs $\{u^i\}$ are processed by a vision encoder $\mathcal{F}$, typically a transformer-based architecture followed by a connector module such as an MLP, to produce vision tokens aligned to the language embedding space,
\begin{align}
v^i_j = {\mathcal{F}}(\{u^i\}; \theta),
\label{eq:vision_output}
\end{align}
where the upper index $i$ ranges over different raw visual inputs such as images and videos, the index $j$ ranges over the resulting tokens, and $\theta$ denotes the combined parameters of the vision encoder and connector. The connector may also compress the token sequence before it is fed to the LLM. In practice, the vision encoder is typically loaded with pretrained weights, while the connector is newly initialized for each LLM and trained alongside it.

For VLMs the language modeling task is that the model must output text tokens $y$ given an interleaved set of vision tokens $v$ and prompt text tokens $x$ as input. Unlike in regular LLM autoregression where there is only one type of token, in VLMs the model is not expected to generate vision tokens and this is achieved by masking out the vision tokens from the loss. Mathematically, this autoregressive objective can be written as,
\begin{align*}
{\mathcal{P}}(y_0, y_1, &\dots, y_n \mid v_0, v_1, \dots, v_k, x_0, x_1, \dots, x_l; \psi) \\
\quad= {\mathcal{P}}(y_0 &\mid v_0, v_1, \dots, v_k, x_0, x_1, \dots, x_l; \psi) \notag\\
\qquad\times \prod_{i=1}^{n}
{\mathcal{P}}&\!\left(
y_i \mid
v_0, \dots, v_k,\,
x_0, \dots, x_l,\,
y_0, \dots, y_{i-1};
\psi
\right),
\end{align*}
where $\{v_k\}$ represent all vision tokens and we have omitted the superscript index from Eq.~(\ref{eq:vision_output}), $\{x_l\}$ represent input text tokens, and $\{y_n\}$ represents output text tokens which are available as ground truth during supervised fine tuning or instruction tuning, and $\psi$ are the parameters of an LLM.

The overall learning objective is, therefore,
\begin{align*}
\operatorname*{arg\,max}_{\theta,\psi}\;
{\mathcal{P}}\!\left(
y_0,y_1,\dots,y_n
\,\middle|\,
v_0,v_1,\dots,v_k,\,
x_0,x_1,\dots,x_l;\psi
\right),
\end{align*}
where the dependence on the vision encoder and connector parameters, $\theta$ arise through the vision tokens $\{v_k\}$.

\subsection{Position embeddings}
 
Self-attention is natively permutation equivariant, requiring positional encodings to inject order information. While early approaches used fixed or learned absolute position embeddings \cite{vaswani2017attention}, Rotary Position Embeddings (RoPE) \cite{Roformer-Su} have become the dominant choice in modern LLMs due to their superior length generalization. RoPE rotates queries and keys based on their sequence position so that attention scores depend only on relative position.%, parameterized by a base frequency $b$ and the hidden dimension size $d$.
 
However, standard RoPE is defined over a 1D sequence and does not naturally extend to the 2D or 3D structure of images and videos. Some implementations \cite{deitke2025molmo, clark2026molmo2} simply rasterize the image and apply 1D RoPE, while V2PE \cite{ge2024v2peimprovingmultimodallongcontext} reduces the rate at which position indices increment over visual tokens to account for their greater redundancy relative to text. A family of Multimodal RoPE (MRoPE) approaches \cite{wang2024qwen2, bai2025qwen2.5, bai2025qwen3} instead split the hidden dimensions into separate chunks rotated by 1D RoPE corresponding to time, height, and width respectively, with subsequent work \cite{bai2025qwen3, li2025hope, wei2025videorope, wang2026circlerope, huang2026revisiting} addressing the frequency allocation across these components. Some architectures also inject positional information through other means such as explicit temporal tokens \cite{bai2025qwen2.5} or end-of-row tokens \cite{li2024llava, deitke2025molmo}.
 
A related challenge is handling the heterogeneous sizes and resolutions of natural images. Early approaches resized and cropped all images to a fixed template, discarding information from high-resolution inputs. To address this, AnyRes \cite{liu2024llavanext} splits images into fixed-size tiles compatible with vision encoders trained on absolute position embeddings, an approach also adopted by other architectures \cite{zhu2025internvl3exploringadvancedtraining, deitke2025molmo}. More recently, vision encoders with 2D RoPE can process images at native or near-native resolution, requiring only that dimensions be divisible by the patch size \cite{kimiteam2025kimivltechnicalreport, bai2025qwen2.5, niu2025nativevisualunderstandingresolving, yang2025kwaikeyevl15technical}.

\subsection{Mixture of Experts}
VLMs are typically designed for a diverse range of tasks and need to be scaled adequately to have sufficient capacity to represent the key features in their datasets. One approach to scaling LLMs while keeping the active compute under control is MoE. The idea is to introduce input dependent conditional computation route \cite{shazeer2017} such that only a fraction of model weights are active per input. MoE architectures have been successfully implemented for LLMs from switch transformer \cite{switch-transformer-Fedus} to more recent models like Mixtral~\cite{jiang2024mixtralexperts} and DeepSeekMoE~\cite{dai2024deepseekmoeultimateexpertspecialization}. In large-scale LLMs, MoE models have become ubiquitous due to their compelling flop-efficiency in both inference and, to a lesser extent, in training \citep{liu2024deepseek,team2025kimi}.

In the context of VLMs, MoE-LLaVA \cite{lin2026moe} explored a strategy for adopting MoE to VLMs and preventing model degradation caused by sparsity, showing competitive performance with models that activate more parameters per token. DeepSeek-VL2 \cite{wu2024deepseekvl2mixtureofexpertsvisionlanguagemodels} leveraged DeepSeekMoE models of various sizes, achieving competitive performance with similar or smaller activated parameters compared to existing open-source dense models. Qwen3-VL \cite{bai2025qwen3} also released two MoE VLM variants in addition to four dense models. The issue of MoE in the vision encoder is analyzed in ViMoE \cite{Han2024ViMoEAE}, where the authors leveraged shared experts to address unreliable routing and enable capture of common knowledge. Separately, recent native multimodal models such as NaViL \cite{tian2025navil} and Mono-InternVL \cite{luo2025mono} employ modality-specific feed-forward layers, routing vision and text tokens through dedicated MLPs rather than a shared network, though without a trainable router as in MoE architectures. Our ZAYA1-VL-8B model also demonstrates the compelling advantages of MoE in VLMs, where we find that the benefits of MoEs in language seamlessly transfer to VLM tasks.

\subsection{VLM training strategies}
VLM training typically follows a multi-stage curriculum: (1) vision encoder pretraining, (2) alignment of the connector to a pretrained LLM, (3) supervised instruction tuning of all components, and (4) reinforcement learning post-training. Most efforts leverage existing pretrained vision encoders, though some \cite{bai2025qwen2.5, bai2025qwen3} train one from scratch, and Penguin-VL \cite{zhang2026penguinvlexploringefficiencylimits} initializes a vision encoder from a small pretrained LLM. During alignment, the vision encoder and LLM are usually frozen; during instruction tuning, all parameters are updated. We leave the fourth stage for future consideration.

This is the general recipe followed in LLaVA \cite{liu2023visual}, Qwen2.5-VL \cite{bai2025qwen2.5}, and Molmo \cite{deitke2025molmo}, though implementations vary. LLaVA-OneVision \cite{li2024llava} adds a high-quality knowledge learning stage after alignment. Molmo2 \cite{clark2026molmo2} includes a long-context SFT stage. PerceptionLM \cite{cho2025perceptionlm} pursues large-scale midtraining with synthetic data before SFT on human-annotated data. A common theme is a training curriculum in which data complexity, in terms of task difficulty or context length \cite{bai2025qwen3}, increases gradually alongside data quality.

Recently, native multimodal pretraining \cite{zhu2025internvl3exploringadvancedtraining, kimiteam2026kimik25visualagentic} has attracted interest, introducing the vision modality early during LLM pretraining. This promises tighter cross-modal integration and avoids inductive biases of separately pretrained vision encoders \cite{diao2026from, shukor2025scalinglaws}, but requires substantially more vision-text data and can destabilize LLM optimization during early training \cite{luo2025mono}.

\subsection{VLM datasets and benchmarks}
Several open source datasets have been released to help with VLM training across various tasks. These include contrastive learning \cite{Schuhmann-LAION-5B}, captioning \cite{Chen2015MicrosoftCC, sharma-etal-2018-conceptual, ShareGPT4V-Chen, zhu2024minigpt} VQA \cite{acharya2019tallyqa}, OCR and text recognition \cite{SynthText-Gupta, ocr-vqa-mishra}, chart and figure understanding \cite{methani2020plotqa, kahou2017figureqa, kafle2018dvqa, yang2025effective}, object detection and grounding \cite{shao2019objects365, kuznetsova2020openimages, refCOCO-Mao}, and graphical user interface (GUI) understanding and computer use \cite{liu2024multiui, wu2024osatlas, GUIWorld-Lei, chen-etal-2025-guicourse}.

To assess the capabilities of VLMs many benchmarks have also been released which test various capabilities such as VQA \cite{yue2024mmmu, yue-etal-2025-mmmu-pro}, STEM understanding and reasoning \cite{lu2023mathvista}, OCR \cite{liu2024ocrbench, poznanski2025olmocr2unittest}, chart and plot understanding \cite{methani2020plotqa}, coding \cite{si-etal-2025-design2code}, video understanding \cite{hu2025videommmuevaluatingknowledgeacquisition, MVBench-Li}, GUI understanding and computer use \cite{Screenspot-Pro-Li, xie2024osworld, he-etal-2024-webvoyager, rawles2025androidworld}, and also general qualities like hallucinations \cite{li-etal-2023-evaluating, Hallusionbench-Guan} and robustness \cite{RBench-Zhao}.

\section{Model Architecture}
\label{sec:architecture}

\begin{figure*}[htbp]
    \centering
    \includegraphics[width=0.7\linewidth]{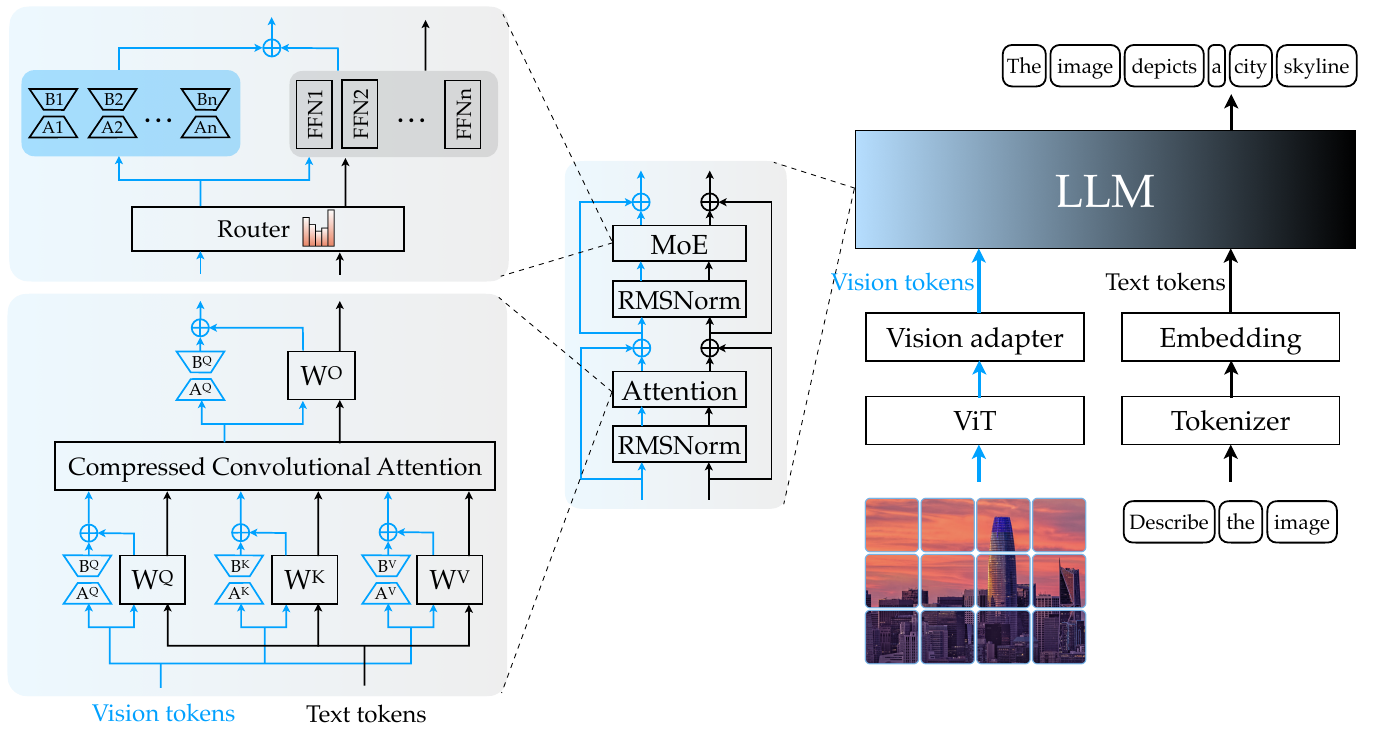}
    \caption{Architecture of ZAYA1-VL-8B. The model uses ZAYA1-8B as the LLM backbone and the Qwen2.5 vision transformer as the vision encoder, connected by a two-layer MLP adapter that projects image features into the language embedding space. Two architectural innovations are introduced: (1) vision-specific LoRA parameters on the MLP and CCA blocks, trained alongside the standard LLM parameters during the main training phase, and (2) bidirectional attention over image tokens within the LLM, allowing all image patches to attend to one another regardless of position.}
    \label{fig:model-arch}
\end{figure*}

ZAYA1-VL-8B builds upon the standard LLaVA-style VLM architecture~\cite{liu2023visual} as shown in Fig.~\ref{fig:model-arch}, comprising four 
components: (1) a decoder-only LLM, (2) a vision encoder that independently 
computes patch embeddings for each input image, (3) an image preprocessor that 
resizes and tiles input raw images into fixed-size patches, and (4) a vision adapter that projects visual features into the LLM's token embedding space.

As the backbone LLM, we adopt ZAYA1-8B-A1B~\cite{anthony2025training}, an MoE model developed 
in-house, which offers state-of-the-art performance per FLOP due to a combination of its novel architecture and training methods and datasets.

For visual encoding, we adopt the vision transformer architecture from Qwen2.5-VL~\cite{bai2025qwen2.5}, 
motivated by its strong empirical performance in our setting. We attribute 
this, in part, to its use of 2D Rotary Position Embeddings (2D RoPE)~\cite{kexuefm-8397,kexuefm-10040}
and its native dynamic resolution processing strategy~\cite{dehghani2023patch,wang2024qwen2}, which avoids fixed-resolution 
distortions and preserves fine-grained spatial structure.
However, we retain the standard 1D RoPE formulation in the LLM rather than adopting a multimodal RoPE formulation. This design choice is driven by empirical observations that such modifications require substantially greater compute and data to realize consistent gains, exceeding the budget allocated for our training setup.

Following Qwen2.5-VL~\cite{bai2025qwen2.5}, each input image is resized to a resolution whose height and width are multiples of 28, while preserving the aspect ratio as much as possible. 
The resized image is next processed by the Vision Transformer (ViT) using a patch size of $14 \times 14$, producing a sequence of patch-level features.
A two-layer MLP adapter then pools each $2 \times 2$ window of patch embeddings into a single vector and projects it into the LLM embedding space, simultaneously reducing the number of vision tokens by a factor of four and aligning their dimensionality with the text embeddings.

Beyond this standard architecture, we further equip the LLM with vision-specific LoRA adapters inserted into linear weights of attention and MLP modules (see Fig.~\ref{fig:model-arch}). These LoRA adapters are only activated when a vision token is passed into the LLM. Our motivation is to increase modality-specific capacity in a parameter-efficient manner. A natural alternative would be to introduce dedicated vision experts into the MoE backbone; however, scaling the number of experts substantially increases the model size and typically demands significantly more training data. Moreover, we find that models trained with shared experts between text and vision perform well already, and thus naturally it is a waste of parameters to use randomly initialized weights for vision experts compared to adapting the existing parameters. Instead, the proposed LoRA adapters provide lightweight vision-specialized pathways (marked in blue in Fig.~\ref{fig:model-arch}) for visual tokens within the LLM, serving as an efficient proxy for modality-specific computation. Note that unlike parameter efficient fine tuning (PEFT), these LoRA adapters are trained {\it alongside} the actual weights in the main training phase; they start from zero and a weight decay term is introduced that prevents vision pathways from diverging indefinitely from text pathways. As shown in our ablation results (Sec.~\ref{sec:ablations}), this design yields a clear performance improvement. For a similar reason, attention weights of the model also use LoRA adapaters for vision tokens to provide vision-specific attention pathways. Again, both the original weights and LoRA weights are trained. We believe this novel approach provides a useful balance of vision-specific and vision-language shared parameters for the model to utilize to both learn the new visual data distribution while preserving more of its original pure language capabilities. 

Our second major architectural change is to use bidirectional attention within the image.  We remove the causal mask for vision tokens within the LLM attention layers, allowing full bidirectional attention among all visual tokens. This design is well-motivated: input images serve as a static conditioning context rather than a temporally ordered sequence, and thus do not exhibit an inherent causal structure. Moreover, this choice aligns the attention pattern of visual tokens inside the LLM with that of the vision encoder, ensuring architectural consistency across modalities.
Text tokens, in contrast, retain standard causal masking and are allowed to attend to all preceding vision tokens as well as prior text tokens. Our attention masking scheme is illustrated in Fig.~\ref{fig:attention-mask}.

During training, examples may contain multi-turn, multi-image conversational data, where multiple question--answer pairs appear for the same set of images (which appear at the beginning of the sequence for each example). Given a multi-turn example for instance, all vision tokens attend to one another bidirectionally (e.g., {Img1} and {Img2} in Example~1 of the right panel in Fig.~\ref{fig:attention-mask}). Subsequent text tokens (e.g., {Txt1} and {Txt2}) attend to all vision tokens via full cross-attention and attend to one another causally. Cross-conversation attention between Txt1 and Txt2 is optionally dropped during training (marked as shaded boxes in the figure). 

During decoding, this custom masking strategy (bidirectional for vision and causal for text) is applied only to the prefill tokens; autoregressive decoding proceeds with a standard causal attention mask.

\section{Training}
\label{sec:training}

\begin{table*}[ht]
\small
\setlength{\tabcolsep}{6pt}
\renewcommand{\arraystretch}{1.12}
\begin{minipage}{\linewidth}
    \centering
    \begin{tabular}{lccccc}
    \toprule
    \textbf{Stage} & \textbf{Training} & \textbf{Total Tokens} & \textbf{Loss Tokens}  & \textbf{Max Seq. Len.} & \textbf{Max Image Res.} \\
    \midrule
    Alignment & Adapter & 230M & 130M & 800 & 0.3MP\\
    Pretraining & Full & 100B & 4B & 16.5k & 0.8MP$\to$6.3MP \\
    Embed expansion & LM Embed Layer & 2.4B & 310M & 16.5k & 6.3MP \\
    Instruction tuning & Full & 34B & 5.2B & 16.5k & 6.3MP \\
    \bottomrule
    \end{tabular}
    \caption{Training stages of ZAYA1-VL-8B. Training proceeds in four stages. Across all stages, images are preserved at their native resolution up to a stage-specific cap, beyond which they are resized. In Stage 1 (Alignment), we train only the MLP adapter with the loss computed over all text tokens. In Stage 2 (Pretraining), we unlock the full model and progressively increase the resolution cap. Stage 3 (Embed Expansion) briefly trains only the LM embedding layer to initialize new chat-template tokens. Stage 4 (Instruction Tuning) performs full training at the highest resolution cap. From Stage 2 onward, the loss is computed exclusively over answer tokens, meaning the model is supervised only on its responses rather than on the input context or question.}
    \label{tab:train-stages}
\end{minipage}
\end{table*}

\subsection{Training Stages}
\label{sec:training-stages}

\begin{figure}[htbp]
    \centering
    \includegraphics[width=0.95\linewidth]{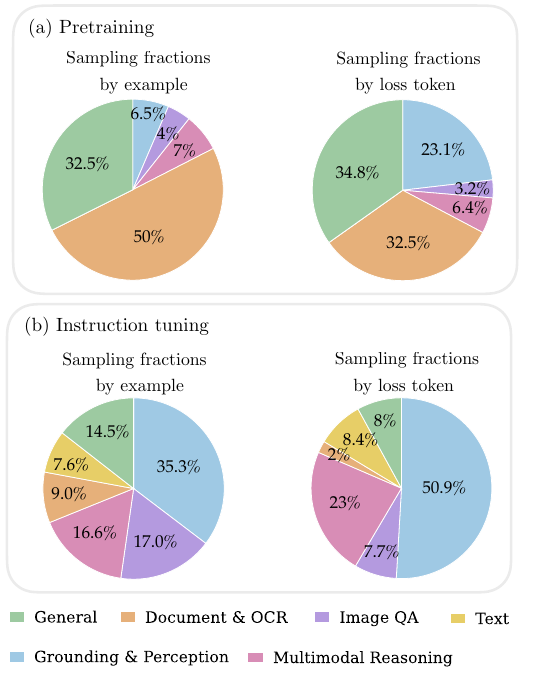}
    \caption{Training data mixtures across the two main training phases: pretraining and instruction-tuning. We report the fraction of each constituent category both by sample count and by the number of answer tokens over which the loss was computed. We distinguish between these two metrics because they can diverge substantially when certain categories contain many short QAs or samples with brief responses. While it is standard to report dataset composition by sample count, we find this can give a misleading picture and prefer the answer-token view, which better reflects what the model actually `sees' during training.}
    \label{fig:data-mixture}
\end{figure}

The model is trained in multiple stages, summarized in Table~\ref{tab:train-stages} which progressively increases the quality of the training data. The pipeline consists of three main phases: (1) alignment, (2) large-scale pretraining (including embedding expansion), and (3) supervised fine-tuning. Data composition for stage (2) and (3) is illustrated in Fig.~\ref{fig:data-mixture}.

\paragraph{Alignment}
We begin by training only the vision adapter on low-resolution image captioning data from LLaVA-ReCap-558K~\cite{llava_recap_558k}, with all LLM parameters frozen. This stage initializes the vision-language interface without disturbing the pretrained language model. We retain the original LLM chat template without introducing new special tokens, and train on short sequences (up to 800 tokens) at low resolution (0.3MP). Despite the frozen LLM, we impose bidirectional attention over vision tokens within the LLM attention modules. We note that our LoRA adapters are not active during this training stage. The goal of this stage is to essentially produce a good initialization for the adapter module before the full VLM training begins.

\paragraph{Pretraining}
We jointly train all model parameters on 30 million multimodal samples. We introduce a new chat template to structure interleaved image-text inputs, shown in Fig.~\ref{fig:attention-mask}. Specifically, \texttt{<|im\_end|>} replaces the original EOS token as the true end-of-sequence marker, while \texttt{<|im\_start|>} serves as a bookkeeping delimiter separating multiple annotations over a shared image set. The base LLM's \texttt{<bos>} token is preserved, as ablations indicate it is important for performance---likely due to its role as an attention sink~\cite{xiao2023efficient}. Finally, \texttt{<|vision\_start|>} and \texttt{<|vision\_end|>} tokens bracket each image, helping the model delineate and distinguish between multiple images, which subsumes the ability to track image count in multi-image inputs. Grounding data at this stage is mostly pointing in the original xml format of PixMo~\cite{deitke2025molmo}.

The maximum allowed image resolution is increased progressively from 0.8MP to 6.3MP (corresponding to $1\text{k}$--$8\text{k}$ vision tokens in the language model) via a stepwise schedule over the first 35\% of training, with the maximum sequence length set to 16.5k tokens to support long-context multimodal reasoning.

We employ the hybrid attention scheme described in Section~\ref{sec:architecture}: vision tokens attend bidirectionally both within and across images, while text tokens are causally masked. Although extending bidirectional attention to question tokens is possible, we find that image-to-question cross-attention introduces undesirable interactions between QA pairs in multi-question examples: under conversation masking, it mixes information across QAs that should be independent, and under causal masking, it allows later answers to see earlier questions through the shared image tokens, effectively violating the causal constraint. We therefore keep image-text attention causally masked throughout.

\paragraph{Embedding Expansion}
Following pretraining, we expand the LLM's embedding layer to accommodate the special tokens required for grounding tasks: \texttt{<|box\_start|>} and \texttt{<|box\_end|>} delimit bounding box coordinates, \texttt{<|point\_start|>} and \texttt{<|point\_end|>} mark point-based object references, and \texttt{<|object\_ref\_start|>} and \texttt{<|object\_ref\_end|>} wrap object mentions that carry either bounding box or point coordinates in grounded responses. Further details on grounding templates are provided in Appendix~\ref{app:grounding}. Only the embedding parameters are updated in this stage, keeping the rest of the model fixed to ensure stable vocabulary integration. We upsample grounding data to constitute 80\% of loss tokens over 3 million examples. This stage in turn prepares the model to train on various formats of the grounding data in the form of referring expressions, bounding boxes and pointing. See Appendix~\ref{app:grounding} for various grounding formats.

\paragraph{Supervised Fine-Tuning (SFT)}
In the final stage, we perform instruction tuning on 20 million curated text and multimodal samples, training end-to-end with the same chat template and attention masking as in pretraining. We introduce grounding data in the form of bounding boxes and pointing here. We use relative coordinate pixels in 0-1000 range for bounding boxes and pointing in this new format, while maintaining 0-100 range (with 1 decimal point) for the xml format to be consistent with our pretraining stage. This is not a bad choice as there is no difference between the significant digits in these two formats and model learns to place a decimal point between the second and third digits when prompted for xml format. See Appendix~\ref{app:grounding} for examples of grounding formats.

Across both pretraining and SFT stages, we apply loss masking such that the cross-entropy loss is computed only over answer tokens, with images and questions treated purely as conditioning context. 
While this scheme implicitly assigns greater weight to longer responses, we find it preferable in practice; in particular, it yields better performance than alternatives that include question or context tokens in the loss.

Our training corpus comprises a heterogeneous mixture of datasets with substantial variation in sequence lengths, making naive padding prohibitively inefficient. We therefore pack multiple examples into fixed-length sequences, using the document- and conversation-level attention masking described in Sec.~\ref{sec:architecture}, implemented efficiently via FlexAttention~\cite{dong2024flex}.

For conversation masking specifically, we apply it stochastically rather than universally. Full conversation masking showed no consistent benefit in our evaluations, and allowing later turns to attend to earlier ones is generally useful for learning multi-turn reasoning over a shared set of images. We therefore apply conversation masking with probability 50\%. An exception is made for grounding data, where earlier turns can inadvertently leak answer-relevant information (such as object counts or locations expressed across different response formats) so we raise the masking probability to 70\% in those cases.

\begin{figure}
    \centering
    \includegraphics[width=0.8\linewidth]{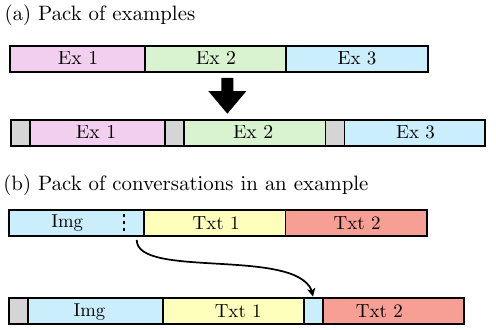}
    \caption{Padding schemes for the CCA module. (a) Each example in a packed sequence is left-padded to prevent convolutions from mixing adjacent examples across document boundaries. (b) For a conversation-masked example containing one image and two QAs, the last few vision tokens are duplicated into the padding region to both isolate the QAs from each other and maintain image-text continuity in the convolutional receptive field.}
    \label{fig:cca-padding}
\end{figure}

Packing introduces an additional complication due to the 1D causal convolution layers in the Compressed Convolutional Attention (CCA)~\cite{figliolia2025compressed} blocks of our LLM. We first note that in CCA, q and k vectors reside in a compressed latent space, but to retain the representability and performance, one performs convolutions both in sequence and channel dimensions (for more details, see~\cite{figliolia2025compressed}). The convolutions in the sequence length are very short range, however for a packed sequence, one should be cautious about different examples leaking into each other through such convolutions; in particular, document boundaries must be respected not only in attention but also in the convolutional receptive field. We address this by left-padding each packed example to enforce proper isolation (Fig.~\ref{fig:cca-padding}(a)). The situation is a bit different for multiple questions based on a single set of images, when we impose conversation masking for later conversations (or QAs) of the same example. In this case, to ensure that each image-question pair is treated as if the question sees the context image independently, we place the corresponding last few vision tokens in the padding region (Fig.~\ref{fig:cca-padding}(b)). 
This ensures that the convolutional receptive field maintains continuity between image and text tokens within each QA while preventing cross-contamination between separate QAs.

\begin{figure}[htbp]
    \centering
    \includegraphics[width=\linewidth]{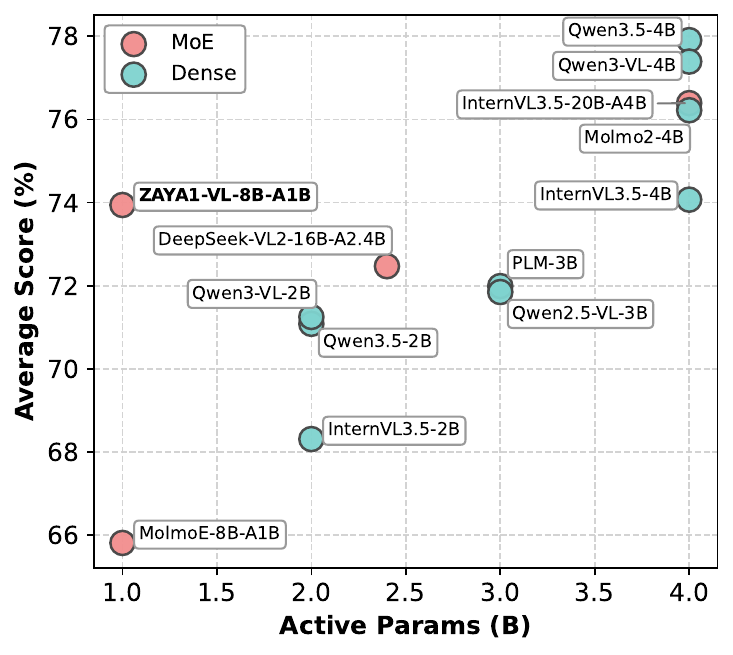}
    \caption{Performance of ZAYA1-VL-8B against models across different parameter scales. Overall average scores are computed from all benchmarks in Table~\ref{tab:comprehensive-eval}. ZAYA1-VL-8B is highly competitive with models of similar active parameter count, particularly against MolmoE, which shares a nearly identical MoE architecture. However, dense models and MoEs with $>$4B active parameters begin to show a clear advantage.}
    \label{fig:scores-vs-params}
\end{figure}

Since the loss is computed only over answer tokens, which constitute a small fraction of the total sequence, the effective batch size in terms of gradient signal is much smaller than in equivalent LLM training. This problem is further accentuated by MoE architectures, where each batch is split across many experts, reducing the number of loss tokens seen by any individual expert. To ensure sufficient gradient signal, we target a minimum of 30k loss tokens per MLP expert per update, which necessitates both a substantially larger batch size than is typical and the use of gradient accumulation. However, even with packing, the number of loss-bearing tokens can vary significantly across devices and microbatches under gradient accumulation. To handle this, rather than normalizing the loss per step, we accumulate the total loss across microbatches and normalize by the total number of answer tokens at each parameter update.

Throughout training, we use the Muon optimizer~\cite{jordan2024muon} for all LLM parameters, consistent with the optimizer used during the base model's pretraining and following the recommendation of~\cite{liu2025muon} to maintain optimizer continuity. The ViT parameters are trained with AdamW to match the optimizer used during its original pretraining. For the vision adapter, which is initialized from scratch, we also adopt Muon based on our ablation results (Section~\ref{sec:ablations}), which show it achieves marginally lower validation loss than AdamW during the alignment stage (see Fig.~\ref{fig:st1-loss-ablations}(b)).

\subsection{Data}

We construct our training recipe by curating and mixing a broad range of open-source datasets, guided by the data strategies developed in PerceptionLM~\cite{cho2025perceptionlm}, Idefics3~\cite{laurencon2024building}, and Molmo~\cite{deitke2025molmo}. We organize the resulting corpus into high-level categories whose proportions vary across training stages, as illustrated in Fig.~\ref{fig:data-mixture}.  Given the aggregate size of the mixture, we stream all data online during training using Mosaic Streaming~\cite{mosaicml_streaming} and apply a greedy bin-packing algorithm to maximize token utilization per batch. During the pretraining stage, general image and document understanding and captioning data constitute a larger portion of our training data. On the other hand, during the instruction tuning stage, we focus more on higher quality data with especial focus on grounding tasks such as bounding-boxes as well as more advanced multimodal reasoning. In appendix \ref{app:dataset_details}, we provide detailed descriptions of our datasets for each phase.

% Requires: \usepackage{booktabs,tabularx,array}
% Shared column types
\newcolumntype{Y}{>{\centering\arraybackslash}X}   % centered, stretchy
\newcolumntype{G}{@{\hspace{8pt}}}                 % fixed gap between groups

\begin{table*}[!htbp]
\centering
\caption{Performance of ZAYA1-VL-8B on general vision evaluations. For DocVQA and InfoVQA we report scores from the original paper since the evaluation requires submission to conference website.}
\label{tab:comprehensive-eval}
\small
\setlength{\tabcolsep}{2pt}
\renewcommand{\arraystretch}{1.12}
\begin{minipage}{\linewidth}\centering

% ===== TOP: best-of (shares columns with bottom) =====
\begin{tabularx}{\linewidth}{l*{10}{Y}G Y Y G Y Y}
\toprule
& \multicolumn{6}{c}{{Chart, Diagram, and Document Understanding}}
  & \multicolumn{6}{c}{{Perception and Reasoning}}
  & \multicolumn{2}{c}{{Counting}}
  \\
\cmidrule(lr){2-7}\cmidrule(lr){8-13}\cmidrule(lr){14-15}
\textbf{Model} 
& \rotatebox{90}{\footnotesize\makecell[l]{\textbf{AI2D}\\(test)~\cite{kembhavi2016diagram}}}
& \rotatebox{90}{\footnotesize\makecell[l]{\textbf{ChartQA}\\(test)~\cite{masry2022chartqa}}}
& \rotatebox{90}{\footnotesize\makecell[l]{\textbf{DocVQA}\\(test)~\cite{mathew2021docvqa}}}
& \rotatebox{90}{\footnotesize\makecell[l]{\textbf{InfoVQA}\\(test)~\cite{mathew2022infographicvqa}}}
& \rotatebox{90}{\footnotesize\makecell[l]{\textbf{TextVQA}\\(val)~\cite{singh2019towards}}}
& \rotatebox{90}{\footnotesize\makecell[l]{\textbf{OCRBench}\\ \cite{liu2024ocrbench}}}
& \rotatebox{90}{\footnotesize\makecell[l]{\textbf{VQA v2.0}\\(val)~\cite{goyal2017making}}}
& \rotatebox{90}{\footnotesize\makecell[l]{\textbf{MathVista}\\(mini)~\cite{lu2023mathvista}}}
& \rotatebox{90}{\footnotesize\makecell[l]{\textbf{MMMU}\\(val)~\cite{yue2024mmmu}}}
& \rotatebox{90}{\footnotesize\makecell[l]{\textbf{SEED}\\(image)~\cite{li2023seed}}}
& \rotatebox{90}{\footnotesize\makecell[l]{\textbf{Blink}\\(val)~\cite{fu2024blink}}}
& \rotatebox{90}{\footnotesize\makecell[l]{\textbf{RealWorldQA}\\ \cite{realworldqa2024}}}
& \rotatebox{90}{\footnotesize\makecell[l]{\textbf{CountBenchQA}\\ \cite{beyer2024paligemma}}}
& \rotatebox{90}{\footnotesize\makecell[l]{\textbf{PixMoCount}\\(test)~\cite{deitke2025molmo}}}
 \\
\midrule
ZAYA1-VL-8B-A1B    & 87.5 & 82.2 & 92.5 & 74 & 74.4 & 79.8 & 80.0 & 64.0 & 46.0 & 72.7 & 45.9 & 65.0 & 88.1 & 83.1 \\
% MolmoE-8B-A1B & 86.4 & 78 & 77.7 & 53.9 & 78.8 & -- & 83.9 & 34 & 34.9 & -- & -- & 60.4 & 87.2 & 79.6  \\
MolmoE-8B-A1B & 73.6 & 77.9 & 77.7 & 53.9 & 78.1 & 55.0 & 82.8 & 39.1 & -- & 68.7 & -- & 60.4 & 77.4 & 45.2  \\
% DeepSeek-VL2-16B-A2.4B & 80 & 84.5 & 92.3 & 75.8 & 83.4 & 83.4 & -- & -- & 48 & -- & --& 65.4 & -- & --  \\
DeepSeek-VL2-16B-A2.4B & 79.6 & 84.6 & 92.3 & 75.8 & 83.4 & 83.3 & 83.7 & 61.2 & 46.0 & 76.8 & 53.3 & 70.0 & 86.0 & 38.6  \\
% InternVL3.5-20B-A4B & 93.5 & 86.6 & 92.9 & 78.1 & 78.5 & 87 & -- & 78 & 72.6 & -- & 59 & 71.2 & -- & -- \\
InternVL3.5-20B-A4B & 85.5 & 87.0 & 92.9 & 78.1 & 78.5 & 86.7 & 78.4 & 73.5 & 72.6 & 76.8 & 58.9 & 71.2 & 82.1 & 47.3 \\

\midrule

% InternVL3.5-2B & 78.8 & 80.7 & 89.4 & 70.8 & 76.5 & 83.6 & -- & 71.8 & 59 & -- & 51.3 & 62 & -- & -- \\
InternVL3.5-2B & 78.9 & 81.6 & 89.4 & 70.8 & 76.5 & 83.4 & 73.6 & 61.4 & 49.9 & 75.2 & 51.3 & 61.6 & 70.0 & 32.8 \\
% Qwen3-VL-2B & 76.9 & 79.1 & 93.3 & 72.4 & -- & 85.8 & -- & 61.3 & 53.4 & -- & 53.8 & 63.9 & 88.4 & -- \\
Qwen3-VL-2B & 77.7 & 78.7 & 93.3 & 72.4 & 79.9 & 84.1 & 78.8 & 51.8 & 40.9 & 74.8 & 53.2 & 66.0 & 87.9 & 55.7 \\
% Qwen3.5-2B & 81.5 &  &  & &  & 85.4 & & 73.9 & 64.2 &  &  & 71.2 & 86.8 &  \\
Qwen3.5-2B & 78.6 & 78.4 & -- & -- & 79.0 & 83.1 & 78.3 & 52.9 & 49.2 & 75.8 & 61.0 & 69.0 & 84.2 & 65.5 \\
% Qwen2.5-VL-3B & 81.6 & 84 & 93.9 & 77.1 & 79.3 & 79.7 & -- & 62.3 & 53.1 & 67.6 & 47.6 & 65.4 & -- & -- \\
Qwen2.5-VL-3B & 79.3 & 83.2 & 93.9 & 77.1 & 79.2 & 82.5 & 79.6 & 63.2 & 45.7 & 73.4 & 48.2 & 65.6 & 77.0 & 60.0 \\
% PLM-3B & 90.9 & 84.3 & 93.8 & 74.6 & 84.3 & 83 & 84.3 & -- & 41.2 & 78.5 & 55.4 & 72.4 & -- & -- \\
PLM-3B & 80.6 & 85.1 & 93.8 & 74.6 & 80.0 & 80.6 & 77.3 & 61.5 & 41.4 & 78.3 & 49.8 & 73.2 & 88.1 & 41.6 \\

\midrule

% Molmo2-4B & 95.6 & 86.1 & 87.8 & 78.6 & 85 & -- & 86.6 & 56.7 & 50.9 & -- & 57.5 & -- & 93.9 & 88.1 \\
Molmo2-4B & 85.4 & 86.1 & 87.8 & 78.6 & 83.1 & 62.0 & 85.3 & 56.5 & 48.8 & 78.0 & 63.5 & 73.8 & 91.2 & 87.0 \\
% Qwen3-VL-4B & 84.1 & 84.6 & 95.3 & 80.3 & -- & 88.1 & -- & 73.7 & 67.4 & -- & 65.8 & 70.9 & 84.9 & -- \\
Qwen3-VL-4B & 84.0 & 81.8 & 95.3 & 80.3 & 81.5 & 84.1 & 80.7 & 63.6 & 51.4 & 77.3 & 63.2 & 71.0 & 87.3 & 89.2 \\
% Qwen3.5-4B & 89.6 &  &  &  &  & 85.0 &  & 85.1 & 77.6 &  &  & 79.5 & 96.3 & -- \\
Qwen3.5-4B & 83.7 & 82.4 & -- & -- & 81.1 & 85.3 & 80.4 & 82.3 & 56.9 & 76.6 & 56.8 & 74.2 & 84.8 & 84.2 \\
% InternVL3.5-4B & 82.6 & 86 & 92.4 & 78 & 77.9 & 82.2 & -- & 77.1 & 66.6 & -- & 58.1 & 66.3 & -- & -- \\
InternVL3.5-4B& 82.1 & 86.4 & 92.4 & 78 & 77.6 & 82.0 & 76.4 & 72.8 & 57.2 & 76.3 & 58.2 & 67.8 & 82.5 & 47.3 \\
\bottomrule
\end{tabularx}

\end{minipage}
\end{table*}

\section{Evaluation}
\label{sec:evaluation}

We evaluate our model on a set of general vision-language benchmarks as well as two benchmarks focused on grounding tasks.

\subsection{General Benchmarks}
We evaluate ZAYA1-VL-8B on a diverse suite of vision-language benchmarks covering document understanding, perception, reasoning, and counting tasks. Results are summarized in Table~\ref{tab:comprehensive-eval}. 
We refer the reader to Appendix~\ref{app:eval-details} for details on the prompting strategy used for each benchmark along with representative sample responses. Our evaluation suite is designed to probe complementary aspects of vision-language competence. Document understanding is assessed through DocVQA and InfoVQA, which test OCR-free reading and infographic comprehension respectively. For general visual perception and knowledge, we include MMMU, which requires college-level multimodal reasoning, and Blink, which targets fine-grained visual perception that is often overlooked by standard benchmarks. Spatial and object-level understanding is measured via PixMo-Count and CountBenchQA, testing precise object enumeration, while broader reasoning capabilities are captured by benchmarks such as RealWorldQA. This selection ensures coverage across the major axes along which modern VLMs are expected to perform: textual grounding in visual contexts, spatial awareness, factual knowledge, and multi-step reasoning.

Despite using a relatively small number of active parameters, ZAYA1-VL-8B achieves strong performance across a wide range of tasks. In Table~\ref{tab:comprehensive-eval}, we organize models into two broad groups: those employing a Mixture-of-Experts architecture and dense models, with the latter further divided by parameter count (below 4B and 4B+). Across these comparisons, our model is competitive with or surpasses larger models on multiple benchmarks, particularly in diagram/document understanding and counting. Notably, it demonstrates balanced performance across task categories rather than excelling narrowly on a single axis, suggesting that our training mixture and architecture yield robust general-purpose visual understanding rather than benchmark-specific gains.

To better contextualize these results, we plot overall average score against the number of active parameters in Fig.~\ref{fig:scores-vs-params}. ZAYA1-VL-8B achieves competitive accuracy while using considerably fewer active parameters than comparable models, making it a practical choice when inference cost or memory is constrained.
Overall, these results highlight the effectiveness of our training pipeline in achieving strong generalization while maintaining computational efficiency.

\begin{table*}[ht]
\centering
\small  % Match font size of later tables
\setlength{\tabcolsep}{8pt}
\renewcommand{\arraystretch}{1.12}
\begin{tabular}{lcccccc}
\toprule
\textbf{Model} & \textbf{Affordance} & \textbf{Spatial} & \textbf{Reasoning} & \textbf{Steerability} & \textbf{Counting} & \textbf{Average} \\
\midrule
Human & 92.3 & 83.6 & 87.8 & 86.3 & 95.6 & 89.1 \\
\midrule
ZAYA1-VL-8B-A1B & 72.2 & 61.5 & 59.1 & 44.0 & 53.1 & 58.0 \\
MolmoE-8B-A1B & 78.8 & 57.4 & 62.2 & 39.0 & 52.6 & 58.0 \\
\midrule
Qwen3-VL-2B & 71.7 & 60.5 & 53.4 & 25.0 & 57.1 & 53.5 \\
Qwen3.5-2B & 59.6 & 47.7 & 39.4 & 7.0 & 49.5 & 40.6 \\
Qwen2.5-VL-3B & 65.6 & 56.9 & 48.2 & 30.5 & 39.8 & 48.2 \\
\midrule
Molmo2-4B & 85.4 & 76.4 & 76.2 & 40.0 & 64.8 & 68.5 \\
Qwen3-VL-4B & 84.8 & 73.8 & 67.4 & 34.5 & 64.8 & 65.1\\
Qwen3.5-4B & 65.2 & 70.8 & 73.6 & 48.0 & 64.8 & 64.4\\
Molmo-7B-D & 82.3 & 68.2 & 72.0 & 27.5 & 58.7 & 61.7 \\
Molmo-7B-O & 85.4 & 63.1 & 63.2 & 44.5 & 56.6 & 62.6 \\
Qwen2.5-VL-7B & 75.2 & 62.6 & 56.5 & 40.5 & 54.1 & 57.8 \\
\bottomrule
\end{tabular}
\caption{Comparison of ZAYA1-VL-8B on Point-Bench benchmark with various open-source models of comparable scale.}
\label{tab:point-arena}
\end{table*}

\emph{Reproducibility notes.}
Results for all models are reproduced using {VLMEvalKit}~\cite{duan2024vlmevalkit}, ensuring a consistent evaluation pipeline across models. For DocVQA and InfoVQA, we report scores from the original papers as these benchmarks require submission to an external evaluation server.
We observe that some reproduced scores differ from those reported in the original works. For PixMo-Count, the official test set contains 540 examples, of which we were able to retrieve and evaluate 531 images within \texttt{VLMEvalKit}. Under this setting, MolmoE-8B-A1B~\cite{deitke2025molmo} scores notably lower than its reported value (79.6). We do not believe this stems from a systematic issue in our prompting, as the same evaluation setup applied to the closely related Molmo2-4B~\cite{clark2026molmo2} produces results consistent with its published numbers (88.1). Furthermore, scores for the Molmo family (including MolmoE) on Point-Bench (see Table~\ref{tab:point-arena}), which prompts the model using the same XML format, are close to their reported values. Additionally, since MMMU and Blink involve multi-image reasoning and MolmoE-8B-A1B was not trained on multi-image inputs, we omit its scores on these benchmarks.

For the next two evaluations probing grounding capabilities, Point-Bench and RefCOCO, we report scores using PointArena~\cite{pointarena} and \texttt{VLMEvalKit}~\cite{duan2024vlmevalkit} respectively. Models absent from these tables were not trained to generate pointing coordinates or bounding boxes.

\subsection{Grounding Benchmarks 1: Point-Bench}

We first evaluate grounding on PointArena~\cite{cheng2025pointarena}, specifically 
its \emph{Point-Bench} evaluation, which probes language-guided pointing across five 
categories: Affordance, which tests fine-grained tool and object identification 
(\emph{e.g.}, ``point to the object you would use to open the bottle''); Spatial, 
which requires understanding positional relationships between objects; Reasoning, 
which involves complex visual inference to identify a target; Steerability, which 
evaluates the model's ability to follow directional or contextual cues relative to a 
reference point; and Counting, which requires enumerating object instances via 
point-based annotations. Unlike standard VQA-style benchmarks, PointArena requires 
the model to localize its answer as a precise point in the image rather than 
producing a text-only response, making it a stricter test of spatial grounding and 
visual understanding. We provide the prompting strategy used for this evaluation 
along with representative sample responses in Appendix~\ref{app:point-bench} (Figs.~\ref{fig:pointbench-image1} and \ref{fig:pointbench-image2}).

As shown in Table~\ref{tab:point-arena}, ZAYA1-VL-8B achieves an average score 
of 58.0, matching MolmoE-8B-A1B. The two models exhibit complementary strengths: 
ZAYA1-VL-8B performs better on Spatial (61.5 vs.\ 57.4), Steerability (44.0 vs.\ 39.0), 
and slightly on Counting (53.1 vs.\ 52.6), whereas MolmoE is stronger on Affordance 
and marginally better on Reasoning. This pattern suggests that our model is 
particularly effective on tasks requiring fine-grained spatial control and relative 
pointing, likely benefiting from the grounding data introduced during embedding 
expansion and instruction tuning (Sec.~\ref{sec:training-stages}).

Compared with dense baselines at smaller scales, ZAYA1-VL-8B outperforms 
Qwen3-VL-2B, Qwen3.5-2B, and Qwen2.5-VL-3B, and is also slightly ahead of the 
larger Qwen2.5-VL-7B in overall average. At the same time, larger grounding-oriented 
dense models such as Molmo2-4B and Qwen3-VL-4B still maintain a clear 
advantage, indicating that pointing-based grounding continues to benefit from 
additional model capacity and more extensive grounding-focused training data 
(\emph{e.g.}, the large-scale spatio-temporal pointing and tracking datasets 
introduced in~\cite{clark2026molmo2} and the human-labeled video grounding data 
in~\cite{cho2025perceptionlm}).

\subsection{Grounding Benchmarks 2: RefCOCO}

We further evaluate grounding on the RefCOCO~\cite{kazemzadeh2014referitgame}, 
RefCOCO+~\cite{kazemzadeh2014referitgame}, and RefCOCOg~\cite{mao2016generation} 
referring expression comprehension benchmarks, which require the model to localize 
a target object in an image given a natural language description by predicting a 
bounding box. Results are summarized in Table~\ref{tab:refcoco}. All scores are 
reproduced using our evaluation pipeline based on VLMEvalKit~\cite{duan2024vlmevalkit}.

ZAYA1-VL-8B achieves an overall average of 84.3 across all splits, placing it 
competitively among models with significantly more active parameters. On RefCOCO, 
the model scores 91.0 on testA (person-centric queries), approaching larger models 
such as InternVL3.5-4B (94.1) and PLM-8B (93.9). Performance on testB 
(object-centric queries) is somewhat lower at 83.5, reflecting the greater 
difficulty of grounding non-person referents. On RefCOCO+, which removes spatial language cues and 
thus demands stronger appearance-based reasoning, our model achieves 81.8 on val 
and 87.5 on testA, outperforming all 2B--3B dense models and matching Qwen3-VL-2B 
on testA (87.6). The gap to top-performing 4B+ dense models remains modest, 
typically within 3--5 points across splits.

Notably, DeepSeek-VL2-16B-A2.4B scores substantially below its reported numbers 
under our evaluation pipeline. We verified that the coordinate format was correct 
(normalized to a 0--1000 pixel range), and observed that while the output formatting 
was properly parsed, the predicted bounding boxes themselves were consistently 
inaccurate. We report these reproduced scores for consistency but note that the 
discrepancy with the originally published results remains unexplained.

Overall, the two grounding benchmarks results demonstrate that ZAYA1-VL-8B attains strong grounding performance across both pointing and bounding-box formats, despite using significantly fewer active parameters and less training data than the top-performing models in these comparisons. While a substantial gap remains to larger dense models and human performance (89.1 average on Point-Bench), our training recipe --- particularly the staged introduction of grounding data and the use of vision-specific LoRA adapters --- yields a model with competitive spatial understanding rather than one that excels only on high-level recognition or text-heavy benchmarks.

\begin{table*}[t]
\centering
\small
\setlength{\tabcolsep}{4pt}
\renewcommand{\arraystretch}{1.15}

\begin{tabularx}{\linewidth}{l *{8}{>{\centering\arraybackslash}X} c}
\toprule
\multirow{2}{*}{\textbf{Model}} 
& \multicolumn{3}{c}{\textbf{RefCOCO}}
& \multicolumn{3}{c}{\textbf{RefCOCO+}}
& \multicolumn{2}{c}{\textbf{RefCOCOg}}
& \multirow{2}{*}{\textbf{Avg}} \\

\cmidrule(lr){2-4}
\cmidrule(lr){5-7}
\cmidrule(lr){8-9}

& val & testA & testB
& val & testA & testB
& val & test \\

\midrule
ZAYA1-VL-8B-A1B    & 88.0 & 91.0 & 83.5 & 81.8 & 87.5 & 73.6 & 84.5 & 84.5 & 84.3   \\
% \href{}{MolmoE-8B-A1B}      & ?  \\
% DeepSeek-VL2-16B-A2.4B  & 93.9 & 95.3 & 91.3 & 89.4 & 92.9 & 84.8 & 92.6 & 92.6 & 91.6 \\
DeepSeek-VL2-16B-A2.4B  & 45.8 & 52.4 & 40.1 & 38.0 & 46.3 & 30.7 & 42.1 & 42.3 & 42.2 \\
% \href{}{InternVL3.5-20B-A4B}$^\ast$  & 91.9 & 94.1 & 88.8  & 87.6 & 92.0 & 82.7 & 89.1 & 90.0 & 89.5  \\
\href{}{InternVL3.5-20B-A4B}  & 91.5 & 93.8 & 88.3  & 87.3 & 91.6 & 82.0 & 88.9 & 89.2 & 89.1  \\
\midrule
% InternVL3.5-2B$^\ast$ & 88.7 & 91.6 & 84.8 & 82.7 & 88.4 & 76.6 & 85.6 & 85.5 & 85.5\\
InternVL3.5-2B & 86.1 & 89.7 & 82.4 & 80.2 & 86.2 & 74.1 & 82.4 & 82.1 & 82.9\\
Qwen3-VL-2B  & 88.2 & 91.4 & 84.9 & 81.6 & 87.6 & 74.9 & 85.4 & 85.8 & 85.0 \\
Qwen3.5-2B  & 84.4 & 88.0 & 77.4  & 77.2 & 83.4 & 68.9 & 81.0 & 80.8 & 80.1 \\
% PLM-3B & 93.3 & 94.9 & 89.5 & 89.8 & 93.6 & 84.2& 90.8 & 90.9 & 90.9 \\
PLM-3B & 90.4 & 92.2 & 86.7 & 85.3 & 89.2 & 79.8 & 87.6 & 87.5 & 87.3 \\
% Qwen2.5-VL-3B$^\ast$  & 89.1 & 91.7 & 84.0 & 82.4 & 88.0 & 74.1 & 85.2 & 85.7 & 85.0 \\
Qwen2.5-VL-3B  & 85.7 & 88.6 & 80.2 & 77.8 & 83.4 & 70.2 & 80.5 & 81.2 & 81.0 \\
\midrule
Qwen3-VL-4B  & 90.9 & 92.6 & 87.4 & 85.7 & 90.2 & 79.8 & 88.0 & 87.6 & 87.8 \\
Qwen3.5-4B  & 90.6 & 92.7 & 87.6 & 85.2 & 90.2 & 79.7 & 88.0 & 87.7 & 87.7 \\
% InternVL3.5-4B & 92.5 & 94.3 & 88.2 & 87.6 & 92.3 & 81.6 & 89.6 & 89.3 & 89.4 \\
InternVL3.5-4B & 91.8 & 94.1 & 87.8 & 87.1 & 91.5 & 81.1 & 88.6 & 88.7 & 88.8 \\
% Qwen2.5-VL-7B$^\ast$  & 90.0 & 92.5 & 85.4 & 84.2 & 89.1 & 76.9 & 87.2 & 87.2 & 86.6 \\
Qwen2.5-VL-7B  & 90.3 & 92.8 & 85.4 & 84.4 & 89.3 & 76.0 & 86.8 & 87.2 & 86.5 \\
% PLM-8B  & 90.6 & 91.8 & 85.9 & 87.3 & 91.3 & 81.1 & 88.8 & 89.2 & 88.2 \\
PLM-8B  & 91.8 & 93.9 & 86.2 & 87.7 & 92.7 & 80.7 & 88.3 & 89.3 & 88.9 \\
\bottomrule
\end{tabularx}
\caption{Performance on referring expression comprehension benchmarks vs open-source models of comparable and larger parameter scales.}
\label{tab:refcoco}
\end{table*}
\section{Ablation Studies}
\label{sec:ablations}

We conduct a series of ablation experiments to validate the key design choices in 
our architecture and training pipeline. Unless otherwise stated, ablations start from the aligned 
checkpoint (end of the first training stage; see 
Sec.~\ref{sec:training-stages}) and run the pretraining stage on 3 million 
examples (${\sim}$6B total tokens, 240M loss tokens).

We report performance using two aggregate metrics derived from the general 
benchmarks in Table~\ref{tab:comprehensive-eval}: \emph{Und.\ Avg.}, averaging 
over chart, diagram, and document understanding benchmarks (AI2D, ChartQA, DocVQA, 
InfoVQA, TextVQA, and OCRBench), and \emph{Perc.\ Avg.}, averaging over 
perception and reasoning benchmarks (VQAv2.0, MathVista, MMMU, SEED, Blink, and 
RealWorldQA). The first two ablations (resolution and image masking) primarily target improvements on Und.\ Avg., as they manipulate 
vision-side processing and most directly affect benchmarks that probe 
high-resolution image handling. The remaining ablations are expected to influence 
both categories more broadly.

\begin{figure}[htbp]
    \centering
    \includegraphics[width=\linewidth]{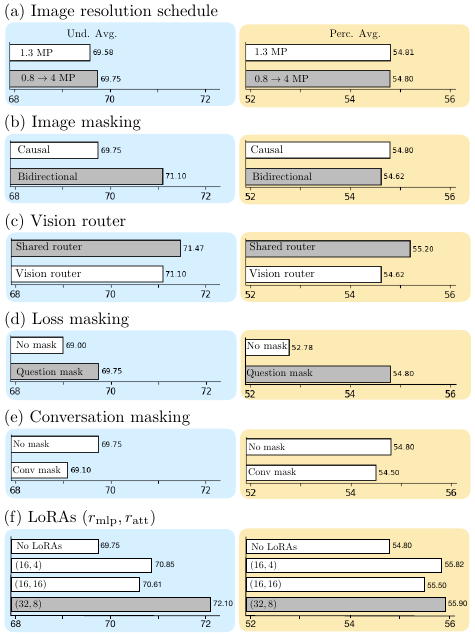}
    \caption{Ablation results. For each experiment we report two average scores: one over image and document understanding benchmarks (Und. Avg.) and one over perception and reasoning benchmarks (Perc. Avg.). The selected configuration is shaded in gray, except for conversation masking, where the mask is applied randomly.}
    \label{fig:ablations}
\end{figure}

\paragraph{Image resolution schedule (Figure~\ref{fig:ablations}(a))} 
We compare training at a fixed lower resolution (1.3MP) against a progressive 
schedule that ramps from 1k to 5k visual tokens via a discrete step function (or equivalently approximately from 0.8MP to 4MP), under a 
comparable compute budget in terms of total training tokens. The progressive 
schedule does not degrade average performance and yields 1--2 point improvements on 
benchmarks that specifically probe high-resolution inputs (ChartQA, DocVQA, 
InfoVQA). Although the aggregate gain in this short ablation is modest, we adopt 
the progressive schedule as it is expected to be increasingly beneficial over 
longer training horizons.

\begin{figure}[htbp]
    \centering
    \includegraphics[width=0.9\linewidth]{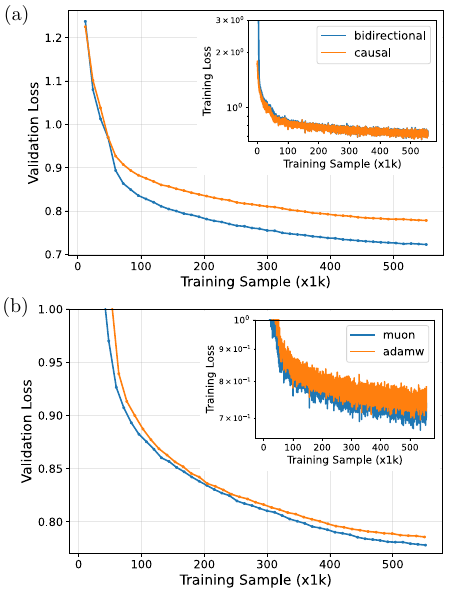}
    \caption{Impact of (a) vision attention mask and (b) vision adapter optimizer 
    on the validation loss during the alignment stage.}
    \label{fig:st1-loss-ablations}
\end{figure}

\paragraph{Image attention masking (Figure~\ref{fig:ablations}(b))} 
By default, the LLM applies causal attention to all tokens, including vision 
tokens. We ablate replacing the causal mask with bidirectional self-attention among 
vision tokens, motivated by the observation that images are a static conditioning 
context without inherent causal structure. As shown in 
Fig.~\ref{fig:st1-loss-ablations}(a), bidirectional masking leads to clearly 
lower validation loss during the alignment stage (where all LLM parameters are 
frozen). In terms of downstream performance, we observe a slight degradation on 
perception and reasoning benchmarks but a consistent benefit for document and image 
understanding tasks, with 1--3 point gains on AI2D, ChartQA, and InfoVQA. 

\paragraph{Vision-specific router (Figure~\ref{fig:ablations}(c))} 
Given the benefit of vision-specific LoRA adapters, a natural hypothesis is that a 
dedicated vision-specific router might similarly help, since the router is a 
relatively small module. We test this by duplicating the MoE router per layer so 
that vision and text tokens are dispatched by separate routers. Although the 
per-modality average expert entropy converged to its maximal value during training 
--- suggesting that both routers learned balanced load distributions --- this did 
not translate into any benchmark improvement. We therefore retain the shared router, 
partly to avoid the additional implementation complexity of splitting tokens by 
modality, routing them through separate modules, and fusing the outputs back into 
the original sequence order. This comparison was conducted using the $(32, 8)$ 
LoRA configuration.

\paragraph{Loss masking (Figure~\ref{fig:ablations}(d))} 
We compare including all tokens in the cross-entropy loss against masking out 
question and context tokens so that loss is computed only over answer tokens. 
Excluding non-answer tokens from the loss is beneficial on both Und.\ Avg.\ and 
Perc.\ Avg., consistent with the intuition that treating images and questions 
purely as conditioning context yields cleaner gradients for the generation 
objective.

\paragraph{Conversation masking (Figure~\ref{fig:ablations}(e))} 
Several datasets in our training mixture contain multiple question--answer pairs 
over a shared set of images, where related questions may reduce the difficulty of 
subsequent ones. We experiment with blocking cross-attention between conversation 
turns (i.e., disabling the shaded regions in 
Fig.~\ref{fig:attention-mask}). Applying conversation masking universally leads 
to a marginal performance reduction. We attribute this to the fact that allowing 
later turns to attend to earlier ones is generally useful for learning multi-turn 
reasoning over shared images. We therefore apply conversation masking 
stochastically at 50\% probability, which preserves the benefits of multi-turn 
context while mitigating information leakage. An exception is made for grounding 
data, where earlier turns can inadvertently reveal answer-relevant information 
(such as object counts or locations expressed in different formats); for these 
examples, we raise the masking probability to 70\%.

\paragraph{Vision-specific LoRA adapters (Figure~\ref{fig:ablations}(f))}
As described in Section~\ref{sec:architecture}, we introduce vision-specific 
capacity by applying LoRA adapters~\cite{hu2022lora} to the linear layers 
processing vision tokens, i.e., $W \to W + B_r A_r$ where $r$ denotes the LoRA 
rank. We additionally apply LoRA adapters to the linear layers in the CCA modules. We use separate ranks for the expert MLP 
and attention modules, denoted $(r_\text{mlp}, r_\text{att})$. As shown in 
Fig.~\ref{fig:ablations}(f), all LoRA configurations improve over the baseline 
without adapters, and the $(32, 8)$ configuration achieves the best overall 
performance across both evaluation categories. We observe that increasing the LoRA rank generally increases performance with the MLP LoRAs being especially important.

\paragraph{Muon optimizer for the vision adapter}
Since the vision adapter is trained from scratch during the alignment stage, we 
compare Muon~\cite{jordan2024muon} and AdamW as its optimizer. As shown in 
Fig.~\ref{fig:st1-loss-ablations}(b), Muon achieves a slightly lower validation loss.

\section{Conclusions}
\label{sec:conclusions}

We presented ZAYA1-VL-8B, a compact mixture-of-experts vision-language model built on our in-house language model ZAYA1-8B~\cite{anthony2025training}. Despite being trained on approximately 140B multimodal tokens, a fraction of the trillions of tokens used by models such as the Qwen-VL family \cite{bai2025qwen3vltechnicalreport}, ZAYA1-VL-8B achieves performance competitive with leading base models such as Molmo2-4B and InternVL3.5-4B, while surpassing Qwen2.5-VL-3B, PLM-3B, and MolMoE-8B across image understanding, reasoning, and counting benchmarks. Two architectural innovations were central to this result: vision-specific LoRA adapters integrated into the LLM to increase modality-specific capacity without adding experts, and bidirectional attention over image tokens to strengthen visual understanding. Equally important was our training infrastructure, which features a fully streaming data pipeline with efficient sequence packing and carefully designed attention masking, enabling high GPU utilization throughout all training stages. Together, these contributions demonstrate that a well-designed MoE backbone paired with a data-efficient pipeline can power a highly capable VLM at a fraction of both the active compute and the data budget of comparable models. By open-sourcing ZAYA1-VL-8B, we aim to provide a practical foundation for the research community to build upon.

Several promising directions emerge from this work. On the architectural side, we plan to incorporate multimodal RoPE directly into the LLM backbone, enabling the model to encode 2D spatial and temporal position information natively rather than relying solely on the vision encoder for positional structure. We expect this to improve spatial reasoning and resolution generalization, particularly for document understanding and fine-grained grounding tasks. On the data side, we intend to significantly scale up our training corpus and extend ZAYA1-VL to video understanding. Recent work \cite{cho2025perceptionlm,clark2026molmo2} has shown that training on spatio-temporal grounding data, including point-driven tracking and video pointing, substantially improves grounding performance even on static images. Incorporating such data into our pipeline is a natural next step toward a more versatile multimodal system.

Finally, the capabilities of any VLM are ultimately bounded by its language backbone. We are actively developing larger and more powerful in-house LLMs that will serve as the foundation for future iterations. Scaling the backbone in tandem with richer multimodal data and improved positional encoding is, in our view, the most direct path toward closing the remaining gap with frontier proprietary systems while preserving the efficiency and openness that define ZAYA1-VL-8B.

\section*{Acknowledgements}

We thank our colleagues at Zyphra; in particular, Xiao Yang for helping with data processing, Rishi Iyer, Anothony Ndirango, Yury Tokpanov, and Robert Washbourne for insightful discussions. We thank Danny Martinelli and Paul White for their help with the ZAYA1-VL public release.

\bibliographystyle{unsrt}
\bibliography{refs.bib}

\clearpage

\onecolumn

\appendices

\section{Dataset details}
\label{app:dataset_details}

In this appendix we provide a detailed description and enumeration of the open-source datasets we used to train the ZAYA1-VL-8B model in each dataset class. 

\paragraph{General}
This category comprises non-document images including real-world scenes, natural photographs, and diverse visual content used for broad visual understanding and instruction tuning. The most notable datasets we sample from are as follows.
MAmmoTH-VL~\cite{guo2024mammothvl} is a 12M instruction-response pair dataset constructed via a scalable pipeline using open-source models, where some fraction contains also chain-of-thought rationales for reasoning-intensive multimodal tasks.
PixMo-Cap~\cite{deitke2025molmo} is a highly detailed image caption dataset collected entirely from human annotators using speech-based descriptions, serving as the pre-training data for the Molmo family of VLMs.
M4-Instruct~\cite{li2024llavainterleavem4} is a 1.18M-sample multi-image instruction dataset spanning across 14 tasks and 41 datasets, compiled for training interleaved multimodal capabilities.
FineVision~\cite{wiedmann2025finevision} is a rigorously curated 24M-sample corpus unifying over 200 open sources via a semi-automated human-in-the-loop pipeline with deduplication and decontamination against 66 benchmarks.
OpenImages~\cite{kuznetsova2020openimages} is a dataset of 9.2M images with unified annotations for image classification, object detection, and visual relationship detection. We use the latter two subsets as part of our grounding datasets.

\paragraph{Document and OCR}
This category focuses on datasets designed for document understanding, optical character recognition, and visually-situated language comprehension from document images, charts, and tables. We consider a mix of the following datasets. The PDF Association (PDFA) dataset~\cite{montalvo2024pdfa} is an extensive OCR dataset containing 2.1 million PDFs with transcriptions, used as a foundation for generating document understanding training data.
The UCSF dataset~\cite{montalvo2024ucsf} is a document dataset derived from the UCSF Industry Documents Library, providing diverse real-world document images for training and evaluation.
DocMatix~\cite{laurencon2024building} is a massive Document Visual Question Answering dataset especially tailored to boost DocVQA tasks.
PixMo-Docs~\cite{deitke2025molmo} is a synthetic dataset within the PixMo collection targeting document understanding capabilities including reading documents and charts in VLMs.
UniChart~\cite{masry2023unichart} contains a large chart corpus for chart comprehension and reasoning, with chart-specific pretraining tasks for low-level element extraction and high-level understanding.
FineVision~\cite{wiedmann2025finevision} also contributes substantial document, OCR, chart, and table reasoning subsets as part of its 24M-sample corpus.
ECD-10K~\cite{yang2025effective} is a synthetic chart dataset of 10K+ images and 300K+ QA pairs spanning 25 topics, designed to improve chart comprehension.

\paragraph{Grounding and Perception}
This category contains datasets for spatial grounding, pointing, counting, and UI/GUI element localization, enabling models to identify and interact with specific regions in images. We use the following datasets as part of our data mixture.
PixMo-Point and PixMo-Count~\cite{deitke2025molmo} are two innovative 2D pointing datasets within the PixMo collection, pairing images with referring expressions and annotated points to support grounding and counting. In the case of counting, the model learns to accurately enumerate objects in images via point-based annotations.
FineVision~\cite{wiedmann2025finevision} further includes grounding and counting subsets as part of its comprehensive multi-task corpus.
MultiUI~\cite{liu2024multiui} is a 7.3M-sample dataset from 1M websites covering diverse multimodal UI tasks which is claimed to achieve strong generalization to non-web domains.
OS-Atlas~\cite{wu2024osatlas} is a cross-platform GUI grounding corpus with over 13M GUI elements spanning Windows, Linux, macOS, Android, and web, for training foundational GUI action models.
UGround~\cite{gou2024uground} provides the largest GUI visual grounding training dataset with 10M elements and referring expressions over 1.3M screenshots, enabling universal visual grounding for GUI agents.
AutoGUI~\cite{li2025autogui} is a pipeline for automatically annotating UI elements with detailed functionality descriptions at scale by leveraging LLMs to infer element functionality from interaction-induced UI changes.
Aria-UI~\cite{yang2024ariaui} provides a diverse set of grounding instructions across platforms which are generated synthetically using a pure-vision approach.
OpenImages~\cite{kuznetsova2020openimages} and Objects365~\cite{shao2019objects365} additionally provide bounding box annotations that support grounding and perception tasks.

\paragraph{Image QA}
This category includes datasets for visual question answering across various domains, combining image understanding with natural language reasoning.
The Cauldron~\cite{laurencon2024matters} is an extensive collection of 50 visual instruction-tuning datasets aggregated for fine-tuning vision-language models targeting academic benchmarks for image, chart, and document understanding. We also separately use several academic datasets tailored to improve performance on the benchmarks including VQA v2.0~\cite{goyal2017making}, OK-VQA~\cite{marino2019ok}, TextVQA~\cite{singh2019towards}, AI2D~\cite{kembhavi2016diagram}, ChartQA~\cite{masry2022chartqa}, DocVQA~\cite{mathew2021docvqa}, InfographicVQA~\cite{mathew2022infographicvqa}, A-OKVQA~\cite{schwenk2022okvqa}, ScienceQA~\cite{lu2022learn}, TabMWP~\cite{lu2022dynamic}, TallyQA~\cite{acharya2019tallyqa}, DVQA~\cite{kafle2018dvqa}, FigureQA~\cite{kahou2017figureqa}, and PlotQA~\cite{methani2020plotqa}.
ArxivQA~\cite{li2024multimodalarxiv} is a question-answering dataset generated by prompting GPT-4V on scientific figures from ArXiv papers, greatly enhancing mathematical reasoning.
Molmo2-MultiImage~\cite{clark2026molmo2} refers to the multi-image extension data from the PixMo collection~\cite{deitke2025molmo}, enabling multi-image reasoning capabilities in VLMs.
M4-Instruct~\cite{li2024llavainterleavem4} also provides multi-image QA capabilities through its interleaved data format.
UReader~\cite{ye2023ureader} is a finetuning dataset based on several academic datasets across documents, tables, charts, natural images, and webpages via a unified instruction format.
SynCLock~\cite{yang2022its} is a code base to synthetically generate clock faces which we use for training to accurately read analog clocks.

\paragraph{Multimodal Reasoning}
This category encompasses datasets that require complex multi-step reasoning integrating both visual and textual modalities, often with chain-of-thought annotations.
M$^3$CoT~\cite{chen2024m3cot} is a benchmark for multi-domain, multi-step, multi-modal chain-of-thought reasoning with 11.4K samples spanning science, math, and commonsense domains.
Geometry3K~\cite{lu2021intergps} is a dataset of 3,002 geometry problems with formal language annotations, enriching geometric problem types across diverse shapes and variable operators for multimodal numerical reasoning.
Geo170K~\cite{gao2023gllava} is a large-scale geometric visual-text dataset comprising approximately 170K question-answer pairs synthesized from existing datasets using LLMs, significantly surpassing prior geometry datasets in scale.
ViRL39K~\cite{wang2025vl} is a curated collection of 39K visual question-answer pairs spanning math, physics, chemistry, biology, chart and diagram reasoning, and broader STEM and social science topics, designed as a reinforcement learning training set for incentivizing self-reflection in vision-language models.

\paragraph{Text}
This category contains text-only datasets (without paired images) used for enhancing language understanding, mathematical reasoning, and code generation capabilities in multimodal model training.
GSM8K~\cite{cobbe2021gsm8k} is a dataset of 8.5K high-quality, linguistically diverse grade school math word problems requiring 2--8 steps to solve, designed to support multi-step mathematical reasoning.
SmolTalk2-SFT is a curated supervised fine-tuning dataset from HuggingFace for training small language models with diverse conversational and instruction-following capabilities.
Furthermore, we sample from NVIDIA AceMath~\cite{liu2024acemath} which provides a suite of frontier math reasoning models and associated SFT data using carefully curated prompts and synthetically generated responses for targeted mathematical domain fine-tuning.

\section{Grounding example formats}

\label{app:grounding}
\setcounter{subsection}{0}
\renewcommand{\thesubsection}{\thesection.\arabic{subsection}}
\renewcommand{\thesubsectiondis}{\arabic{subsection}.}
\renewcommand{\theHsubsection}{appendix.\thesection.\arabic{subsection}}

In this appendix, we provide explicit examples from our training data to illustrate the chat template formatting and grounding conventions used for pointing and bounding box tasks. These examples show how spatial coordinates are represented within the model's input and output sequences, covering both point annotations and rectangular bounding box specifications.

\subsection{Pointing}

Figures~\ref{fig:pixmopoint-images1} and \ref{fig:pixmopoint-images2} show pointing examples with a single image. We use two coordinate formats. The first is an XML format: \texttt{<points x1="." y1="." x2="." y2="." ... alt="obj\_name">obj\_name</points>}, where coordinates are normalized to the range $[0, 100]$ with one decimal place. The second is a format we introduce using special tokens: \texttt{<point\_start>(x1, y1)<point\_end>} where coordinates are integers varying between 0 and 1000. They can further be composed into various structured representations as specified in the prompt, including JSON, Python lists, dictionaries, and markdown tables. When we refer to objects with coordinates this format is wrapped in \texttt{<|object\_ref\_start|>} and \texttt{<|object\_ref\_end|>} as illustrated in Fig.~\ref{fig:pixmopoint-images2}.

For multi-image pointing, shown in Figs.~\ref{fig:multiimage-point-images1} and \ref{fig:multiimage-point-images2}, coordinates are provided after each image, with images enumerated as \texttt{image\_xx} where \texttt{xx} ranges from 1 to the total number of images.

\begin{figure*}[htbp]
    \centering
    \includegraphics[width=0.84\linewidth]{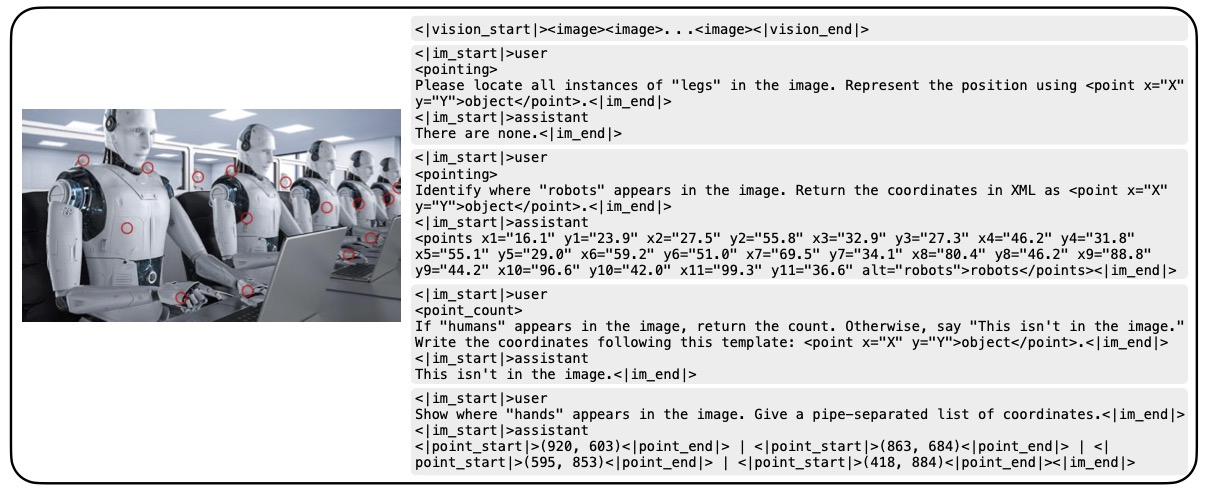}
    \caption{Random examples from PixMo-point.}
    \label{fig:pixmopoint-images1}
\end{figure*}

\begin{figure*}[htbp]
    \centering
    \includegraphics[width=0.84\linewidth]{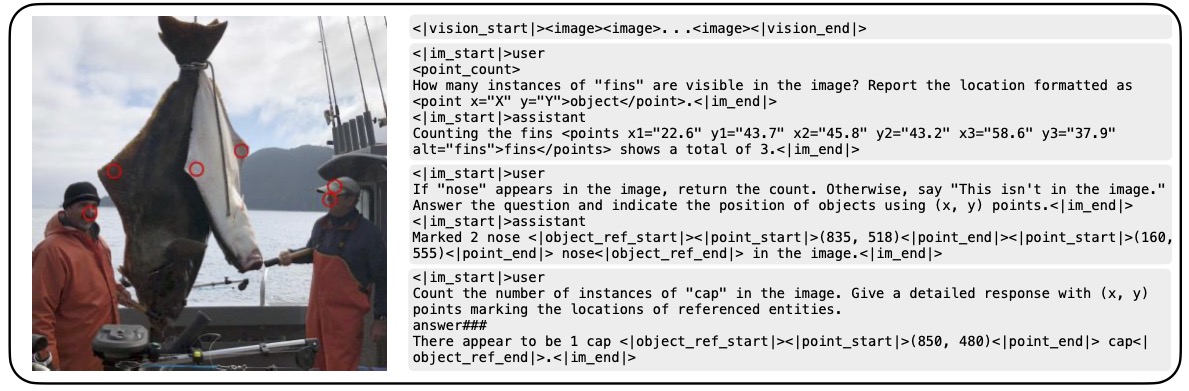}
    \includegraphics[width=0.84\linewidth]{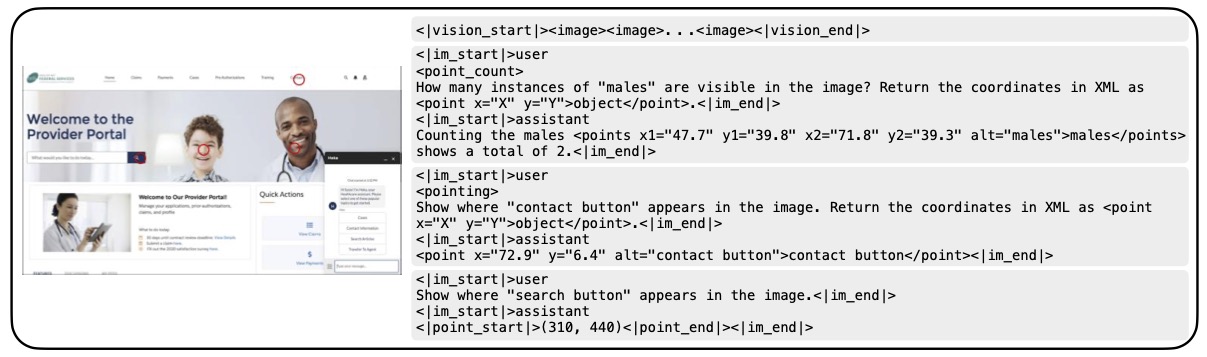}
    \includegraphics[width=0.84\linewidth]{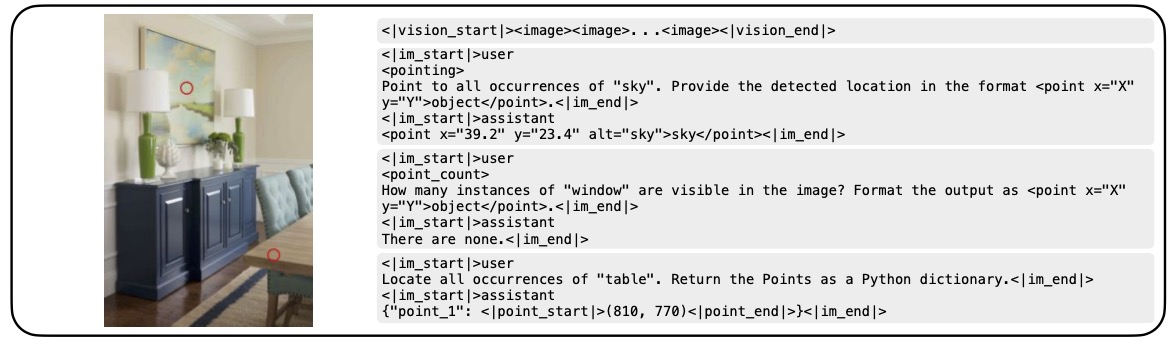}
    \includegraphics[width=0.84\linewidth]{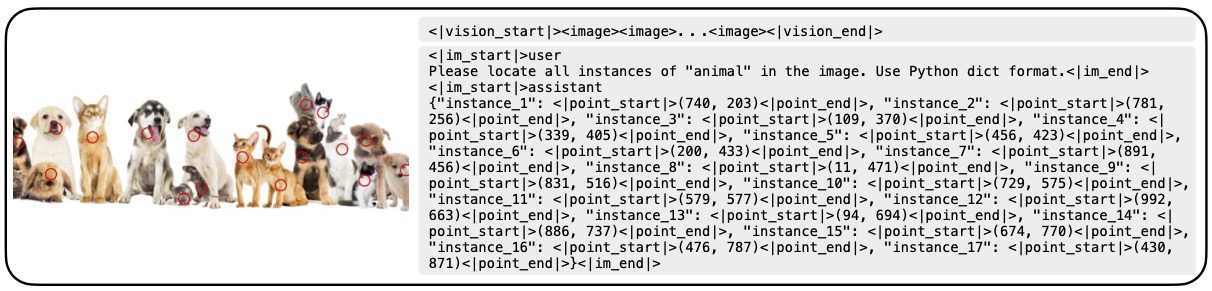}
    \caption{(Cont.) Random examples from PixMo-point.}
    \label{fig:pixmopoint-images2}
\end{figure*}

\begin{figure*}[h]
    \centering
    \includegraphics[width=0.84\linewidth]{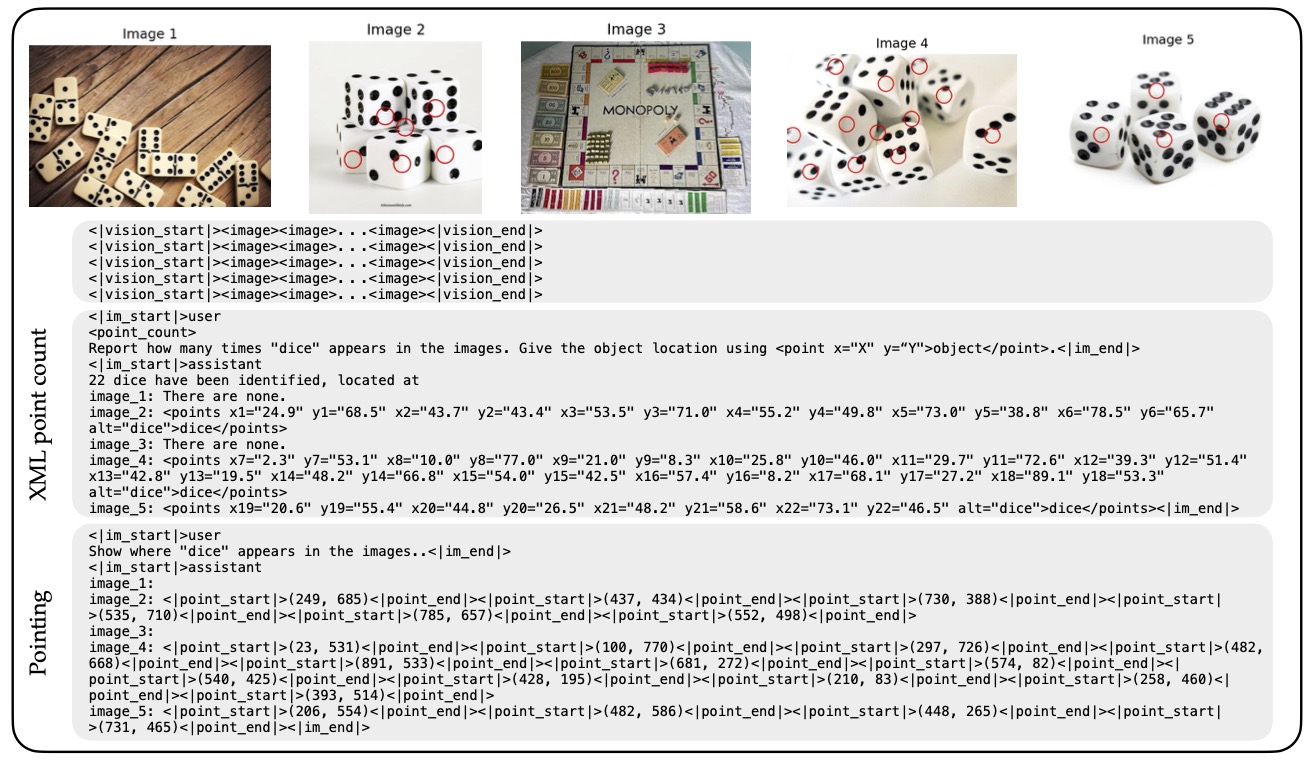}
    \includegraphics[width=0.84\linewidth]{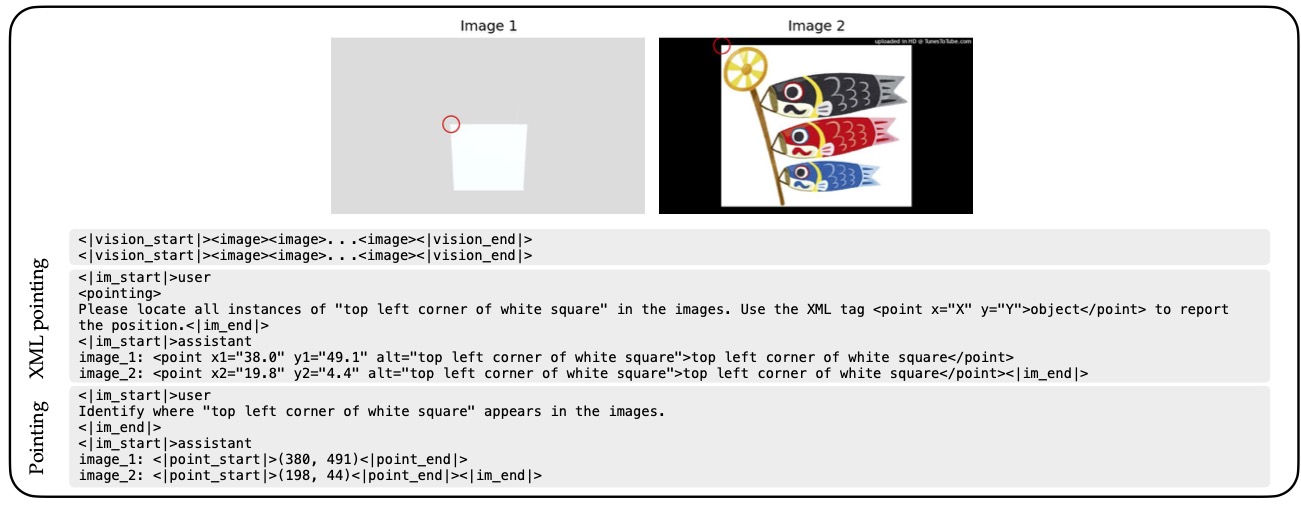}
    \includegraphics[width=0.84\linewidth]{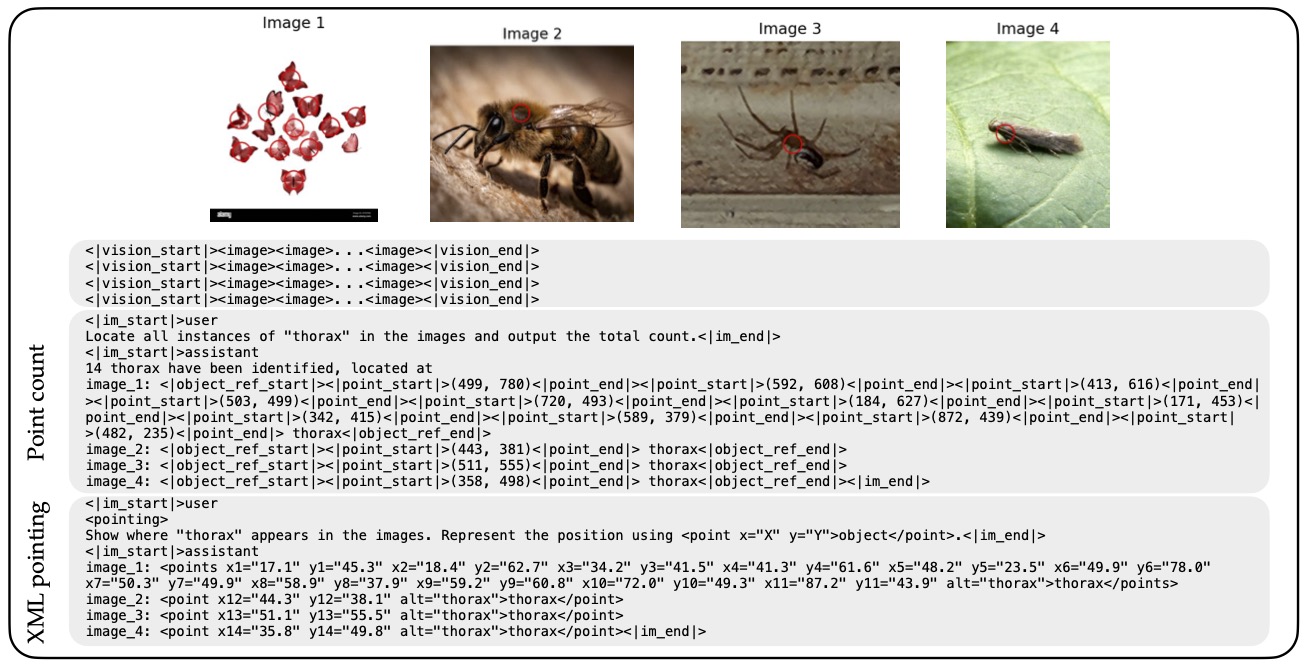}
    \caption{Random examples from Molmo2-multi-image-pointing.}
    \label{fig:multiimage-point-images1}
\end{figure*}

\begin{figure*}[htbp]
    \centering
    \includegraphics[width=0.83\linewidth]{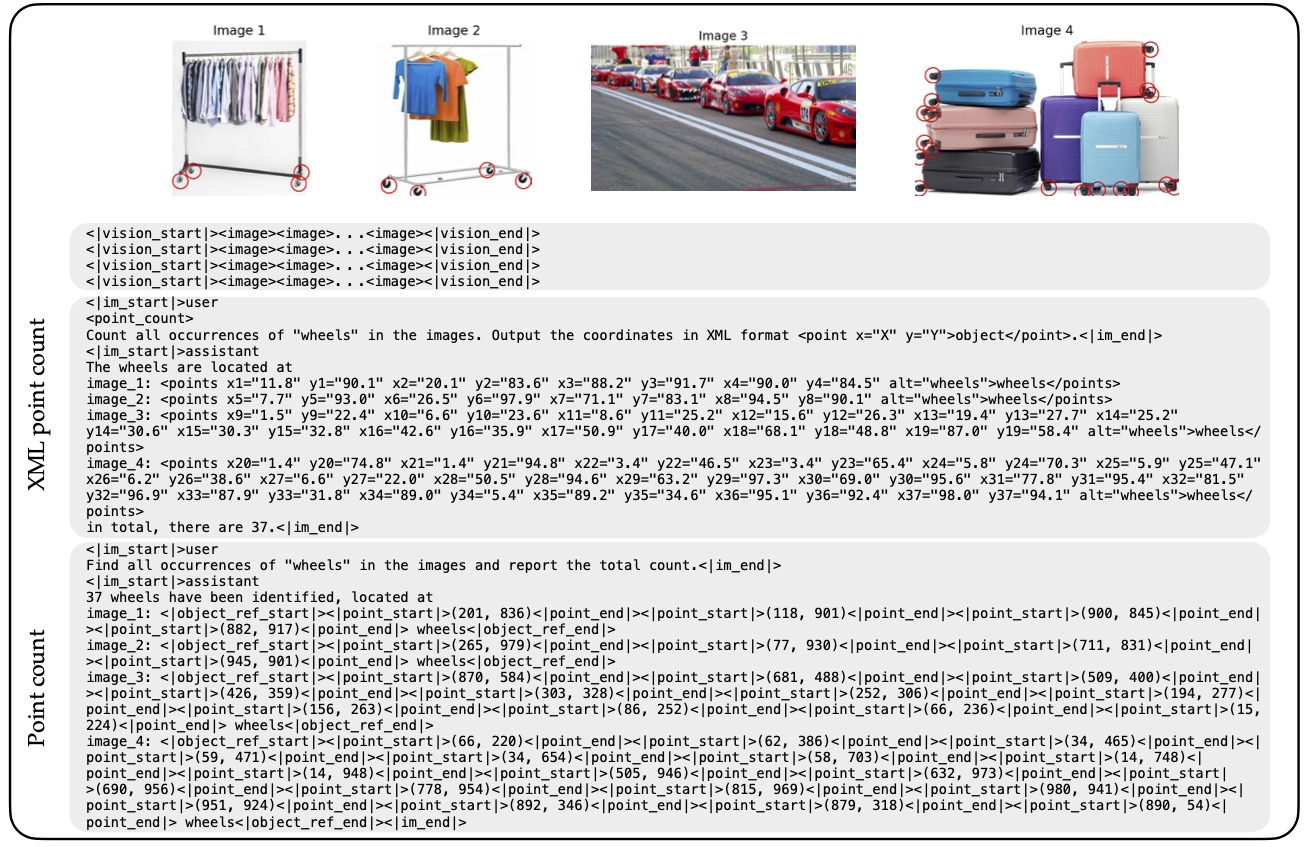}
    \caption{(Cont.) Random examples from Molmo2-multi-image-pointing.}
    \label{fig:multiimage-point-images2}
\end{figure*}

\begin{figure*}[h]
    \centering
    \includegraphics[width=0.83\linewidth]{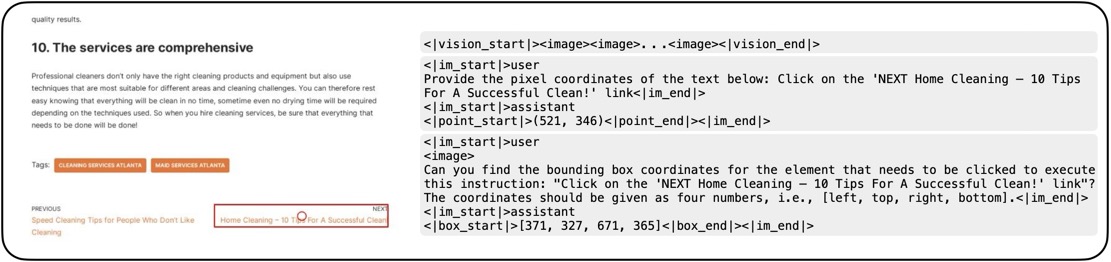}
    \includegraphics[width=0.83\linewidth]{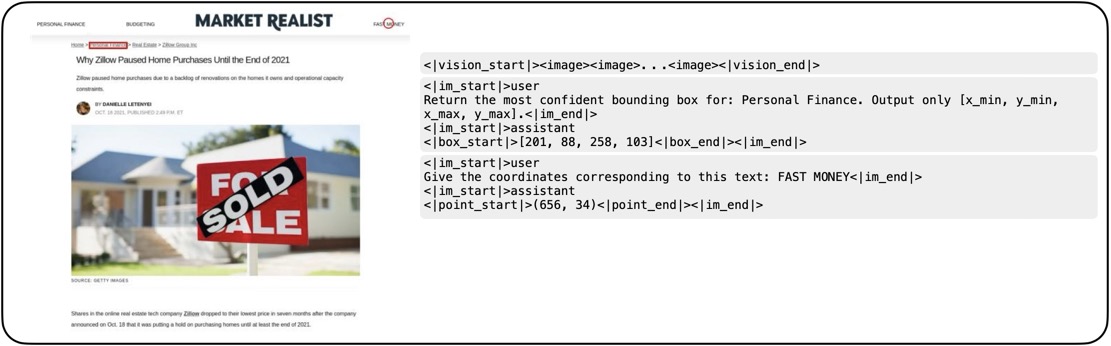}
    \includegraphics[width=0.83\linewidth]{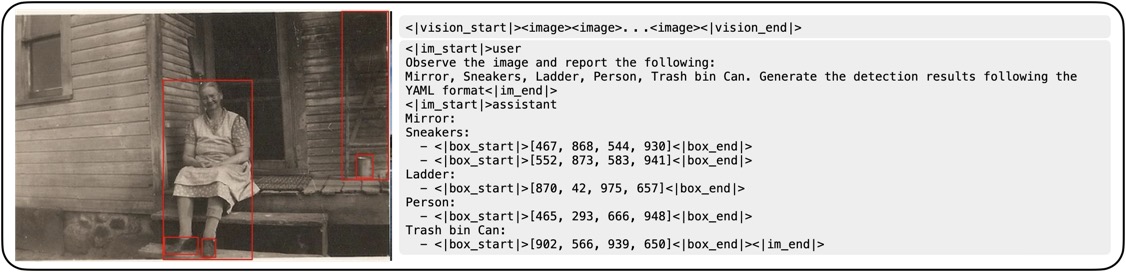}
    \caption{Random examples from grounding datasets.}
    \label{fig:grounding-images1}
\end{figure*}

\clearpage
\newpage

\subsection{Bounding box}

For bounding boxes, as shown in Figures~\ref{fig:grounding-images1}--\ref{fig:grounding-images4}, we use the format \texttt{<|box\_start|>[x1, y1, x2, y2]<|box\_end|>}, where coordinates are integers in the range $[0, 1000]$ representing a relative coordinate system. As with pointing, this format can be composed into structured representations such as JSON. Bounding boxes can additionally be wrapped in \texttt{<|object\_ref\_start|>} and \texttt{<|object\_ref\_end|>} tokens when referring to specific objects.

\begin{figure*}[htbp]
    \centering
    \includegraphics[width=0.84\linewidth]{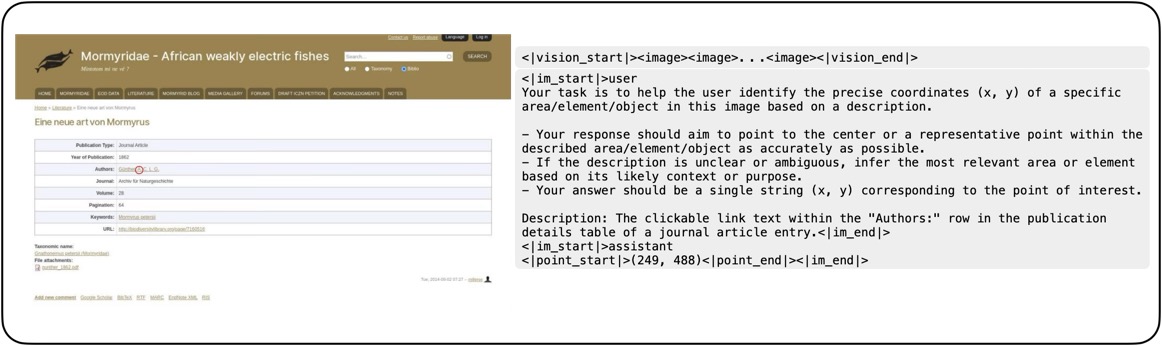}
    \includegraphics[width=0.84\linewidth]{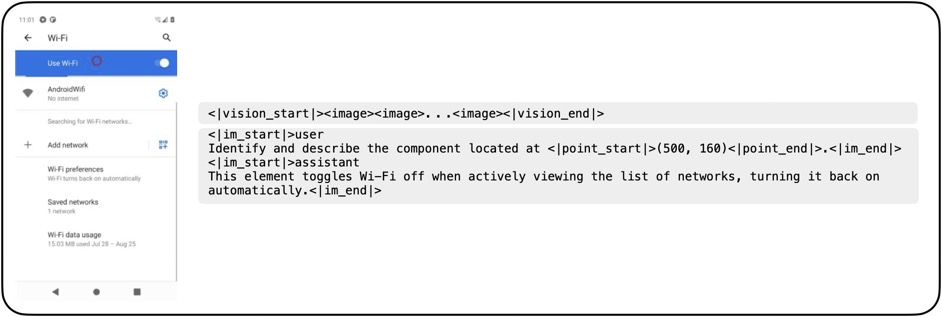}
    \includegraphics[width=0.84\linewidth]{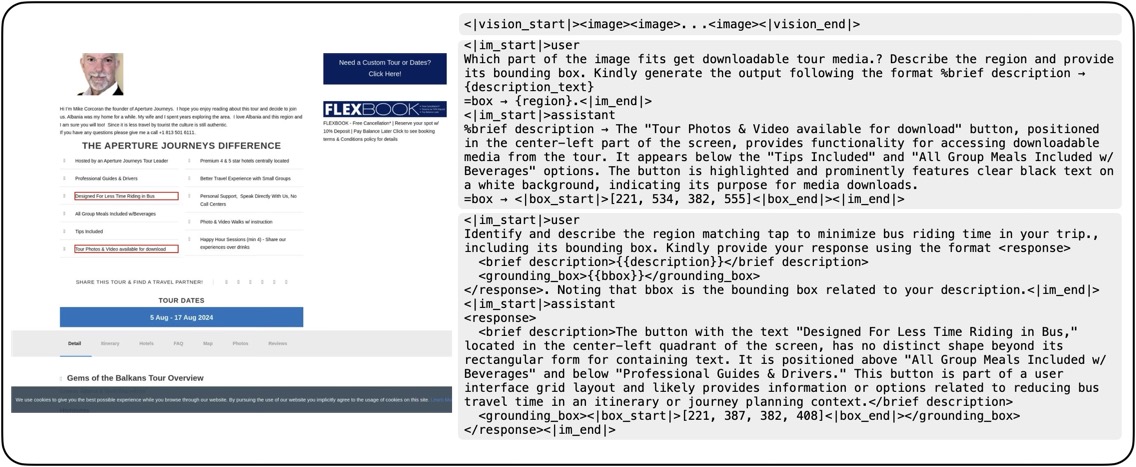}
    \includegraphics[width=0.84\linewidth]{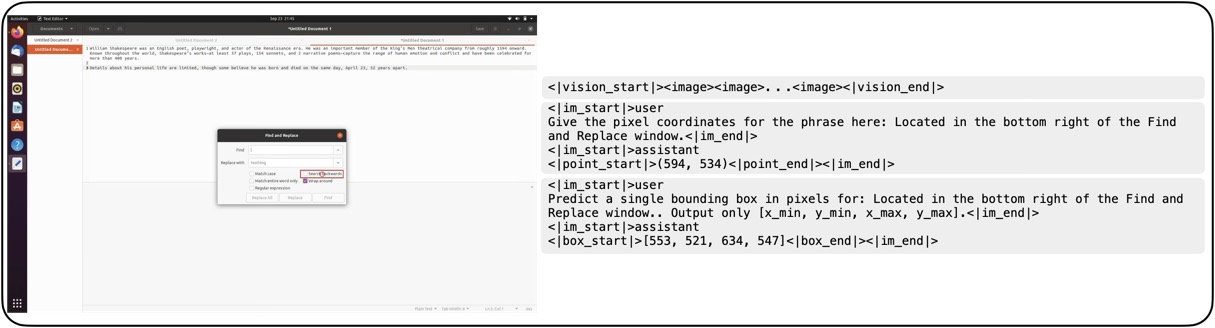}
    \caption{(Cont.) Random examples from grounding datasets.}
    \label{fig:grounding-images2}
\end{figure*}

\begin{figure*}[htbp]
    \centering
    \includegraphics[width=0.84\linewidth]{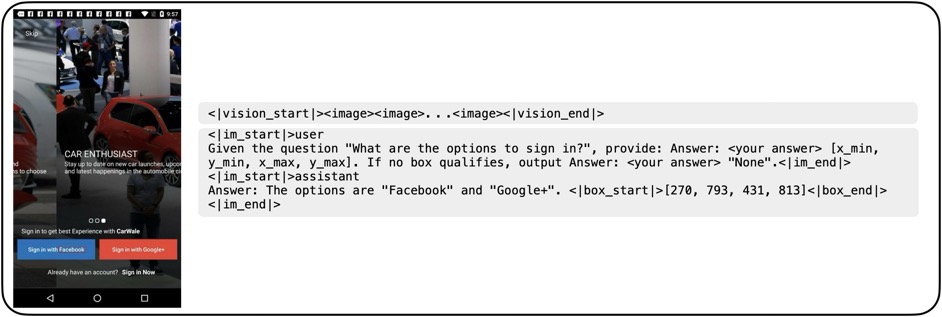}
    \includegraphics[width=0.84\linewidth]{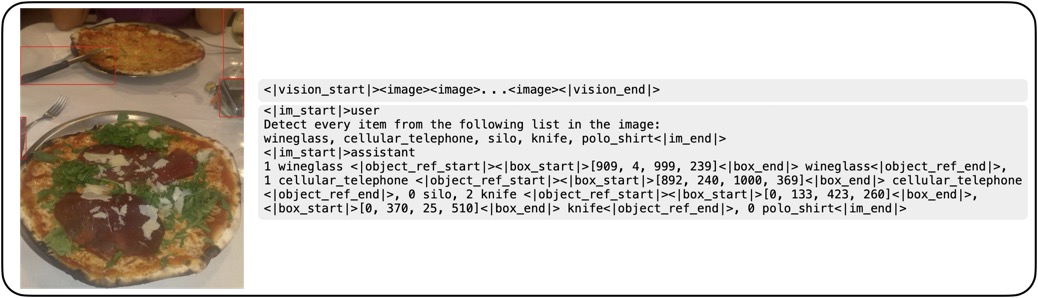}
    \includegraphics[width=0.84\linewidth]{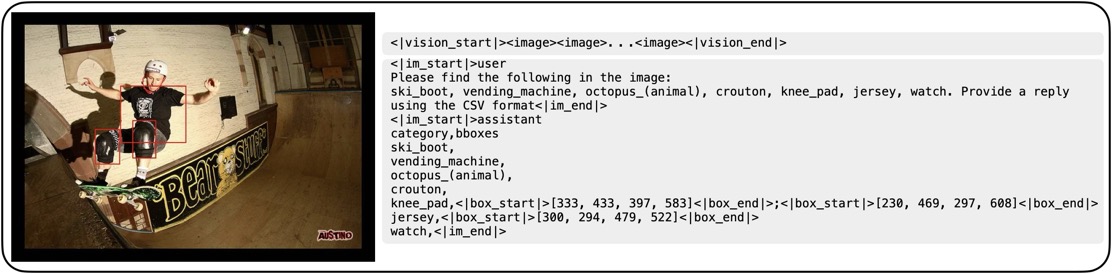}
    \includegraphics[width=0.84\linewidth]{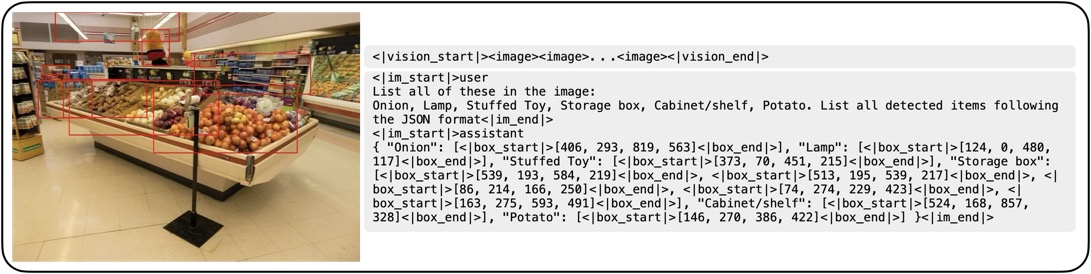}
    \includegraphics[width=0.84\linewidth]{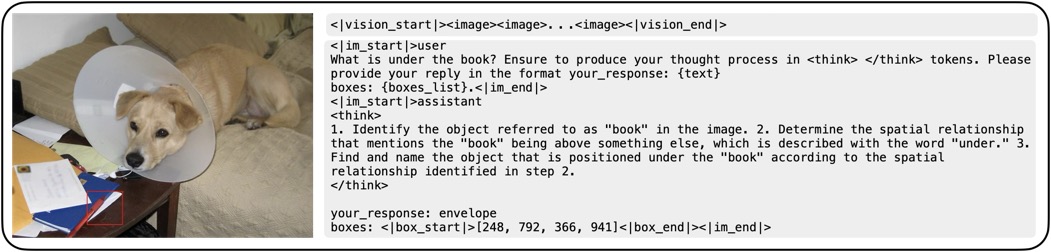}
    \caption{(Cont.) Random examples from grounding datasets.}
    \label{fig:grounding-images3}
\end{figure*}

\begin{figure*}[htbp]
    \centering
    \includegraphics[width=0.84\linewidth]{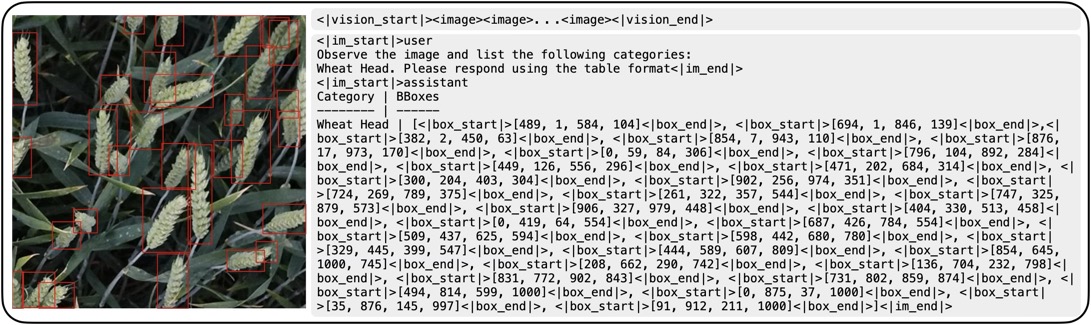}
    \includegraphics[width=0.84\linewidth]{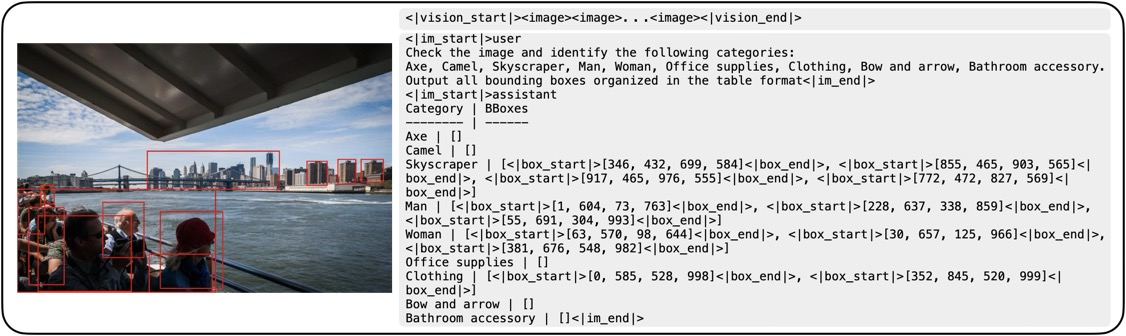}
    \includegraphics[width=0.84\linewidth]{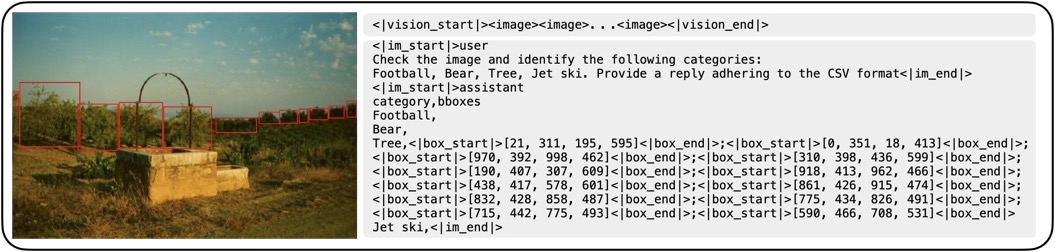}
    \includegraphics[width=0.84\linewidth]{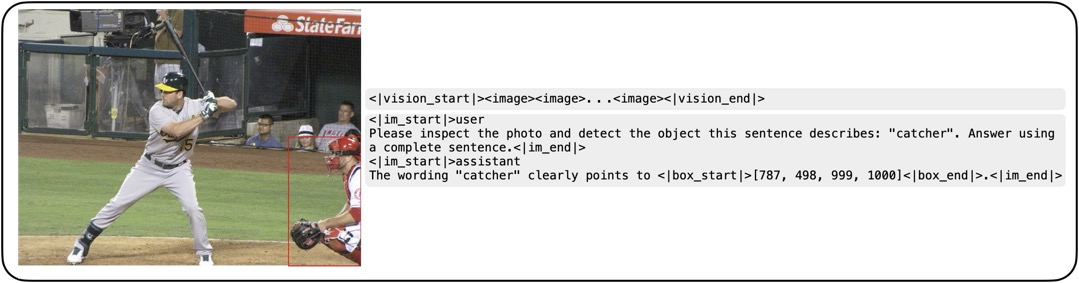}
    \includegraphics[width=0.84\linewidth]{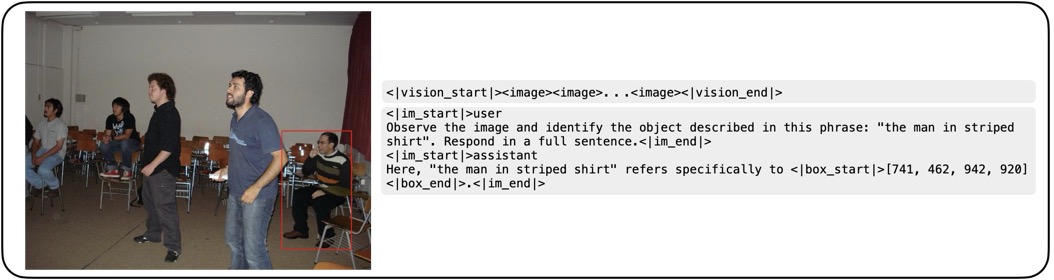}
    \caption{(Cont.) Random examples from grounding datasets.}
    \label{fig:grounding-images4}
\end{figure*}

\clearpage
\newpage

\section{Evaluation details and examples}
\label{app:eval-details}

\setcounter{subsection}{0}
\renewcommand{\thesubsection}{\thesection.\arabic{subsection}}
\renewcommand{\thesubsectiondis}{\arabic{subsection}.}
\renewcommand{\theHsubsection}{appendix.\thesection.\arabic{subsection}}

In this appendix, we provide examples of model responses to various evaluation benchmarks summarized in Tables~\ref{tab:comprehensive-eval}, \ref{tab:point-arena}, and \ref{tab:refcoco}. We highlight two aspects in particular: the prompt format used to query the model (shown in gray boxes) and the format in which the model returns its response (shown in blue boxes). For certain academic benchmarks (AI2D, ChartQA, DocVQA, InfoVQA, TextVQA, VQA v2.0, and counting), we adopt a tagging scheme similar to Molmo~\cite{deitke2025molmo} both during training on the respective train splits and at evaluation time, as shown explicitly in the figures.

\subsection{Chart, Diagram, and Document Understanding}
\begin{figure*}[htbp]
    \centering
    \includegraphics[width=0.75\linewidth]{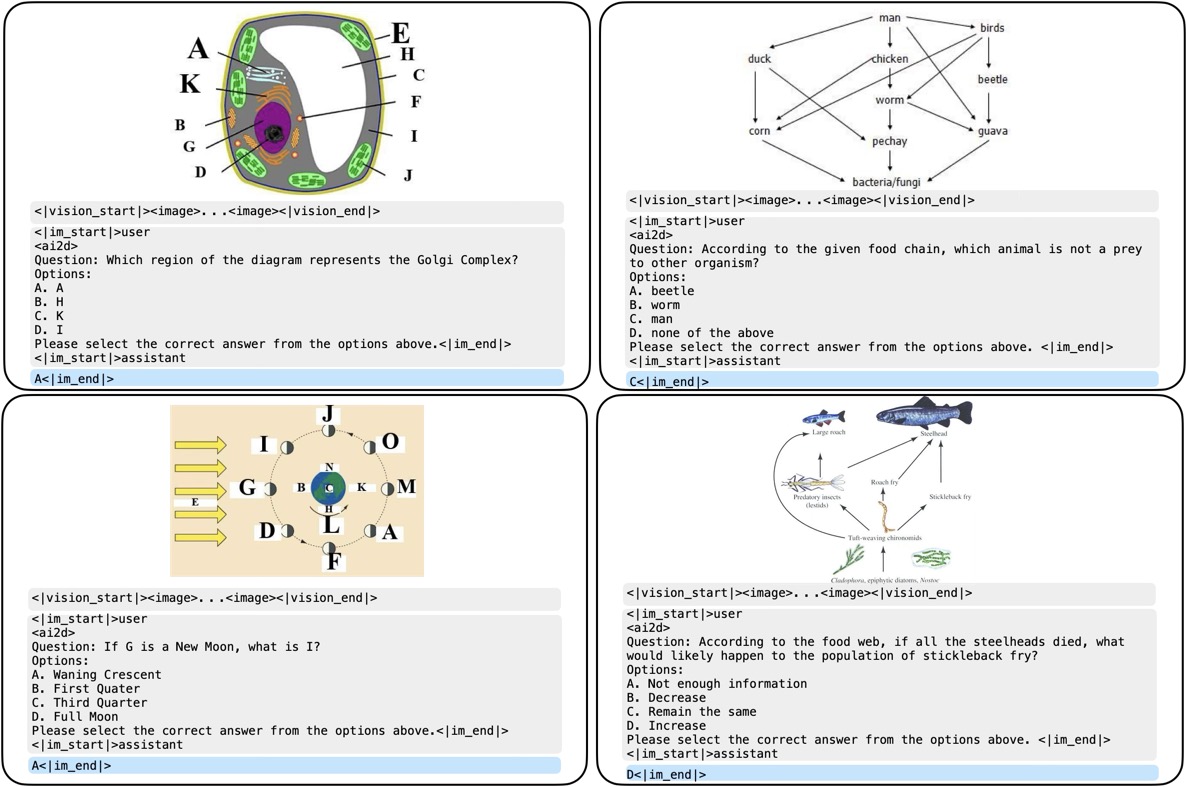}
    \caption{Random examples from AI2D benchmark and model response.}
    \label{fig:ai2d-images}
\end{figure*}

\begin{figure*}[htbp]
    \centering
    \includegraphics[width=0.75\linewidth]{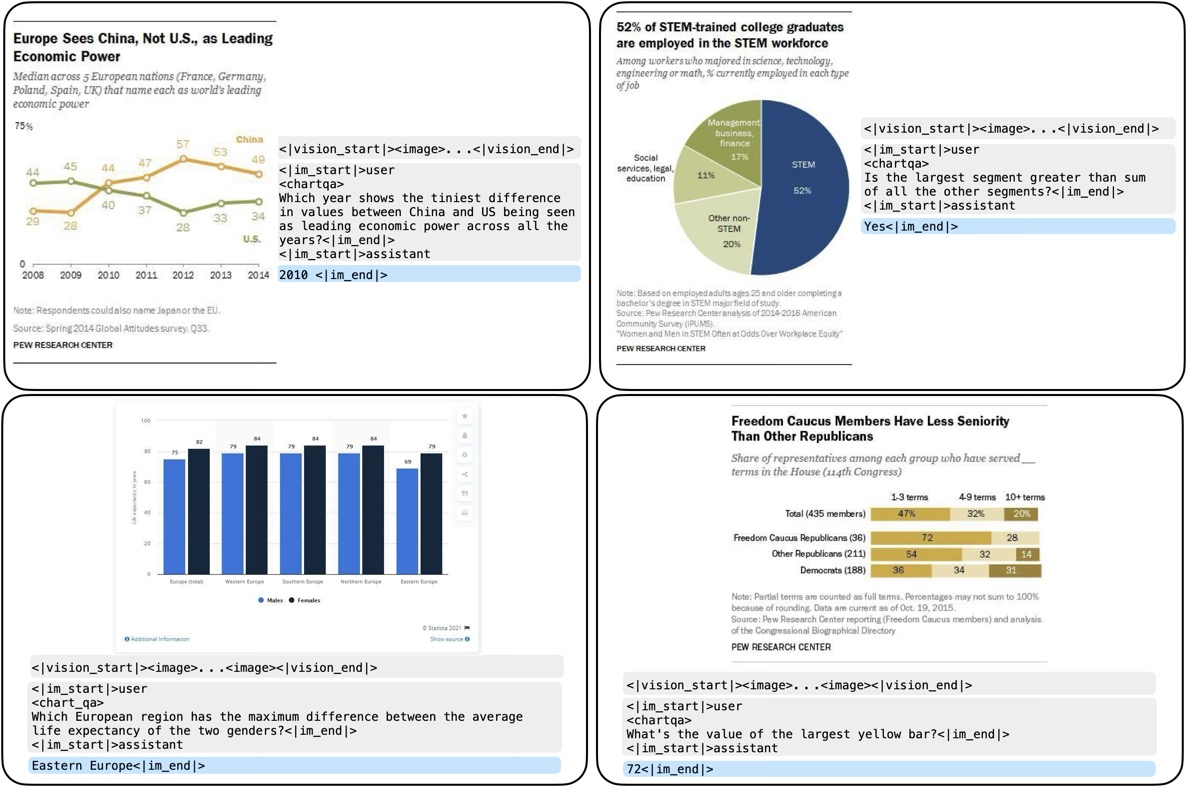}
    \caption{Random examples from ChartQA(test) benchmark and model response.}
    \label{fig:chartqa-images}
\end{figure*}

\begin{figure*}[htbp]
    \centering
    \includegraphics[width=0.8\linewidth]{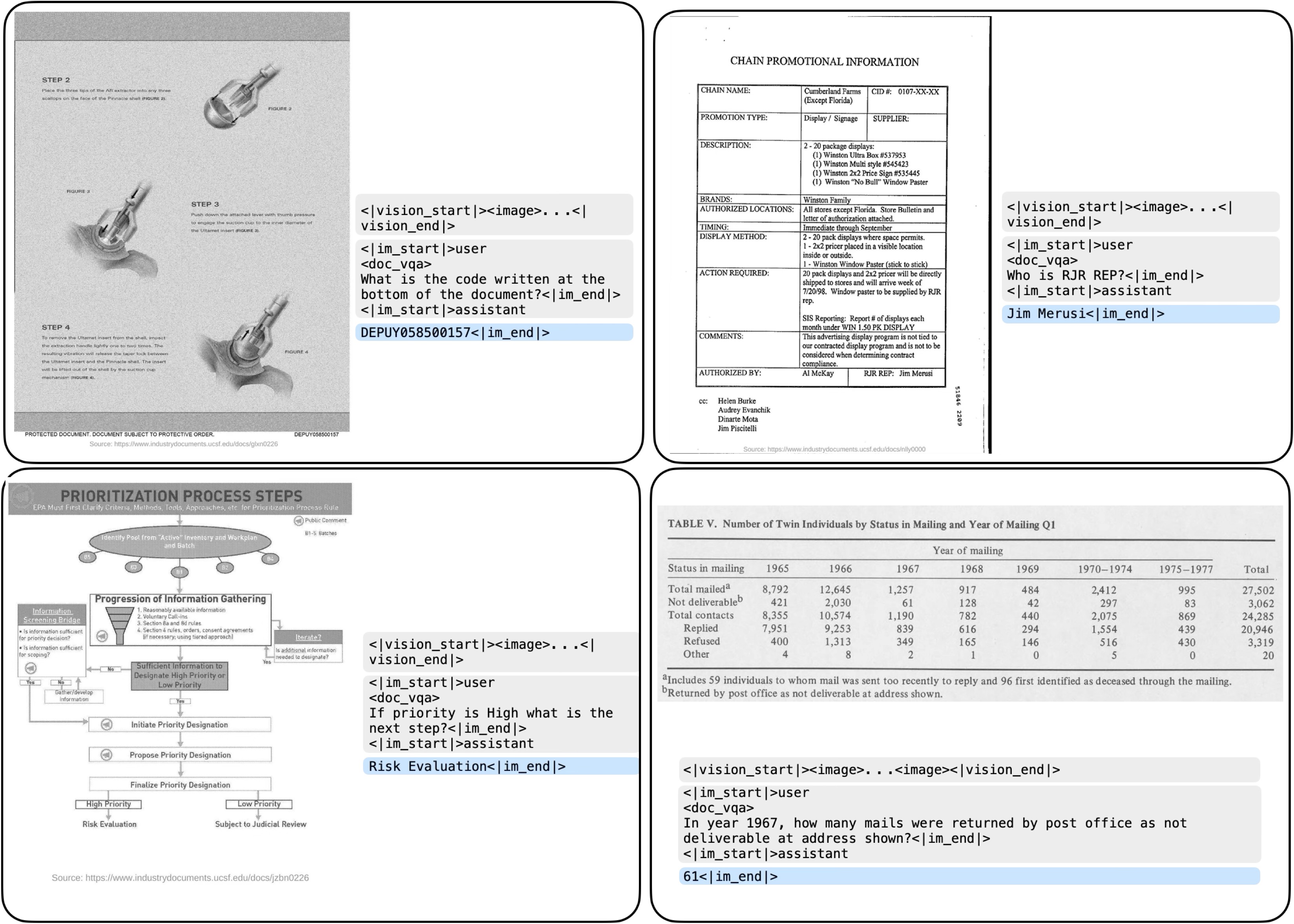}
    \caption{Random examples from DocVQA(val) benchmark and model response.}
    \label{fig:docvqa-images}
\end{figure*}

\begin{figure*}[htbp]
    \centering
    \includegraphics[width=0.8\linewidth]{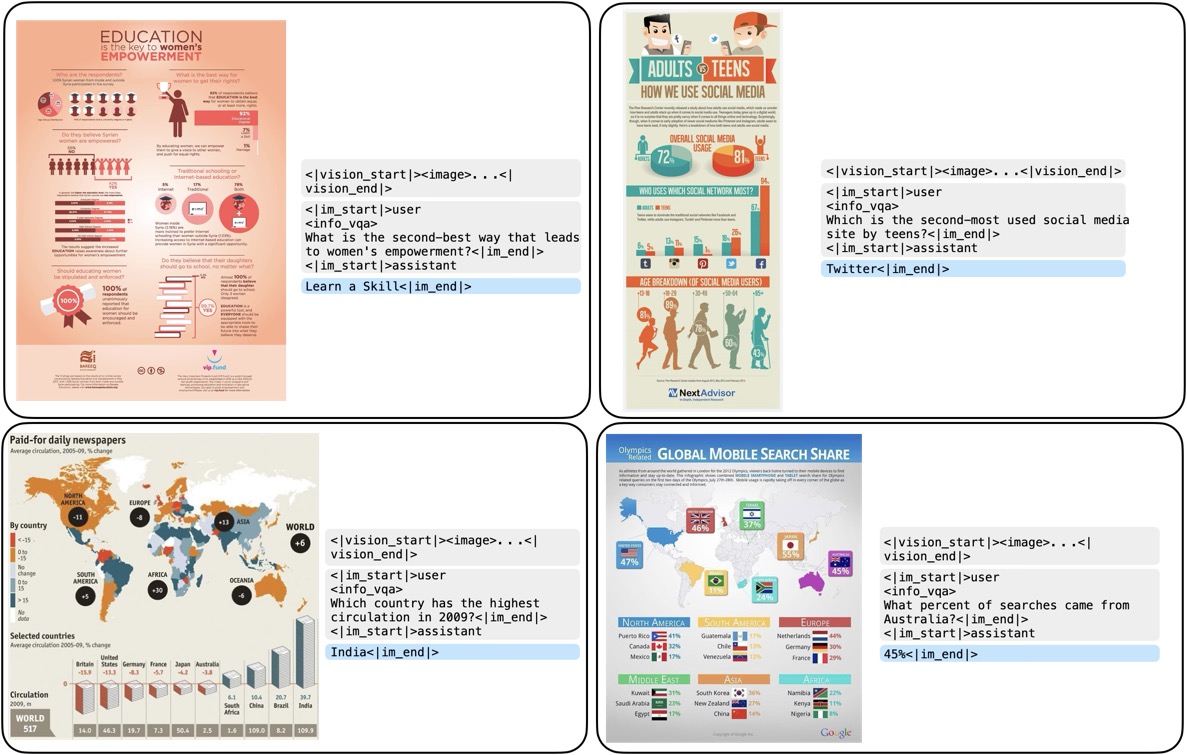}
    \caption{Random examples from InfoVQA(val) benchmark and model response.}
    \label{fig:infovqa-images}
\end{figure*}

\begin{figure*}[htbp]
    \centering
    \includegraphics[width=0.8\linewidth]{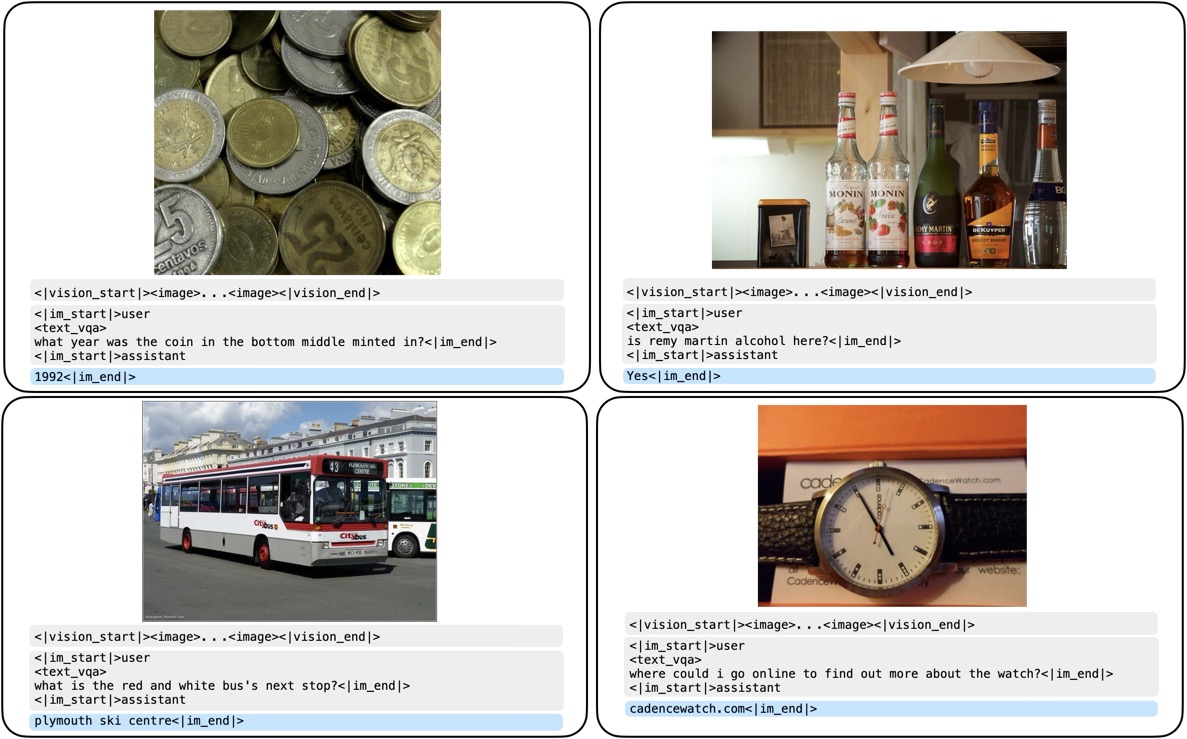}
    \caption{Random examples from TextVQA(val) benchmark and model response.}
    \label{fig:textvqa-images}
\end{figure*}

\begin{figure*}[htbp]
    \centering
    \includegraphics[width=0.8\linewidth]{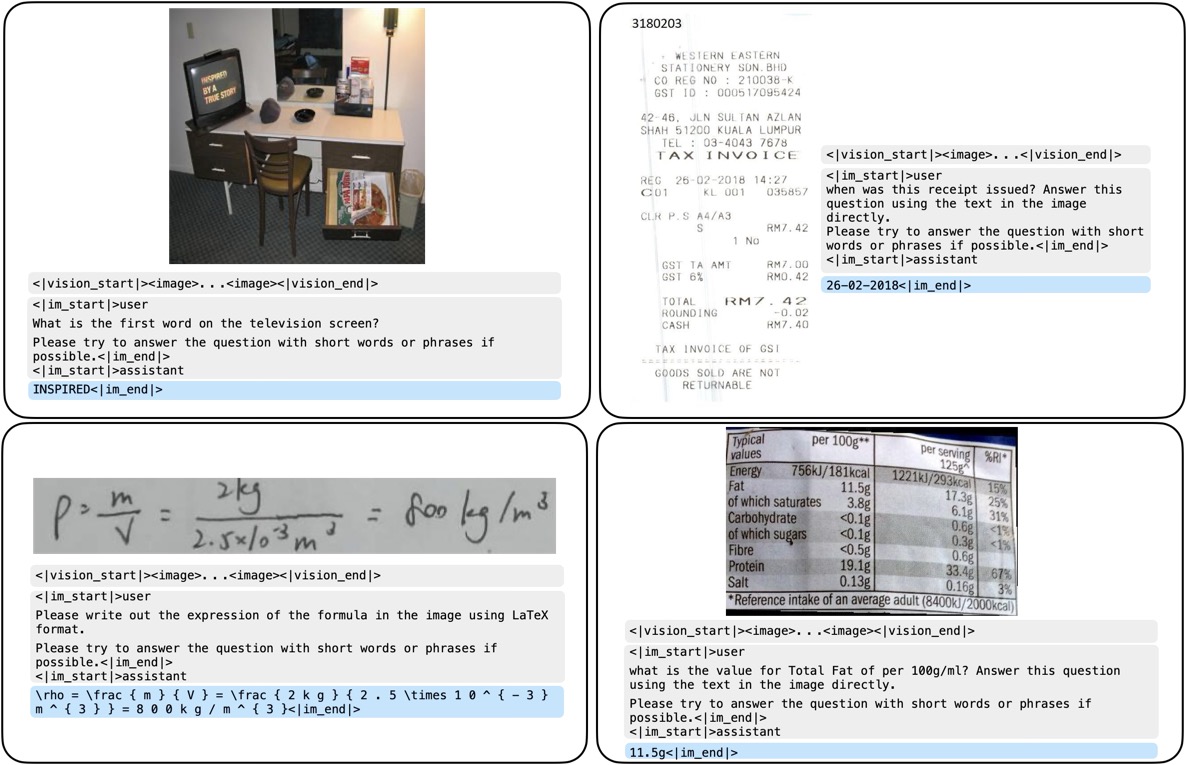}
    \caption{Random examples from OCRBench benchmark and model response.}
    \label{fig:ocrbench-images}
\end{figure*}

\clearpage
\newpage

\subsection{Perception and reasoning}

\begin{figure*}[htbp]
    \centering
    \includegraphics[width=0.84\linewidth]{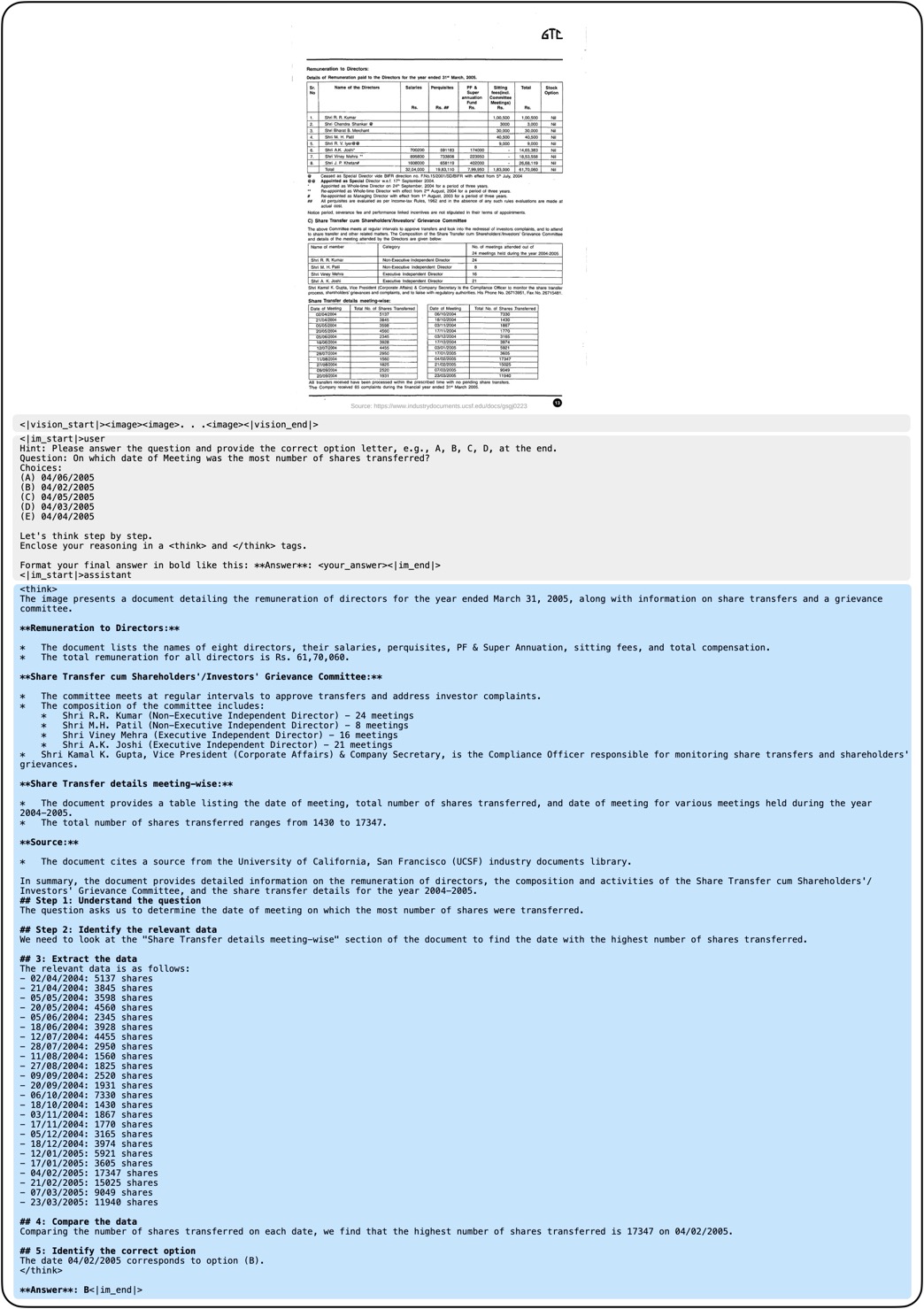}
    \caption{Random examples from MathVista-Mini benchmark and model response.}
    \label{fig:mathvista-image2}
\end{figure*}

\begin{figure*}[htbp]
    \centering
    \includegraphics[width=0.84\linewidth]{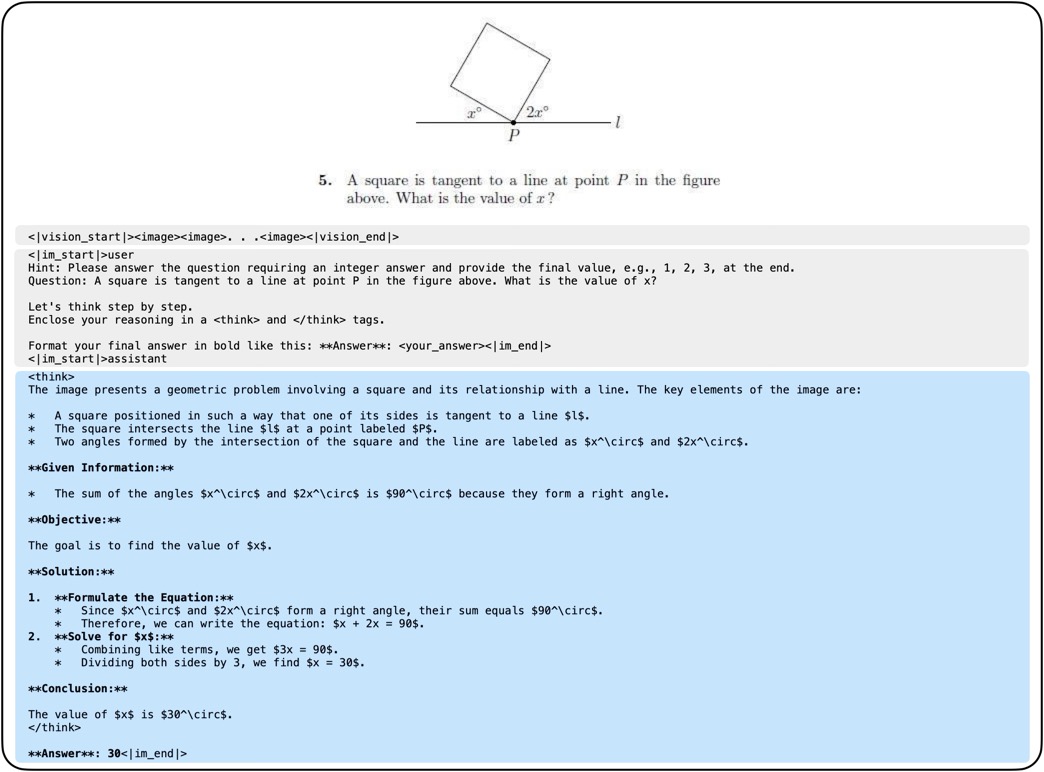}
    \caption{Random examples from MathVista-Mini benchmark and model response.}
    \label{fig:mathvista-image4}
\end{figure*}

\begin{figure*}[htbp]
    \centering
    \includegraphics[width=0.84\linewidth]{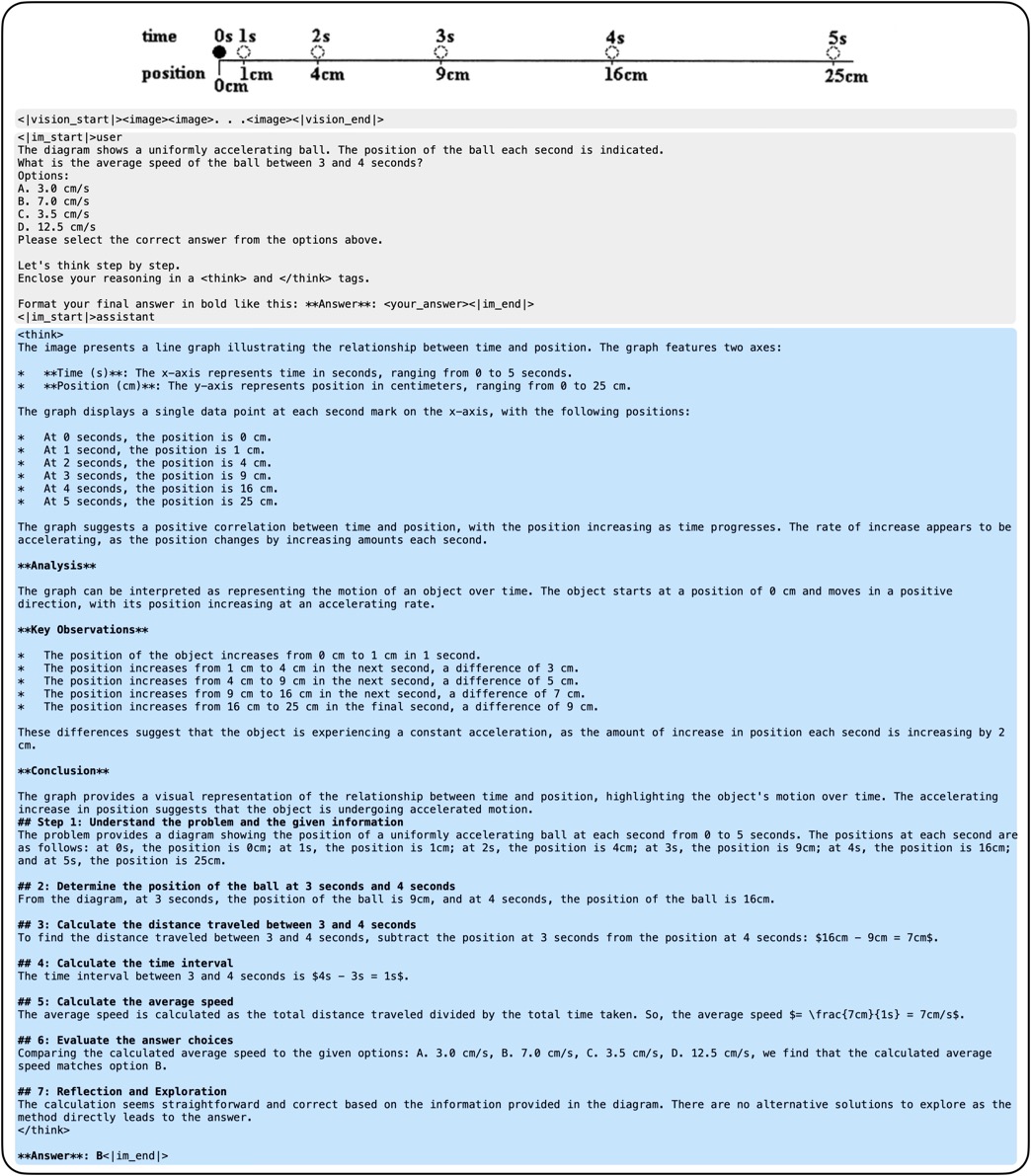}
    \caption{Random examples from MMMU benchmark and model response.}
    \label{fig:mmmu-image1}
\end{figure*}

\begin{figure*}[htbp]
    \centering
    \includegraphics[width=0.84\linewidth]{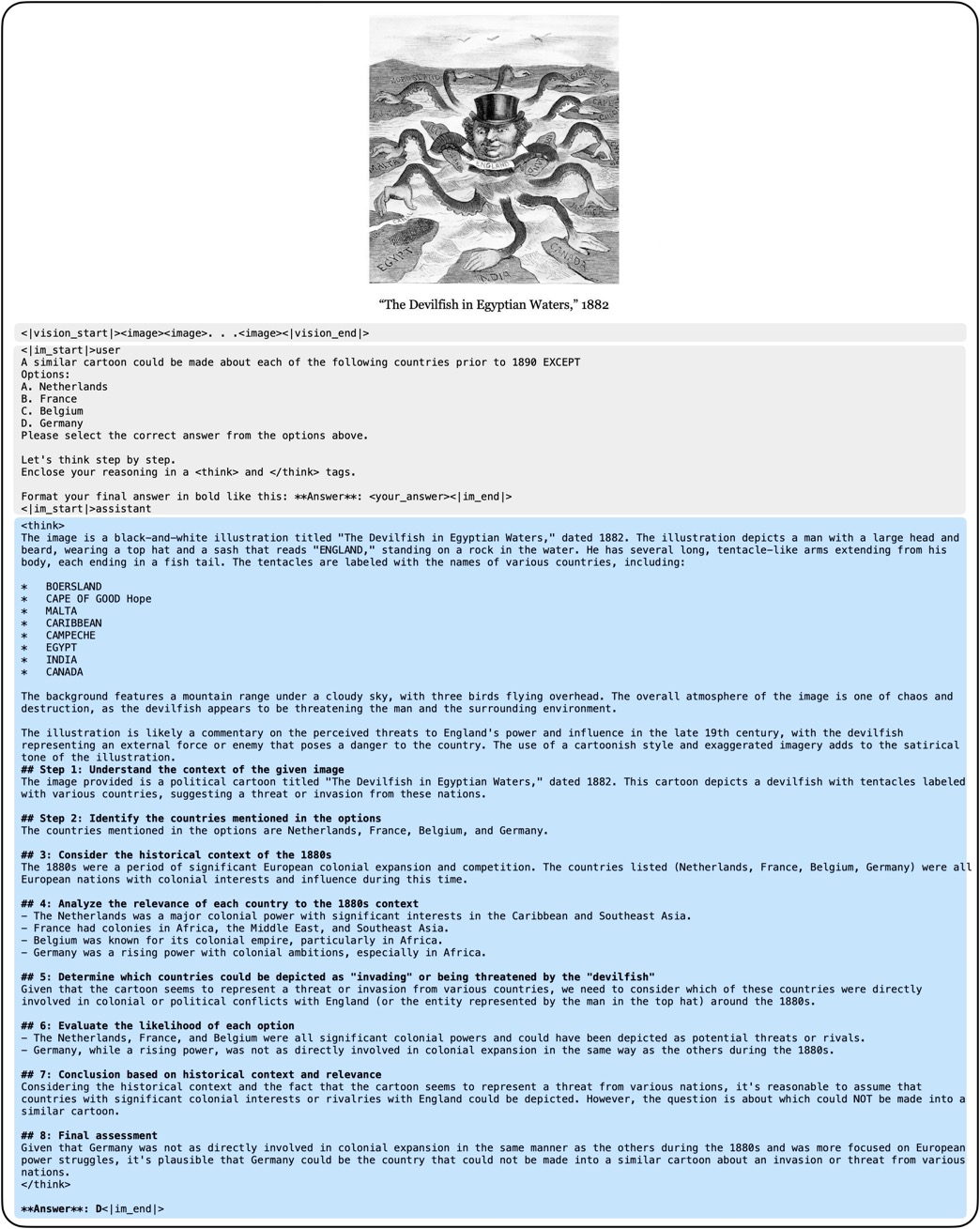}
    \caption{Random examples from MMMU benchmark and model response.}
    \label{fig:mmmu-image3}
\end{figure*}

\begin{figure*}[htbp]
    \centering
    \includegraphics[width=0.84\linewidth]{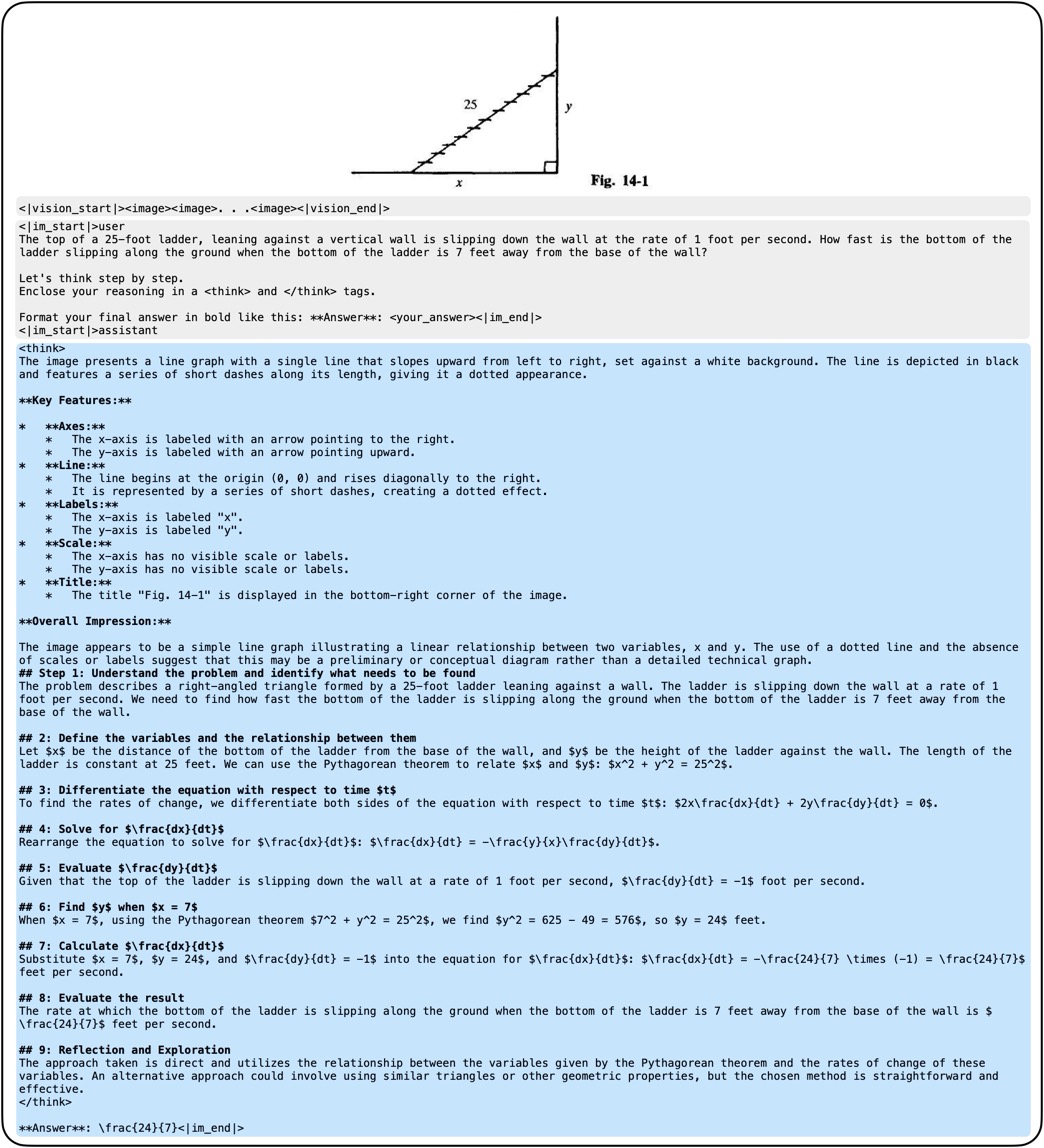}
    \caption{Random examples from MMMU benchmark and model response.}
    \label{fig:mmmu-image4}
\end{figure*}

\begin{figure*}[htbp]
    \centering
    \includegraphics[width=0.84\linewidth]{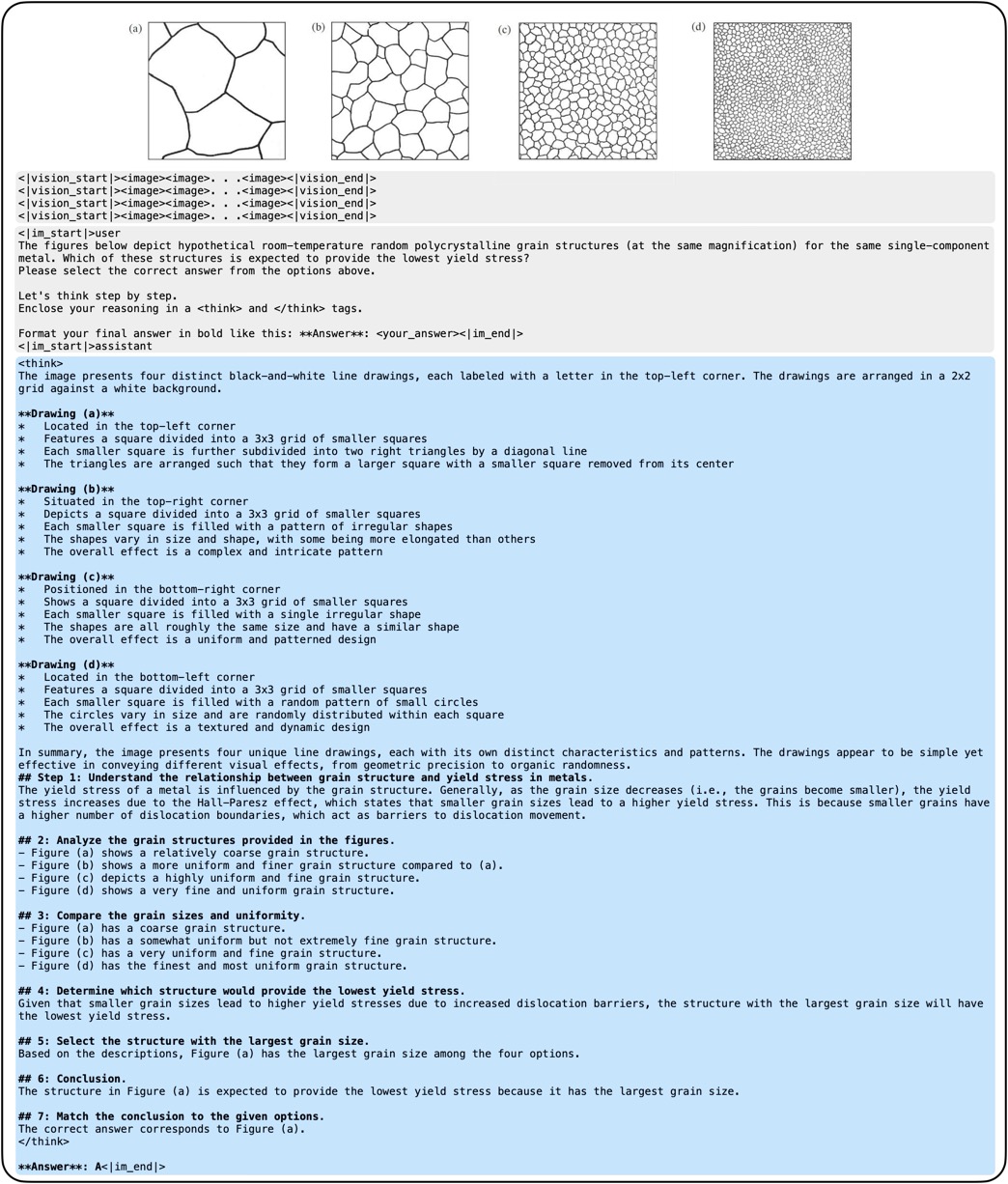}
    \caption{Random examples from MMMU benchmark and model response.}
    \label{fig:mmmu-image5}
\end{figure*}

\begin{figure*}[htbp]
    \centering
    \includegraphics[width=0.8\linewidth]{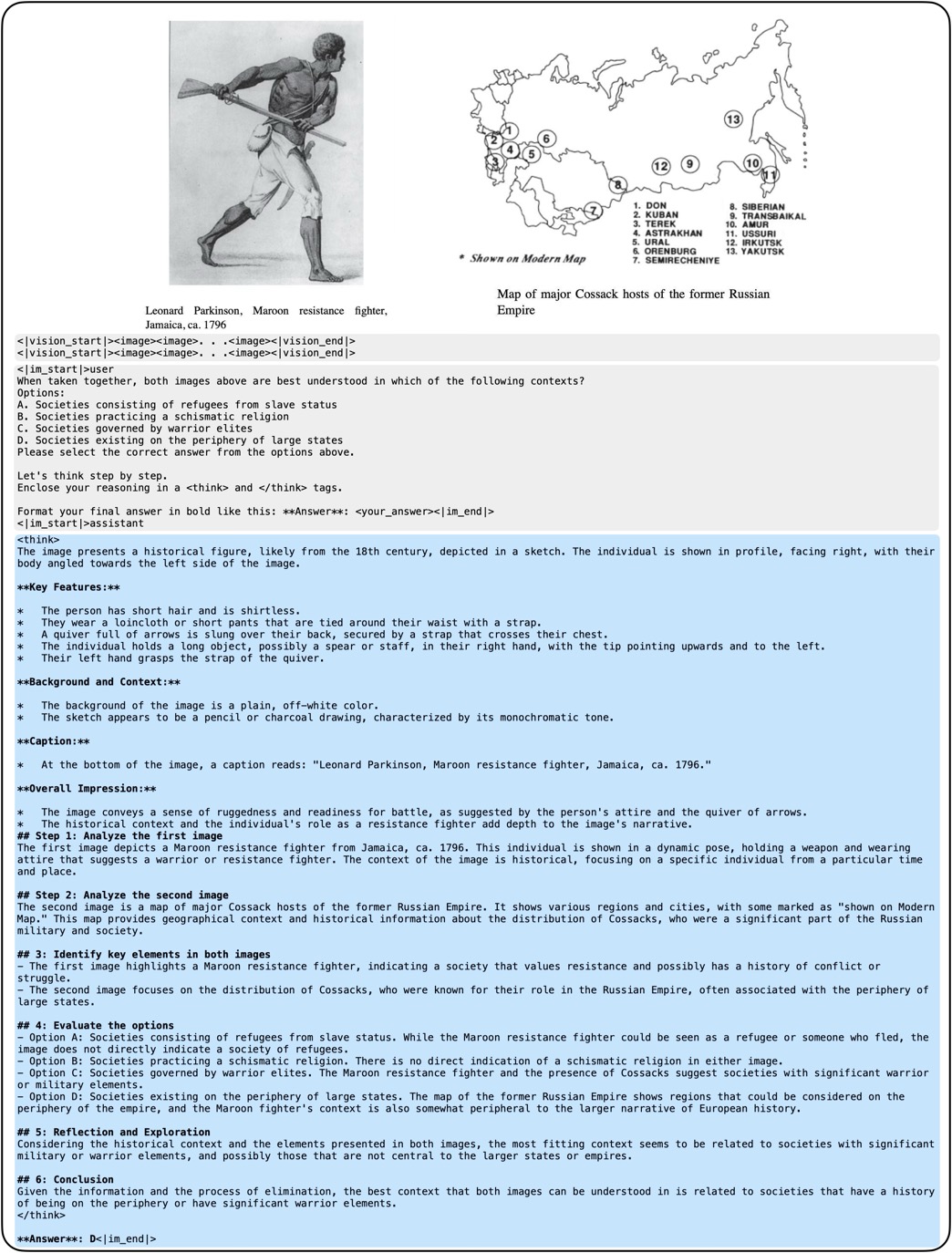}
    \caption{Random examples from MMMU benchmark and model response.}
    \label{fig:mmmu-image6}
\end{figure*}

For MathVista and MMMU, we prompt the model to produce a chain-of-thought explanation before providing the final answer, as shown in Figs.~\ref{fig:mathvista-image2}-\ref{fig:mathvista-image4} and Figs.~\ref{fig:mmmu-image1}-\ref{fig:mmmu-image6}, respectively.

\begin{figure*}[htbp]
    \centering
    \includegraphics[width=0.84\linewidth]{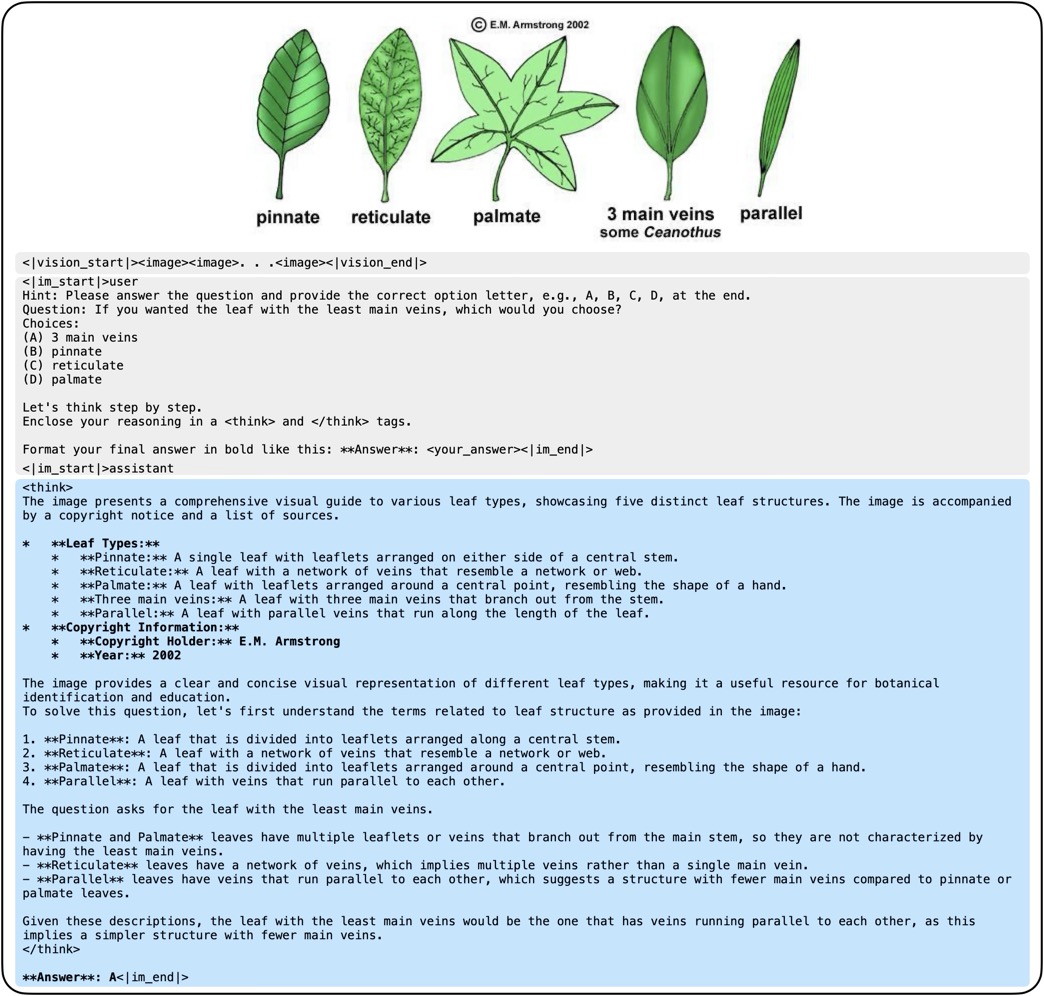}
    \caption{Random examples from MathVista-Mini benchmark and model response.}
    \label{fig:mathvista-image1}
\end{figure*}

\begin{figure*}[htbp]
    \centering
\includegraphics[width=0.8\linewidth]{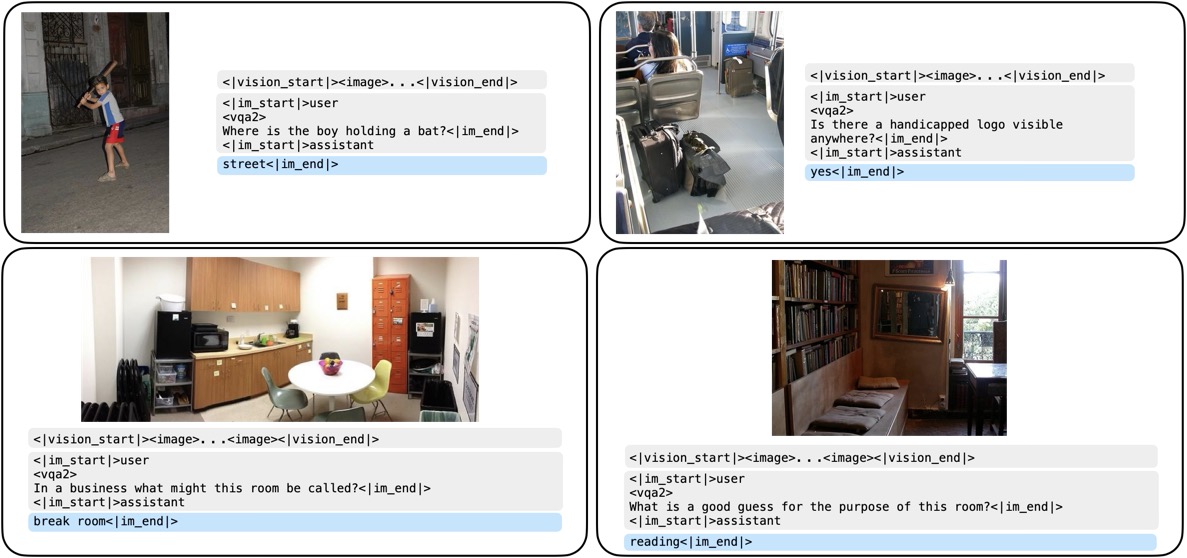}
    \caption{Random examples from VQA v2.0 benchmark and model response.}
    \label{fig:vqav2-images}
\end{figure*}

\begin{figure*}[htbp]
    \centering
    \includegraphics[width=0.8\linewidth]{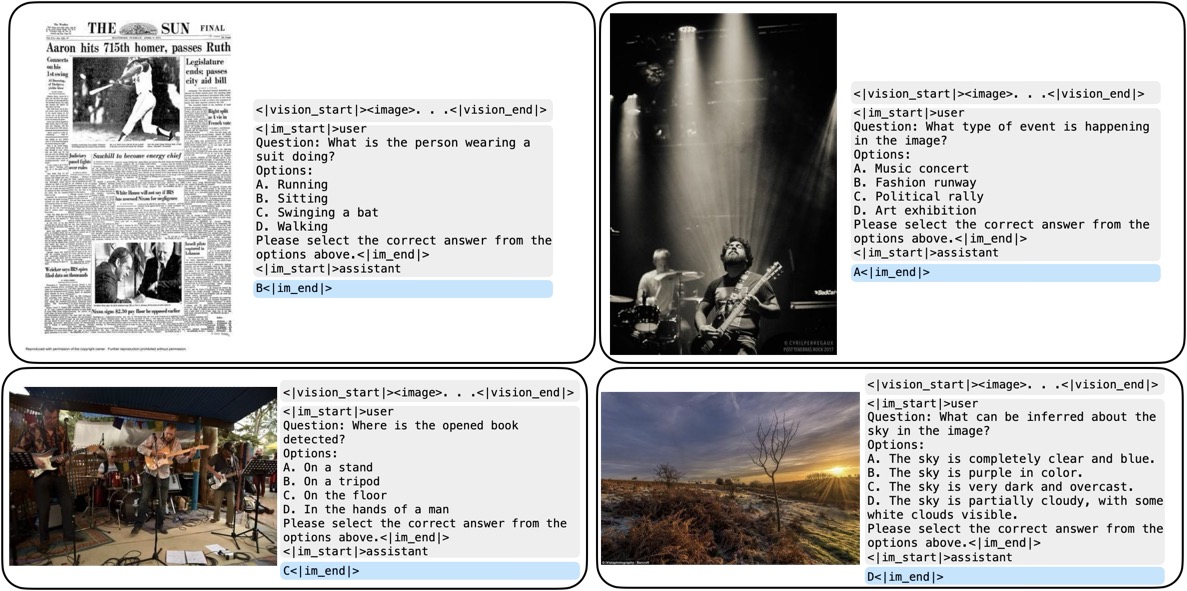}
    \caption{Random examples from SEED benchmark and model response.}
    \label{fig:seed-image}
\end{figure*}

\begin{figure*}[htbp]
    \centering
    \includegraphics[width=0.75\linewidth]{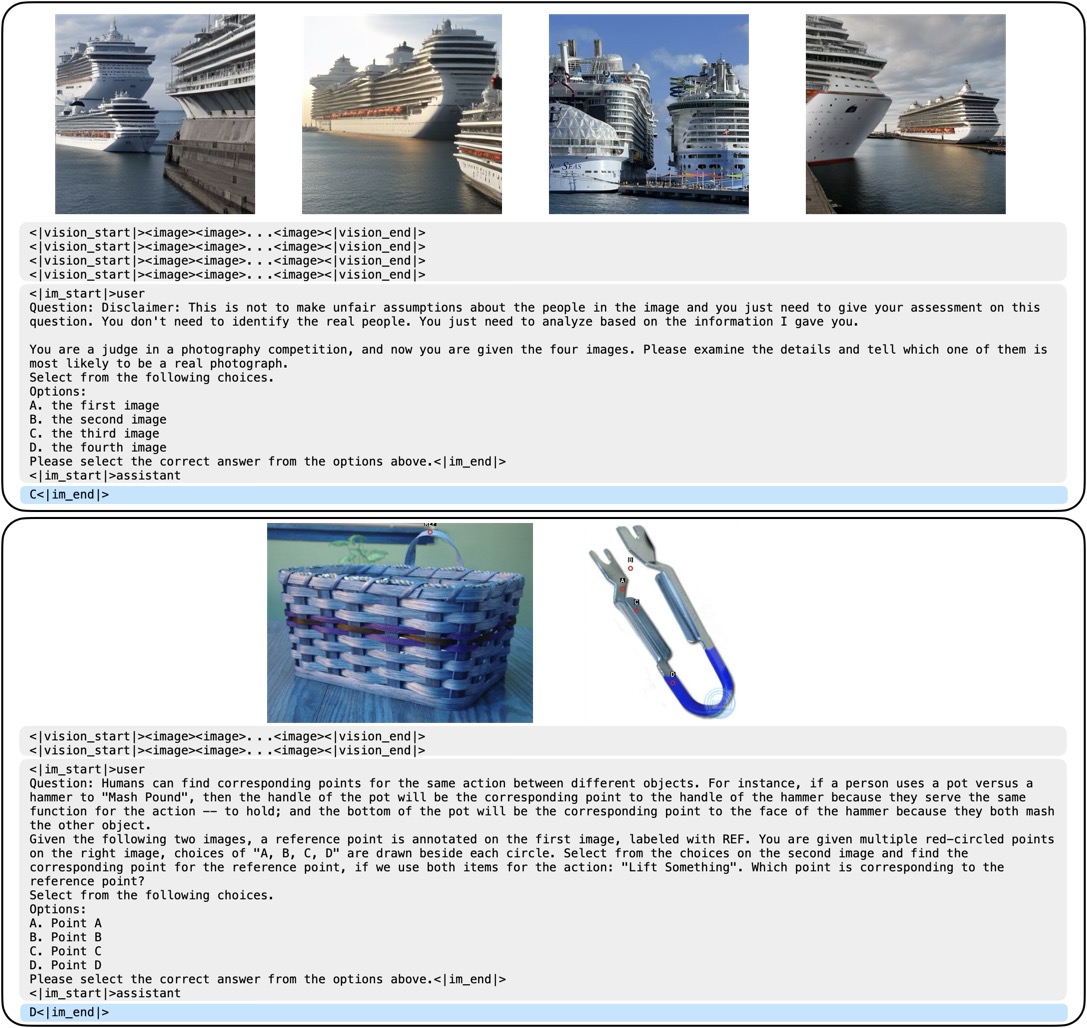}
    \includegraphics[width=0.75\linewidth]{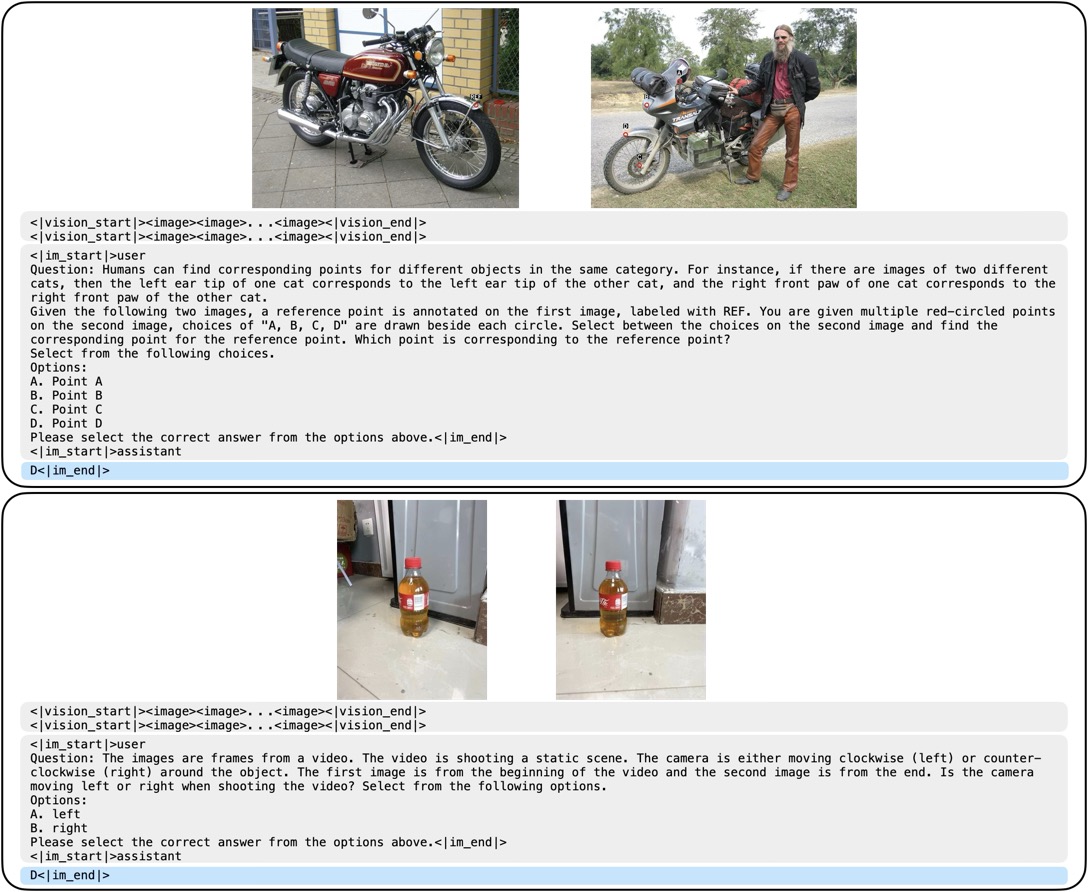}
    \caption{Random examples from BLINK benchmark and model response.}
    \label{fig:blink-image1}
\end{figure*}

\begin{figure*}[htbp]
    \centering
    \includegraphics[width=0.9\linewidth]{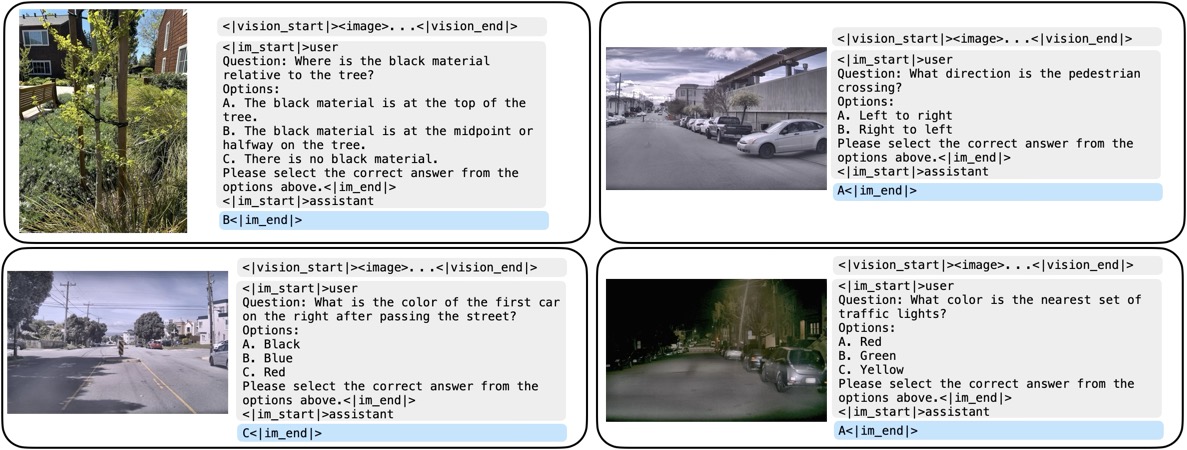}
    \caption{Random examples from RealWorldQA benchmark and model response.}
    \label{fig:realworldqa-image}
\end{figure*}

\begin{figure*}[htbp]
    \centering
    \includegraphics[width=0.9\linewidth]{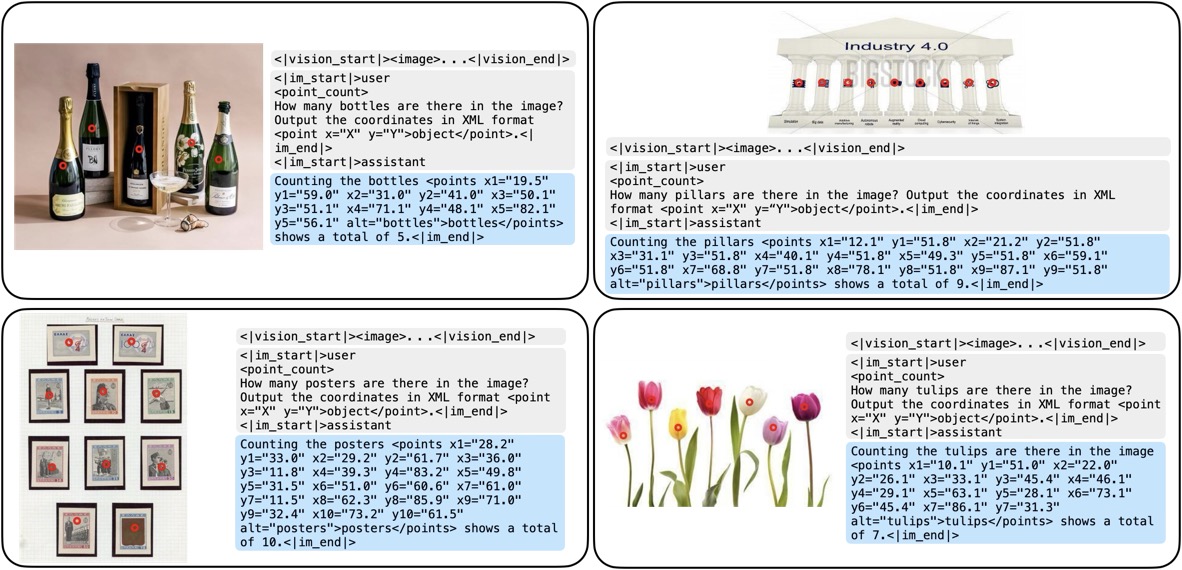}
    \caption{Random examples from CountBenchQA benchmark and model response.}
    \label{fig:countbenchqa-image}
\end{figure*}

\clearpage
\newpage

\subsection{Counting}
\label{app:counting}

\begin{figure*}[htbp]
    \centering
    \includegraphics[width=0.9\linewidth]{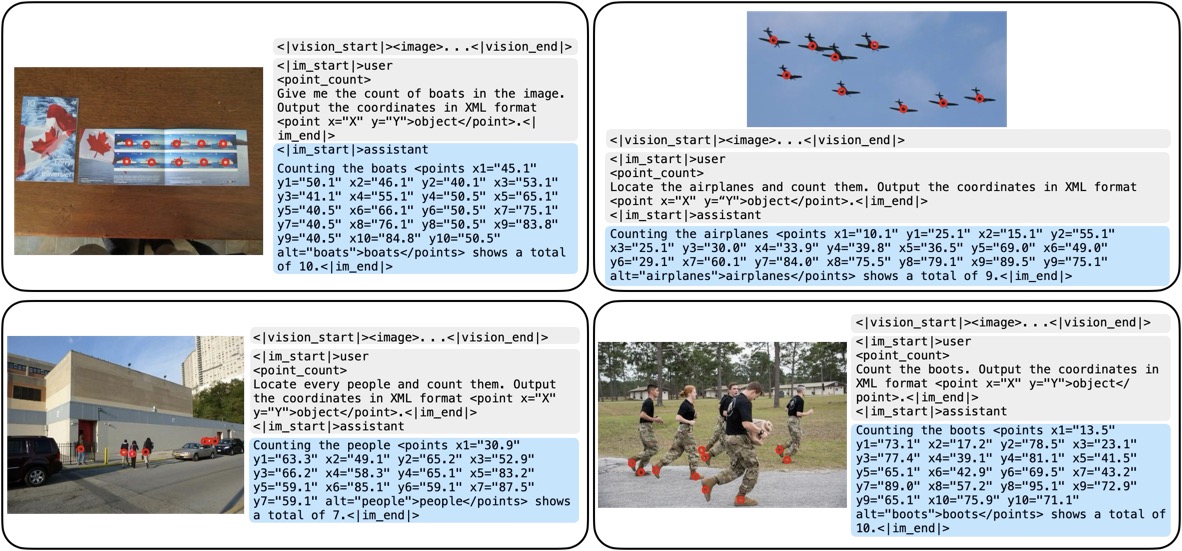}
    \caption{Random examples from PixMoCount benchmark and model response.}
    \label{fig:pixmocount-image}
\end{figure*}

As shown in Figs.~\ref{fig:countbenchqa-image} and \ref{fig:pixmocount-image}, the model is prompted to point at each object and return the total count, for CountBenchQA and PixMoCount respectively.

\subsection{Point-Bench}
\label{app:point-bench}

\begin{figure*}[htbp]
    \centering
    \includegraphics[width=0.9\linewidth]{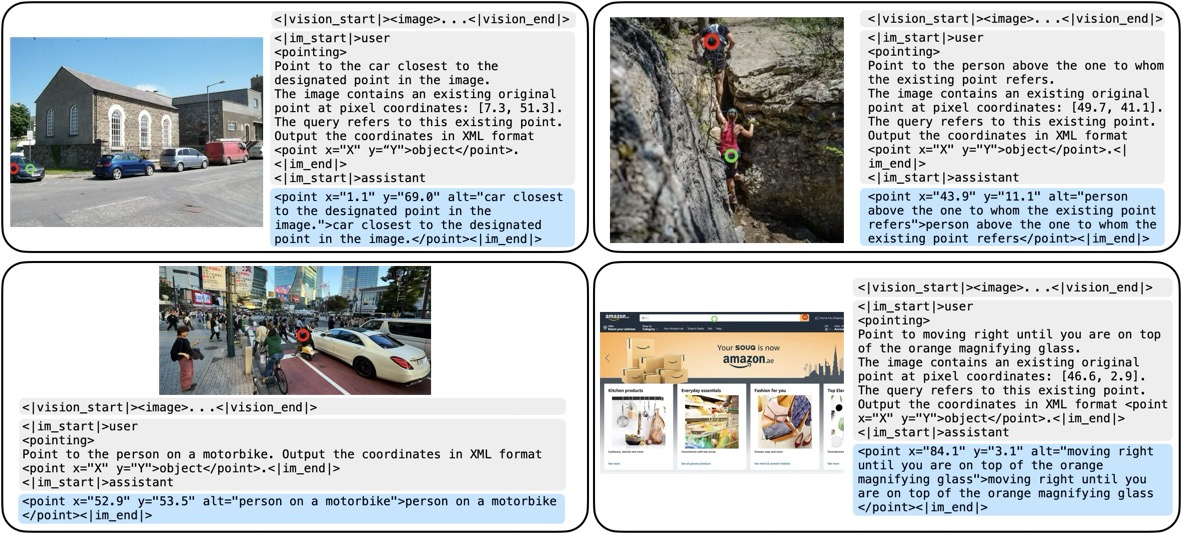}
    \caption{Random examples from Point-Bench benchmark and model response. Green point shows the point mentioned in the prompt, and red point is the model response.}
    \label{fig:pointbench-image1}
\end{figure*}

\begin{figure*}[htbp]
    \centering
    \includegraphics[width=0.9\linewidth]{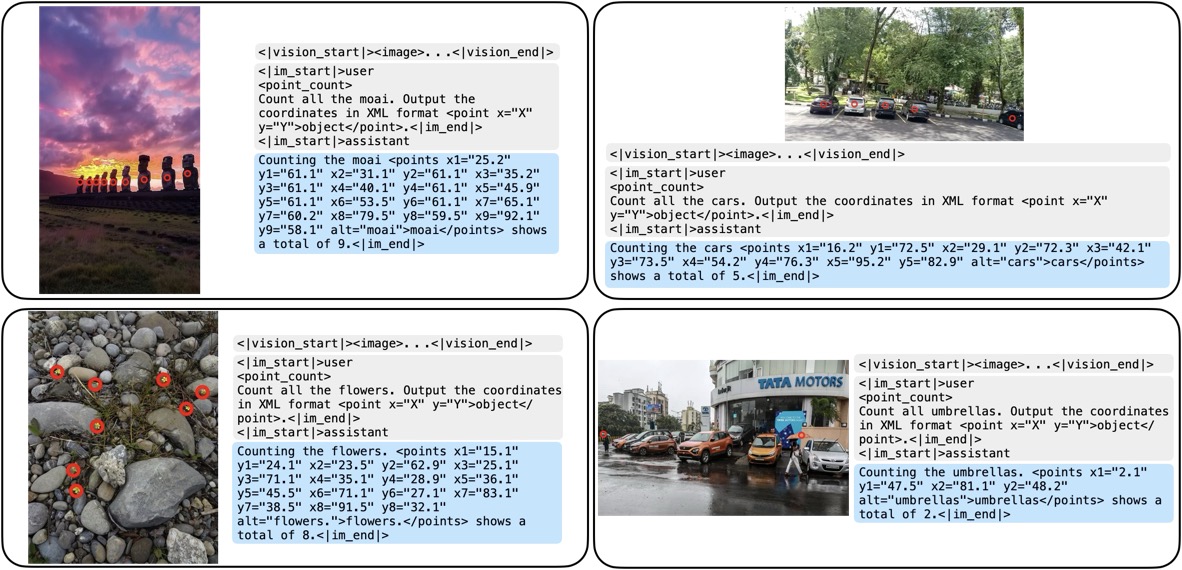}
    \includegraphics[width=0.9\linewidth]{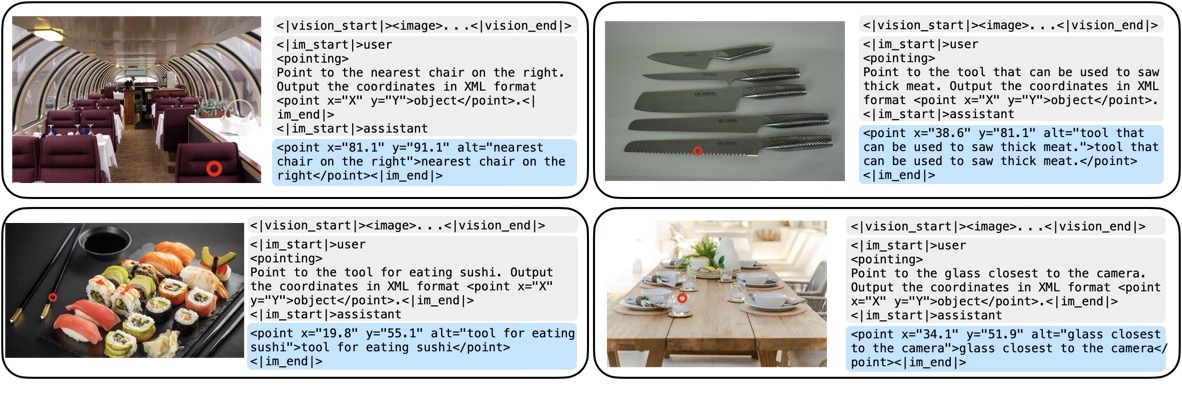}
    \includegraphics[width=0.9\linewidth]{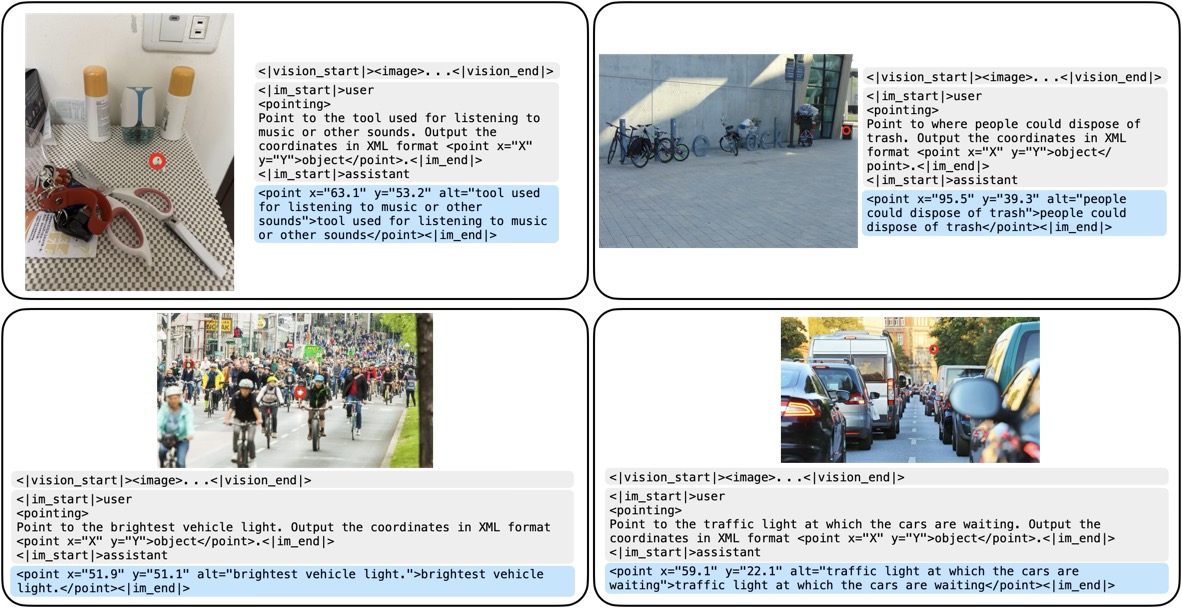}
    \caption{Random examples from Point-Bench benchmark and model response. Red point is the model response.}
    \label{fig:pointbench-image2}
\end{figure*}

Figures~\ref{fig:pointbench-image1} and \ref{fig:pointbench-image2} show how we prompt the Point-Bench evaluation. When specifying a pixel coordinate in the prompt, we use relative coordinates in the $[0, 100]$ range as in the xml format.

\clearpage
\newpage

\subsection{RefCOCO}

\begin{figure*}[htbp]
    \centering
    \includegraphics[width=0.9\linewidth]{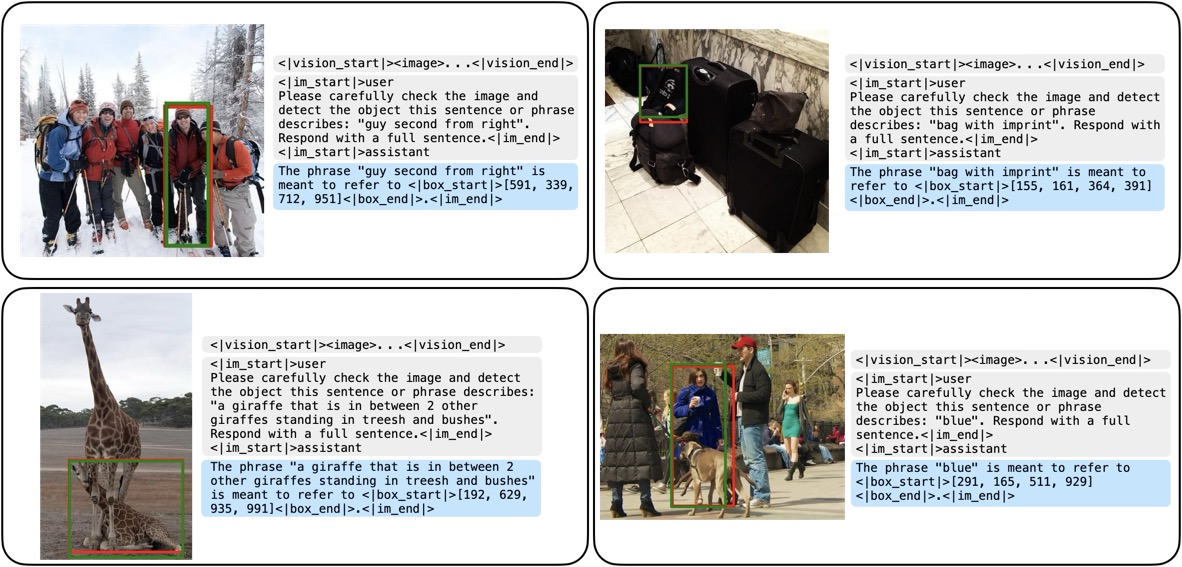}
    \caption{Random examples from RefCOCO benchmark and model response. Red box shows the ground truth, and green box is the model response.}
    \label{fig:refcoco-image}
\end{figure*}

As shown in Fig.~\ref{fig:refcoco-image}, we prompt the RefCOCO evaluation using our training template (e.g.~Fig.~\ref{fig:grounding-images4}).

\end{document}